\documentclass[10pt,twocolumn,letterpaper]{article}

\usepackage[pagenumbers]{cvpr} 

\usepackage{multirow}
\usepackage{tcolorbox}

\definecolor{beaublue}{rgb}{0.9, 0.95, 0.9}
\definecolor{blackish}{rgb}{0.2, 0.2, 0.2}

\usepackage{amsmath,amsfonts,bm}

\def\eqref#1{equation~\ref{#1}}

\def\1{\bm{1}}

\def\vx{{\bm{x}}}

\def\mX{{\bm{X}}}

\DeclareMathAlphabet{\mathsfit}{\encodingdefault}{\sfdefault}{m}{sl}
\SetMathAlphabet{\mathsfit}{bold}{\encodingdefault}{\sfdefault}{bx}{n}
\newcommand{\tens}[1]{\bm{\mathsfit{#1}}}

\def\tF{{\tens{F}}}

\def\tX{{\tens{X}}}

\def\gF{{\mathcal{F}}}

\def\gI{{\mathcal{I}}}

\def\gS{{\mathcal{S}}}
\def\gT{{\mathcal{T}}}

\def\sR{{\mathbb{R}}}

\definecolor{cvprblue}{rgb}{0.21,0.49,0.74}
\usepackage[pagebackref,breaklinks,colorlinks,allcolors=cvprblue]{hyperref}

\def\paperID{9116} 
\def\confName{CVPR}
\def\confYear{2025}

\title{When Spatial meets Temporal in Action Recognition}

\author{Huilin Chen$^{1,}$\thanks{These authors contributed equally.} \qquad Lei Wang$^{1,2, *, }$\thanks{Corresponding author.} \qquad Yifan Chen$^{1, *}$ \qquad Tom Gedeon$^3$ \qquad Piotr Koniusz$^{2, 1}$\\
$^1$Australian National University, $^2$Data61/CSIRO, $^{3}$Curtin University\\
\{u7326198, lei.w, u7725118\}@anu.edu.au, \\tom.gedeon@curtin.edu.au, piotr.koniusz@data61.csiro.au
}

\begin{document}
\maketitle
\begin{abstract}
Video action recognition has made significant strides, but challenges remain in effectively using both spatial and temporal information. While existing methods often focus on either spatial features (\eg, object appearance) or temporal dynamics (\eg, motion), they rarely address the need for a comprehensive integration of both. Capturing the rich temporal evolution of video frames, while preserving their spatial details, is crucial for improving accuracy.
In this paper, we introduce the Temporal Integration and Motion Enhancement (TIME) layer, a novel preprocessing technique designed to incorporate temporal information. The TIME layer generates new video frames by rearranging the original sequence, preserving temporal order while embedding $N^2$ temporally evolving frames into a single spatial grid of size $N \times N$. This transformation creates new frames that balance both spatial and temporal information, making them compatible with existing video models. When $N=1$, the layer captures rich spatial details, similar to existing methods. As $N$ increases ($N\geq2$), temporal information becomes more prominent, while the spatial information decreases to ensure compatibility with model inputs.
We demonstrate the effectiveness of the TIME layer by integrating it into popular action recognition models, such as ResNet-50, Vision Transformer, and Video Masked Autoencoders, for both RGB and depth video data. Our experiments show that the TIME layer enhances recognition accuracy, offering valuable insights for video processing tasks.
\end{abstract}    
\section{Introduction}
\label{sec:intro}

Video action recognition~\cite{simonyan2014two,tran2015learning,feichtenhofer2016convolutional,wang2016temporal,carreira2017quo,wang2019hallucinating,Ryoo2020AssembleNet,wang2021self,bertasius2021space,patrick2021keeping,tong2022videomae,wang2023videomae,girdhar2023omnimae,wang2024flow,wang2024high,wang2024internvideo2,chen2024motion,msad2024} has become a pivotal area in computer vision, with applications spanning from autonomous vehicles and surveillance systems to human-computer interaction and sports analytics. Despite the progress made by advanced 3D convolutional networks~\cite{tran2015learning,carreira2017quo} and transformer-based models~\cite{dosovitskiy2021an,patrick2021keeping,bertasius2021space}, a core challenge remains: efficiently capturing both visual appearances and temporal dynamics within videos without excessive computational cost. Many models rely on sparse frame sampling~\cite{wang2016temporal} to limit memory usage, potentially overlooking critical motion that is essential for understanding complex actions~\cite{patrick2021keeping,chen2024motion}.

Recent research has emphasized the need for approaches that can flexibly adapt to both spatial and temporal demands of video data~\cite{wangtaylor,chen2024motion}. However, methods that heavily rely on densely sampled sequences often encounter scalability issues, while those focused on spatial detail alone may underperform in capturing motion cues~\cite{tong2022videomae,wang2023videomae}. To bridge this gap, we introduce the Temporal Integration and Motion Enhancement (TIME) layer, a preprocessing layer designed to maximize the temporal richness in video frames while preserving spatial fidelity, transforming standard image classifiers like ResNet~\cite{he2016deep} and Vision Transformer (ViT)~\cite{dosovitskiy2021an} into efficient video action recognition models. By enhancing input frames with motion cues, the TIME layer addresses the limitations of sparse sampling approaches and enables detailed, continuous temporal analysis within each frame, even for models traditionally optimized for static images.

The TIME layer’s adaptable design offers several advantages: (i) It balances spatial and temporal information through an adjustable grid structure, allowing possible tuning based on the specific action recognition requirements. (ii) The TIME layer operates independently of the model architecture, making it compatible with a wide range of models, from CNNs to transformers, as well as with emerging self-supervised learning approaches like VideoMAE. (iii) It facilitates systematic evaluation of spatial-temporal information by arranging frames in novel configurations, providing a new avenue for investigating how model architectures absorb and interpret motion and visual cues. This versatility enables TIME to serve as both an effective video preprocessing tool and a diagnostic layer, offering insights into model responses to varied spatial-temporal compositions.

In this work, we evaluate the TIME layer on both RGB and depth video datasets across different action recognition benchmarks, examining its impact on small-scale and large-scale datasets, from-scratch training versus fine-tuning, and CNN versus transformer. We also explore how TIME enhances temporal context within each frame, making it a valuable tool for applications requiring a nuanced balance between visual appearance and motion dynamics. Through experiments, we demonstrate that the TIME layer improves model performance and provides a systematic framework for analyzing the role of temporal information in action recognition. Our \textbf{main contributions} are as follows:
\renewcommand{\labelenumi}{\roman{enumi}.}
\begin{enumerate}[leftmargin=0.6cm]
\item We introduce the TIME layer, a novel preprocessing approach that enhances temporal representation in video frames while preserving spatial fidelity.
\item We demonstrate that the TIME layer is compatible with a broad range of models, including CNN, transformer, and self-supervised learning framework, making it a versatile component for existing pipelines.
\item Extensive experiments show that the TIME layer improves performance across diverse action recognition benchmarks, including adapting RGB-pretrained models to depth videos.
\item The TIME layer functions as a tool for systematically evaluating spatial-temporal information processing, offering new insights into model behavior on video data.
\end{enumerate}

\begin{figure}[tbp]
\centering
\begin{subfigure}[b]{0.495\linewidth}
\centering\includegraphics[trim=4.5cm 0 4.5cm 0, clip=true,width=\linewidth]{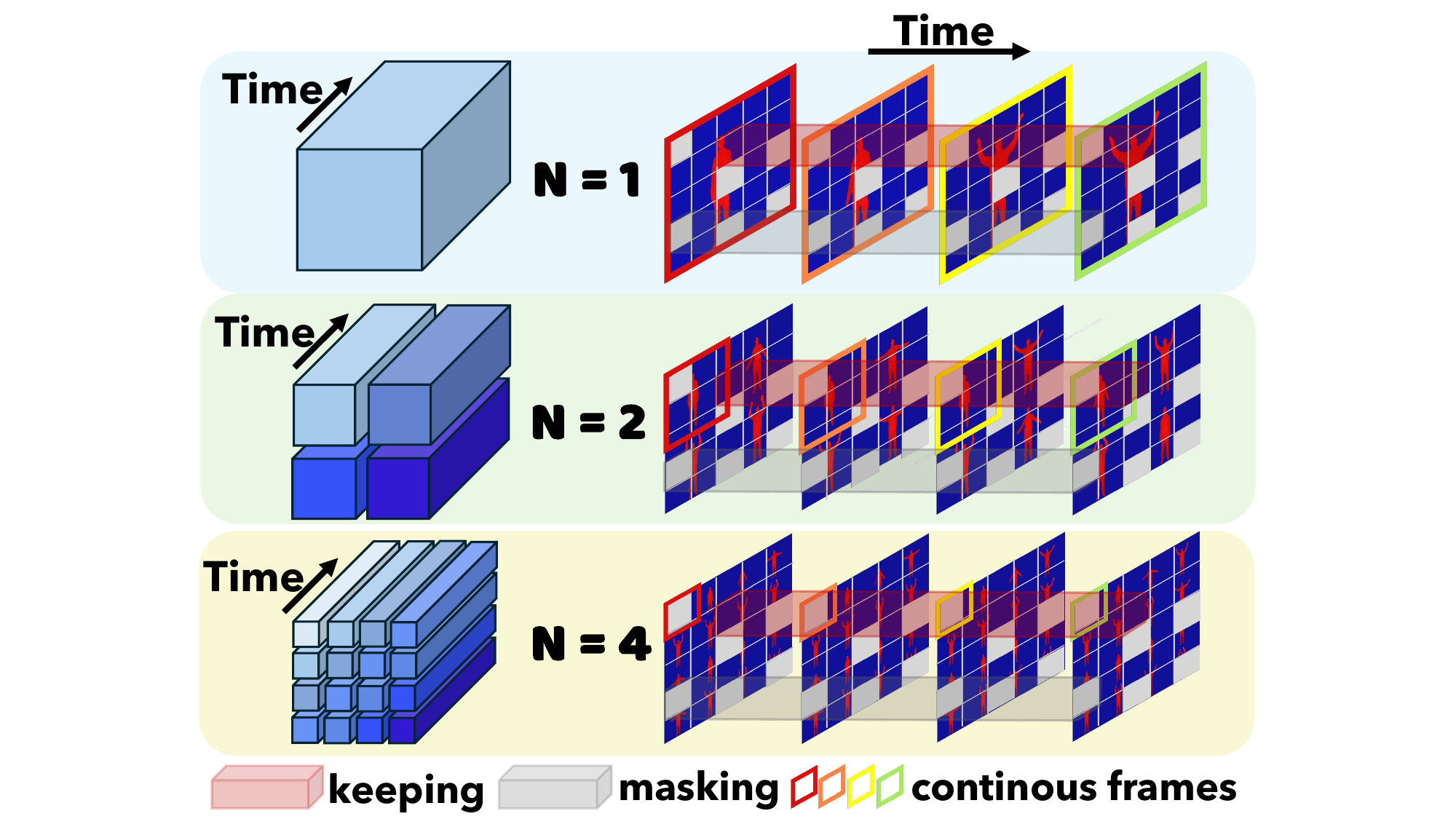}
\caption{\label{fig:spatial-time}Spatial block arrangement.}
\end{subfigure}\hfill
\begin{subfigure}[b]{0.495\linewidth}
\centering\includegraphics[trim=4.5cm 0 4.5cm 0, clip=true,width=\linewidth]{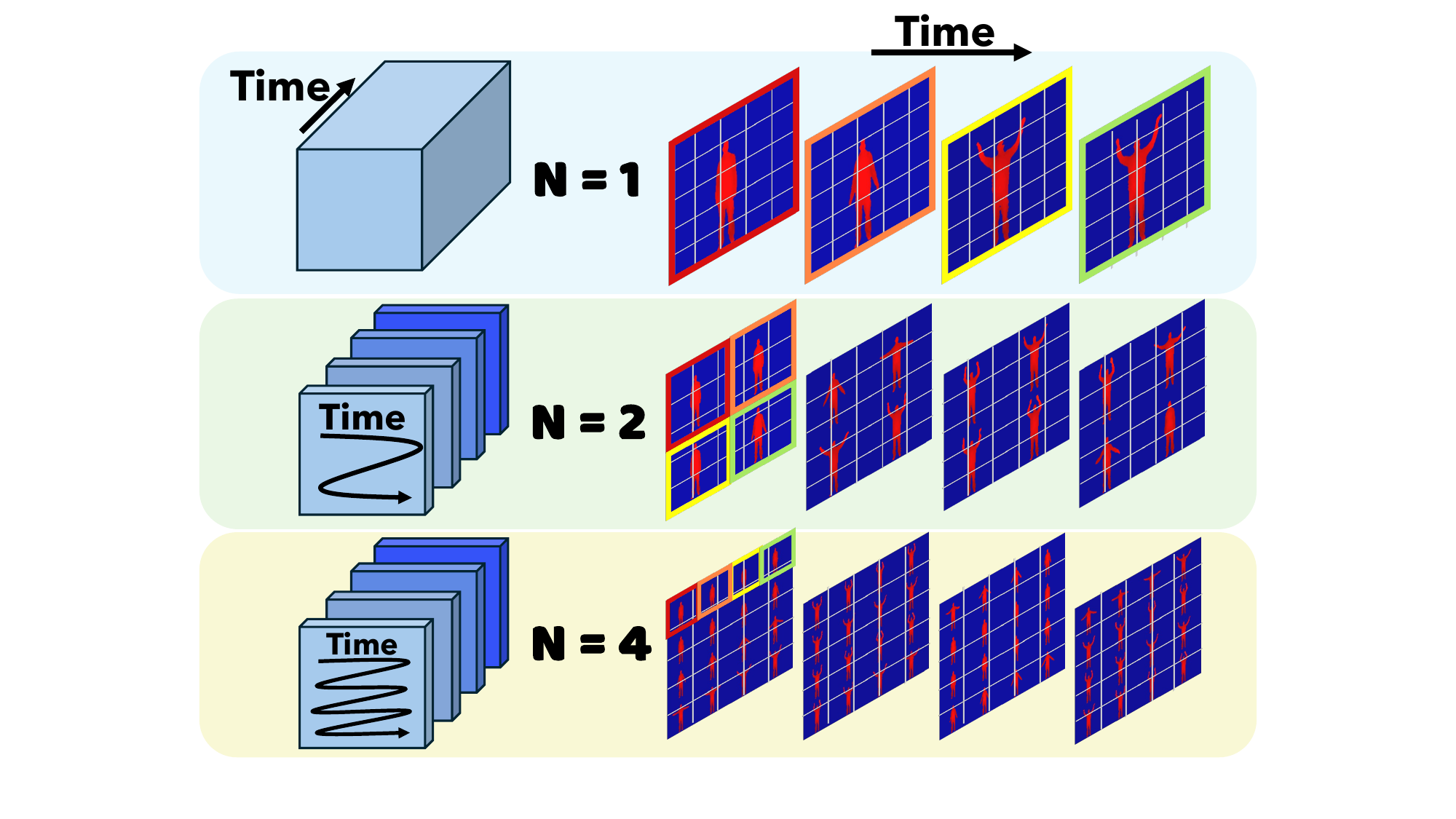}
\caption{\label{fig:temporal-time}Temporal block arrangement.}
\end{subfigure}
\caption{Variants of the TIME layer. The spatial block arrangement captures long-term motion spatially by incorporating broad visual changes across frames, while it emphasizes short-term motion within each frame's temporal sequence (tube masking, as shown in (a), temporally obscures short-term motion details for VideoMAE). In contrast, the temporal block arrangement records short-term motion frame by frame, yet captures long-term motion temporally across the sequence (tube masking here obscures long-term motion). Each approach provides a unique perspective on balancing spatial and temporal information through the spatial-temporal balance parameter, $N$, in video data.}
\label{fig:timelayer}
\end{figure}

\section{Related Work}
\label{sec:related}

In this section, we identify gaps in existing video action recognition methods and highlight the need for adaptable, computationally efficient layers like the TIME layer, which enrich temporal dynamics in standard architectures.

\noindent\textbf{Video action recognition} has progressed from early 2D CNNs that process frames individually~\cite{he2016deep} to more advanced 3D CNNs~\cite{carreira2017quo} and transformer-based models capable of capturing spatial-temporal dependencies~\cite{patrick2021keeping}. Initial approaches, such as C3D~\cite{tran2015learning}, used 3D convolutions to model relationships across frames. Further improvements, like I3D~\cite{carreira2017quo} and SlowFast~\cite{feichtenhofer2019slowfast}, introduced dual-pathway architectures to handle both slow and fast motion patterns, but with high computational costs. Recent transformers, including ViViT~\cite{arnab2021vivit} and TimeSformer~\cite{bertasius2021space}, use self-attention for long-term temporal modeling but are similarly resource-intensive, underscoring the need for efficient temporal processing. To capture motion dynamics, Taylor videos~\cite{wangtaylor} and motion prompts~\cite{chen2024motion} have been introduced to enhance temporal cues in video frames by emphasizing motion and object displacement.

Our TIME layer addresses these challenges by offering an architecture-agnostic, flexible module that enhances temporal information within frames, aiming to bridge gaps in computational efficiency and adaptability across models.  

\noindent\textbf{Sparse sampling and temporal efficiency.} To mitigate the computational demands of video action recognition, sparse sampling techniques reduce the number of processed frames, lowering memory and processing costs. Temporal Segment Networks (TSN)~\cite{wang2016temporal,8454294} popularized uniformly sampling frames to capture temporal moments without exhaustive frame processing. However, sparse sampling may miss fine-grained motion cues in longer sequences~\cite{koniusz2021tensor,wang2024flow}. Recent approaches, such as Temporal Pyramid Networks (TPN)~\cite{yang2020temporal} and Temporal Shift Module (TSM)~\cite{lin2019tsm}, aim to improve temporal representation without added computational cost, although they still rely on selected frames.

In contrast, our TIME layer captures the entire video sequence by restructuring frames to embed richer temporal dynamics while preserving spatial details, achieving a more comprehensive motion representation.

\noindent\textbf{Self-supervised video representation learning.} Self-supervised learning (SSL) has gained traction in video modeling by enabling pretraining without labeled data. Early SSL methods~\cite{wang2019hallucinating, wang2021self} used techniques like Bag of Words and Fisher Vectors to improve action recognition. Recent SSL models, VideoMAE~\cite{tong2022videomae} and its advanced version~\cite{wang2023videomae}, use masked autoencoding to learn spatial-temporal features by reconstructing occluded regions, effectively capturing motion and visual cues. However, SSL methods typically depend on sparse frame sampling or masking on consecutive frames, which may limit their capacity to fully capture both short-term and long-term temporal patterns.

Our TIME layer reorganizes video frames, allowing an adjustable spatial-temporal balance and facilitating a deeper exploration of action recognition trade-offs across varying temporal scales.

\noindent\textbf{Spatial-temporal balancing} is essential for video action recognition. Many traditional approaches face challenges in maintaining this balance, as more temporal information may reduce spatial clarity~\cite{wangtaylor}. Methods like dynamic images~\cite{Bilen_2016_CVPR}, Temporal Context Networks~\cite{dai2017temporal}, and X3D~\cite{feichtenhofer2020x3d} capture multi-scale temporal features but often remain architecture-specific. Additionally, combining consecutive RGB channels into a single frame for better motion detection~\cite{kim2022capturing} has shown potential, as have Taylor videos for capturing dominant motion in actions~\cite{wangtaylor}; however, these methods often lack the flexibility to record either finer temporal or spatial details.

Our TIME layer provides a systematic approach to balancing spatial and temporal information, positioning it as an adaptable tool for examining model sensitivities to spatial and temporal variations, thereby opening new avenues for video-related research, \eg, action recognition.

\section{Approach}
\label{sec:approach}

In this section, we introduce our Temporal Integration and Motion Enhancement (TIME) layer and provide insights into our approach. We begin by defining the notations.

\noindent\textbf{Notations.} Let $\gI_T = \{1, 2, \dots, T\}$ represent an index set. Scalars are denoted by regular fonts, \eg, $x$; vectors by lowercase boldface, \eg, $\vx$; matrices by uppercase boldface, \eg, $\mX$; and tensors by calligraphic letters, \eg, $\tX$.

\subsection{TIME Layer}
\label{sec:time}


Consider a video represented by a sequence $\gF=[\tF_1, \tF_2, \dots, \tF_T]$, where $T$ is the total number of frames, and each frame $\tF_t \in \sR^{H \times W \times 3}$ ($t \in \gI_T$). The TIME layer processes frames in temporal order, efficiently using all frames to preserve fine motion cues in contrast to traditional sparse sampling techniques~\cite{wang2016temporal}.
The TIME layer is independent of the model’s architecture, allowing flexibility in frame resolution and input count. Given a target resolution of $H^*\!\times\!W^*$ and a frame count of $T^*$, the layer arranges each new frame by organizing an $N\! \times\! N$ grid of temporally ordered placeholders. If the video length $T$ is shorter than $T^*N^2$, frames are repeated to reach the necessary length; otherwise, $T^*N^2$ frames are sampled to maintain broad temporal coverage.
To construct each new frame, we use two main approaches, each enhancing temporal detail in distinct ways. \cref{fig:timelayer} shows the two variants of TIME layer. These methods are described below.

\noindent\textbf{Spatial block arrangement.} In this approach, we divide the $T^*N^2$ sequence into $N^2$ temporal blocks, each containing $T^*$ consecutive frames:
\begin{equation}
    \gT_i = [\tF_{(i-1)T^* + 1}, \tF_{(i-1)T^* + 2}, \dots, \tF_{i T^*}],
    \label{eq:spatialblock}
\end{equation}
where $i = 1, 2, \dots, N^2$, and each $\gT_i \in \sR^{H\times W\times 3\times T^*}$ represents a segment of $T^*$ frames from the video. We then arrange these blocks spatially in an $N \times N$ grid to form a new frame $\gF^{\text{spatial}}$, with each grid cell corresponding to frames from a specific segment:
\begin{equation}
    \gF^{\text{spatial}} = 
   \begin{bmatrix}
   \gT_1 & \gT_2 & \dots & \gT_N \\
   \gT_{N+1} & \gT_{N+2} & \dots & \gT_{2N} \\
   \vdots & \vdots & \ddots & \vdots \\
   \gT_{N(N-1)+1} & \gT_{N(N-1)+2} & \dots & \gT_{N^2} 
   \end{bmatrix}.
   \label{eq:spatial}
\end{equation}
This spatial configuration in $\gF^{\text{spatial}}$ captures the progression of $T^*$ frames within each grid cell, offering a temporally enriched view of motion continuity.

\noindent\textbf{Temporal block arrangement.} Alternatively, we organize frames by dividing the $T^*N^2$ sequence into $T^*$ blocks, each containing $N^2$ frames, enabling detailed temporal evolution within each new frame. The blocks are defined as:
\begin{equation}
    \gS_j = [\tF_{(j-1)N^2 + 1}, \tF_{(j-1)N^2 + 2}, \dots, \tF_{jN^2}], 
    \label{eq:temporalblock}
\end{equation}
where $j = 1, 2, \dots, T^*$, and each $\gS_j$ represents a contiguous segment of $N^2$ frames. For each block $\gS_j$, we arrange the frames into an $N \times N$ grid to form a new frame $\tF^{\text{new}}_j$:
\begin{equation}
    \tF^{\text{new}}_j = 
   \begin{bmatrix}
   \tF_{(j-1)N^2 + 1} & \dots & \tF_{(j-1)N^2 + N} \\
   \tF_{(j-1)N^2 + N + 1} & \dots & \tF_{(j-1)N^2 + 2N} \\
   \vdots & \ddots & \vdots \\
   \tF_{(j-1)N^2 + (N-1)N + 1} & \dots & \tF_{jN^2} 
   \end{bmatrix}.
\end{equation}
This results in each new frame $\tF^{\text{new}}_j$ encapsulating temporally rich content within a spatial layout. For simplicity, we denote this operation as $\tF^{\text{new}}_j = r(\gS_j)$. Thus, we construct the sequence:
\begin{equation}
    \gF^{\text{temporal}} = [r(\gS_1), r(\gS_2), \dots, r(\gS_{T^*})],
    \label{eq:temporal}
\end{equation}
which captures the temporal evolution across consecutive frames.
Once the new video frames are structured, we resize them to match the model’s input dimensions ($H^*\! \times\!W^*$, generally with $H^*\!=\!W^*$). Initially, data augmentations such as cropping, scaling, flipping, and rotation are applied to the original high-resolution frames, which are then resized to fit within the $N \times N$ grid, optimizing memory efficiency.

The TIME layer’s flexibility is governed by the parameter $N$, balancing spatial and temporal detail. We refer to $N$ as the spatial-temporal balance parameter. When $N=1$, the layer aligns with conventional frame sampling methods~\cite{wang2016temporal}. As $N$ increases, temporal information becomes more pronounced within each frame, though spatial detail decreases per frame. This trade-off enables the TIME layer to enhance temporal dynamics selectively while retaining essential spatial context.


\subsection{TIME Layer Across Models}
\label{sec:application}

Our experiments apply the TIME layer to a range of foundational architectures for action recognition: CNN-based models (\eg, ResNet-50), Transformer-based models (\eg, ViT), and self-supervised video models (\eg, VideoMAE). These architectures represent key approaches in frame-based and video-based learning, allowing us to explore the TIME layer's adaptability in capturing temporal dynamics.

\noindent\textbf{ResNet-50 with the TIME Layer.} Incorporating the TIME layer into ResNet-50 enables it to handle temporal dependencies directly within frame-based processing, enhancing spatial patterns by embedding long-term and short-term temporal cues into each frame. This allows the CNN, typically optimized for spatial tasks (\eg, image classification), to improve in tasks that require motion sensitivity.

\noindent\textbf{ViT with the TIME Layer.} For ViT, a model structured for efficient spatial patch processing, the TIME layer introduces spatio-temporal patches, capturing both short- and long-term temporal dynamics without altering ViT’s framework. By encoding long-term motion through uniform sampling and capturing immediate motion continuity per frame, the TIME layer enhances ViT’s utility for video data.

\noindent\textbf{VideoMAE with the TIME Layer}. In self-supervised contexts, the TIME layer integrates naturally with VideoMAE's tube masking, enabling it to handle both short-term and long-term dynamics. Extending VideoMAE’s capabilities to depth video, we demonstrate how the TIME layer’s integration across different data modalities provides comprehensive spatio-temporal information, making it valuable for tasks involving complex action dynamics.

These applications illustrate the TIME layer’s flexibility across varied model architectures and data types, confirming its potential as a universal module in video action recognition. By enriching models with temporal awareness and supporting diverse modalities, the TIME layer offers a solid pathway toward more accurate and adaptable action recognition systems. Below, we provide key insights into how the TIME layer enhances video action recognition.

\subsection{The Role of TIME Layer}

\noindent\textbf{Adapting image models for video tasks.} The TIME layer’s integration with standard image classifiers like ResNet-50 and ViT transforms them into effective video processors without requiring major architectural changes. By enriching frames with temporal information, the TIME layer equips these models to handle complex motion cues essential for action recognition, bridging the gap between image-based architectures and video data requirements.

\noindent\textbf{Adjustable spatial-temporal balance.} Using an $N\times N$ grid structure, the TIME layer balances spatial and temporal details within each frame, controlled by $N$. Smaller $N$ values emphasize spatial clarity, preserving fine visual details, while larger values incorporate greater temporal dynamics, revealing motion over time. This adjustable framework enables researchers to finely tune the spatial-temporal balance, offering insights into model sensitivity to these factors and supporting deeper analysis of action recognition needs.

\noindent\textbf{Diagnostic tool for model interpretation.} The TIME layer functions as a diagnostic tool, showcasing how models absorb and interpret spatial and temporal cues across various setups. By testing continuous sequences, temporally ordered grids, and combinations of short- and long-term dynamics, the TIME layer provides insights into how CNNs, transformers, and self-supervised models like VideoMAE respond to detailed spatial-temporal inputs, enhancing our understanding of model interpretability and adaptability.

\noindent\textbf{Systematic evaluation of spatial-temporal information.} Unlike traditional methods that use sparse sampling or treat frames independently, the TIME layer uses the full video sequence, enriching temporal information within each spatial frame structure. This systematic approach allows for precise manipulation of spatial-temporal elements, providing a powerful tool to assess the critical balance of motion and visual cues in action recognition.

\begin{table}[tbp]
\setlength{\tabcolsep}{0.15em}
\renewcommand{\arraystretch}{0.70}
\fontsize{9}{9}\selectfont
\begin{center}
\resizebox{\linewidth}{!}{\begin{tabular}{ l c c c c c c }
\toprule
Datasets & Classes & Subjects & Views & Video clips& Modalities \\ 
\midrule
MSRAction3D~\cite{li_msraction3d} & 20 & 10 & 1 & 567 &  Depth\\
3D Action Pairs~\cite{Oreifej2013}  & 12 & 10 & 1 & 360 &  RGB+Depth\\
UWA3D Activity~\cite{RahmaniHOPC2014}  & 30 & 10 & 1 & 701 &  RGB+Depth\\
UWA3D Multiview Activity II~\cite{Rahmani2016}  & 30 & 9 & 4 & 1,070 &  RGB+Depth\\
HMDB51~\cite{kuehne2011hmdb}  & 51 & - & - & 6,766 & RGB\\
UCF101~\cite{Soomro2012UCF101AD}  & 101 & - & - & 13,320 &  RGB\\
NTU RGB+D~\cite{shahroudy2016ntu}& 60 & 40 & 80 & 56,880 &  RGB+Depth\\
NTU RGB+D 120~\cite{Liu_2019_NTURGBD120} & 120 & 106 & 155 & 114,480 &  RGB+Depth\\
\bottomrule
\end{tabular}}
\caption{We evaluate both RGB and depth videos across a variety of datasets, chosen to represent a wide range of action recognition challenges. These datasets vary from small-scale to large-scale, complex benchmarks, which involve long sequences and intricate temporal dynamics. They cover difficulties such as motion ambiguity, cluttered backgrounds, viewpoint variations, low resolution, and diverse action types, all requiring models that can handle complex temporal and spatial dependencies.
}
\label{tab:datasets}
\end{center}
\end{table}

\section{Experiment}
\label{sec:exp}


This section presents our experiments, and evaluations, followed by an in-depth discussion on the TIME layer. In the Appendix, we include additional experimental results, visualizations, and in-depth discussions for further insights.

\subsection{Setup}

\noindent\textbf{Models \& datasets.} In this work, we evaluate three representative models (\cref{sec:application}), covering both frame-based and video-based architectures, across two modalities: conventional RGB and depth. Our analysis spans eight diverse datasets, ranging from small-scale to large-scale, which collectively present various challenging scenarios. These include datasets capturing sport-related activities, such as \textit{jogging}, \textit{side boxing}, and \textit{golf swing} (\eg, MSRAction3D~\cite{li_msraction3d}), where distinct motion characteristics are critical. Additionally, we evaluate on pairs of actions with highly similar motion trajectories (\eg, \textit{pick up a box} and \textit{put down a box}; \textit{stick a poster} and \textit{remove a poster} in 3D Action Pairs~\cite{Oreifej2013}), testing the models' ability to discern fine-grained differences.
We further analyze performance on actions performed at various heights and speeds in cluttered scenes that feature frequent self-occlusions (UWA3D Activity~\cite{RahmaniHOPC2014}), and actions captured from multiple camera viewpoints (UWA3D Multiview Activity II~\cite{Rahmani2016}), examining the robustness to occlusion and viewpoint changes. The HMDB51 dataset~\cite{kuehne2011hmdb} introduces challenges of low-resolution videos captured in uncontrolled settings, emphasizing motion ambiguity, viewpoint variations, and differing pose orientations that are common in real-world applications. The UCF101 dataset~\cite{Soomro2012UCF101AD} further highlights issues in actions with inherent motion ambiguity, variations in viewpoint, and diverse pose configurations. Lastly, we include two large-scale RGB+D datasets, NTU RGB+D (NTU-60)~\cite{shahroudy2016ntu} and NTU RGB+D 120 (NTU-120)~\cite{Liu_2019_NTURGBD120}, containing longer sequences with complex temporal dynamics, which allow for an in-depth evaluation of the models' ability to understand and recognize actions with extended temporal structures. A summary of these datasets is in \cref{tab:datasets}.

\noindent\textbf{Implementations \& metrics.} Since our TIME layer generates new frame images by rearranging the original video frames, these frames differ from typical real images (\eg, ImageNet-1K~\cite{5206848}) in terms of both visual content and layout. The generated frames contain repeated visual patterns with some motion, and while the scale of objects and subjects may differ, the visual content resembles that of ImageNet-1K. As a result, no existing datasets with image-level labels are directly applicable for pretraining on these transformed frames. Given this, we focus on fine-tuning ImageNet-1K pretrained models for our experiments. For fine-tuning, we use ResNet-50~\cite{he2016deep} and ViT~\cite{dosovitskiy2021an} models, both pretrained on ImageNet-1K, using two V100 GPUs.

For pretraining VideoMAE on the datasets listed in \cref{tab:datasets}, we use the base model with a patch size of 16 and an input resolution of $224\times224$. Training is conducted on four V100 GPUs for all datasets. Following the original paper, we use the AdamW optimizer with a 40-epoch warm-up phase. We adjust the batch size based on the dataset, ranging from 24 to 128 (\eg, 128 for both HMDB51 and UCF101), with the total number of epochs set between 80 and 4800. For instance, NTU-60 and NTU-120 are trained for 200 epochs~\cite{woo2023towards}, UCF101 for 3200 epochs, and HMDB51 for 4800 epochs~\cite{tong2022videomae}, with the number of epochs determined by dataset size and complexity. During fine-tuning, we follow the procedure outlined in \cite{tong2022videomae}, using a base learning rate of 5e-4 for UCF101 and 1e-3 for HMDB51, with a weight decay of 0.7 and a batch size of 128 for both datasets. Fine-tuning is performed for 50 epochs on HMDB51 and 100 epochs on UCF101. For the large-scale NTU-60 and NTU-120 datasets, we perform fine-tuning for 50 epochs.

We primarily report Top-1 recognition accuracy, and in some cases, to deepen the evaluation, we also include Top-5 performance. Additionally, we analyze the effects of the TIME layer by measuring cosine similarity between model weights, layer by layer, comparing models trained with and without the TIME layer. This comparison provides insights into the influence of the TIME layer across different datasets, video modalities, mask ratios in VideoMAE, and both from-scratch training and fine-tuning setups.

\subsection{Analysis and Evaluation}

\begin{figure}[tbp]
\centering
\begin{subfigure}[b]{0.33\linewidth}
\centering\includegraphics[trim=0 0 0 0, clip=true,width=\linewidth]{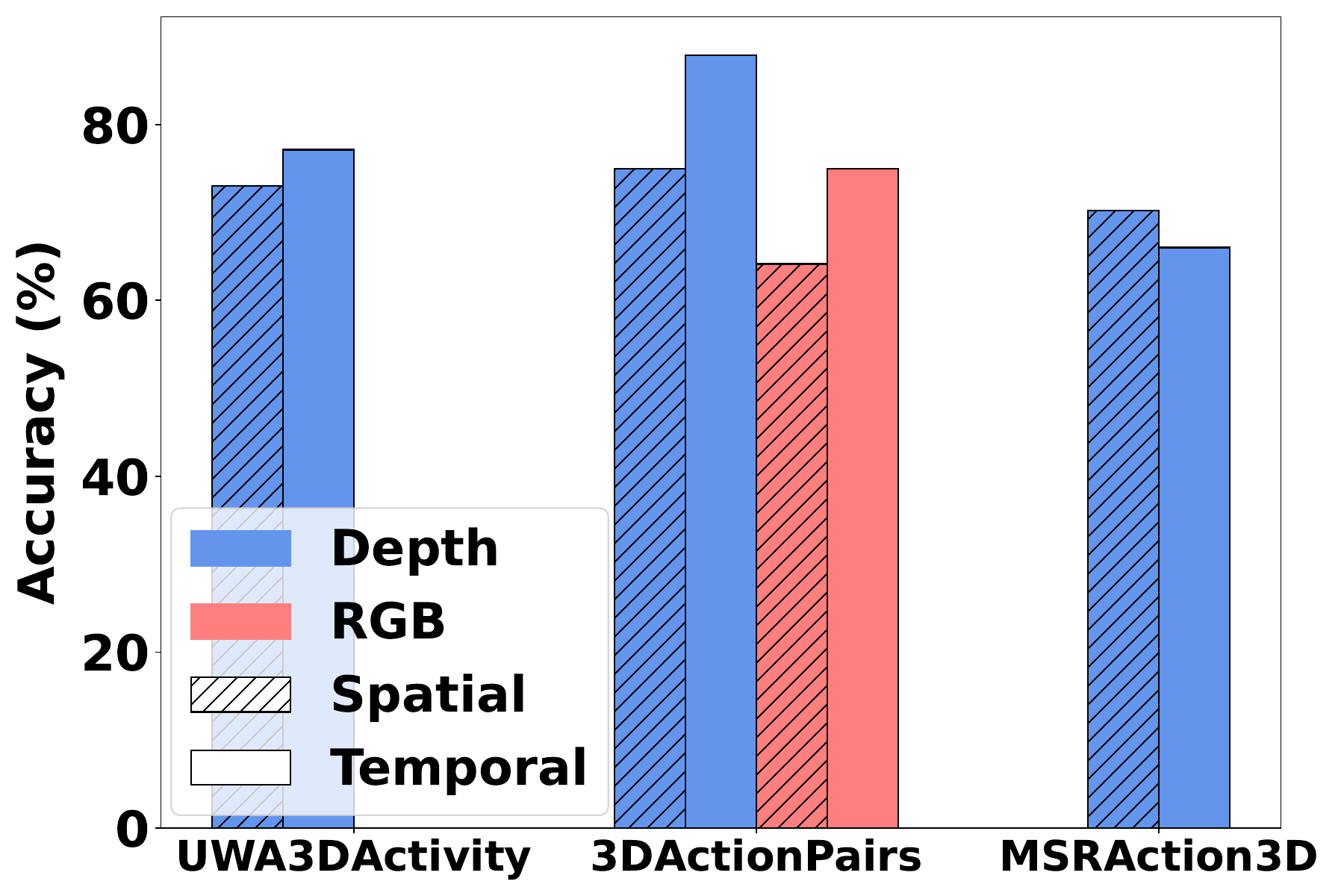}
\caption{\label{fig:resnet}ResNet-50.}
\end{subfigure}\hfill
\begin{subfigure}[b]{0.33\linewidth}
\centering\includegraphics[trim=0 0 0 0, clip=true,width=\linewidth]{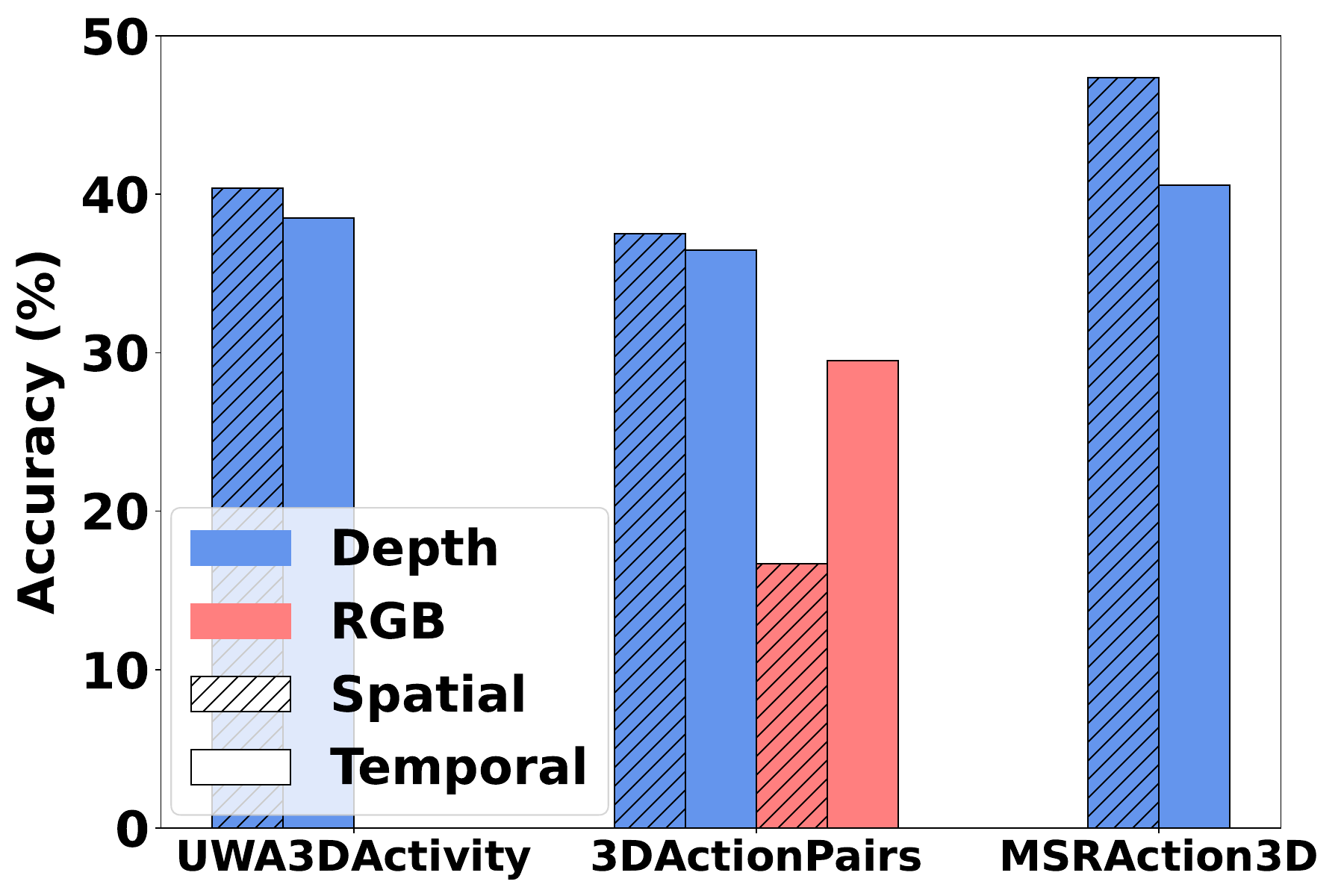}
\caption{\label{fig:vit}ViT.}
\end{subfigure}\hfill
\begin{subfigure}[b]{0.33\linewidth}
\centering\includegraphics[trim=0 0 0 0, clip=true,width=\linewidth]{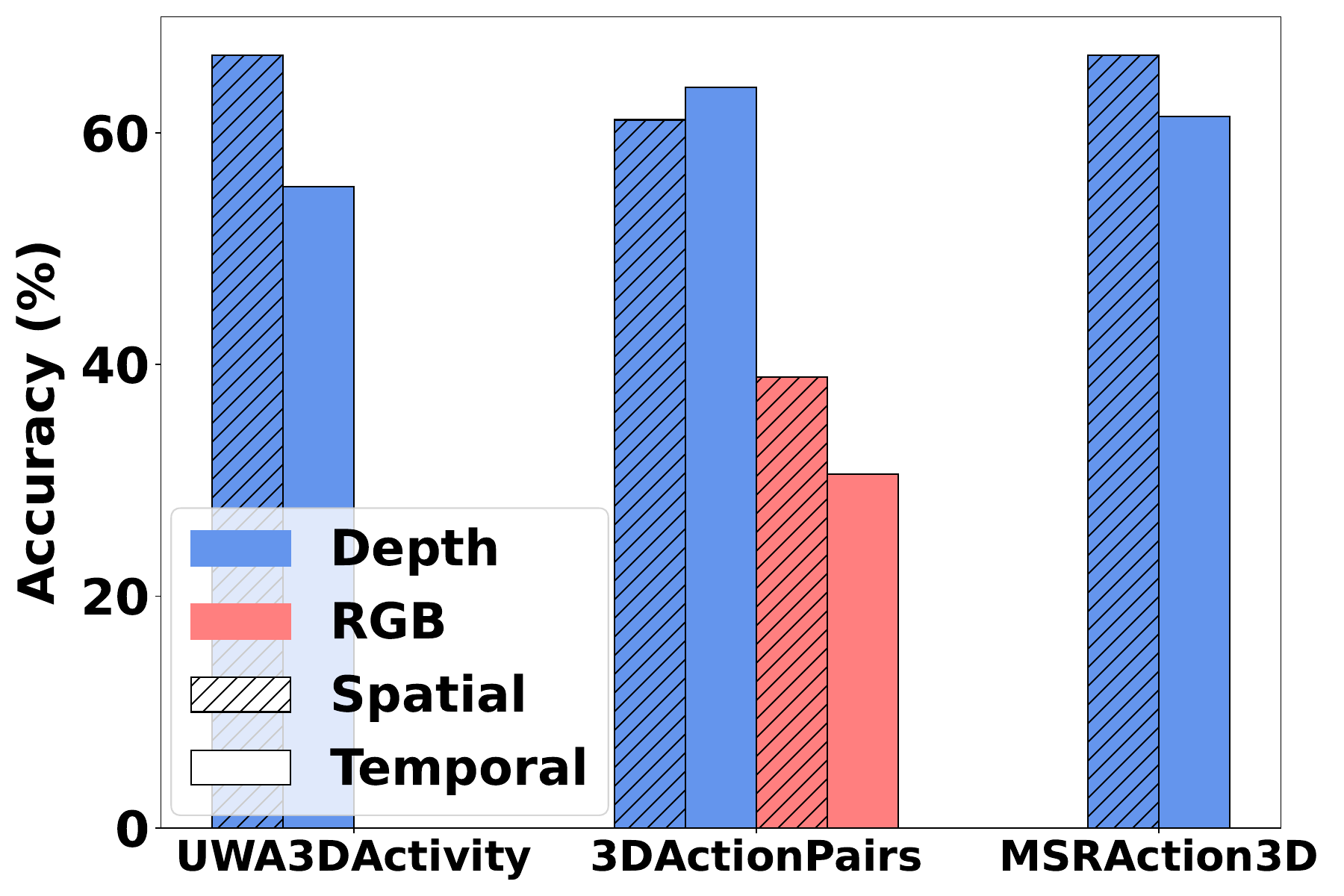}
\caption{\label{fig:videomae}VideoMAE.}
\end{subfigure}
\caption{Comparison of spatial and temporal block arrangements across RGB and depth modalities using ResNet-50 (pretrained on ImageNet-1K), ViT (pretrained on ImageNet-1K), and VideoMAE (trained from scratch) on three datasets. 
}
\label{fig:s-t}
\end{figure}

\noindent\textbf{Spatial vs. temporal block arrangements.} We conduct an ablation study on two TIME layer variants (described in~\cref{sec:time}), using ResNet-50 and ViT models fine-tuned with ImageNet-1K pretrained weights, and training VideoMAE from scratch.
Results show that ViT benefits more from spatial block arrangement on depth videos, while temporal block arrangement excels on 3D Action Pairs (RGB), likely due to the dataset’s action pairs featuring similar motion, where temporal block arrangement better captures short-term and subtle motions.
For ResNet-50, the temporal block arrangement provides better performance, except on the MSRAction3D (Depth). This can be attributed to the pretrained ResNet-50’s ability to capture fine visual details, which benefits from short-term motion captured per frame in the temporal block arrangement. With VideoMAE, spatial block arrangement performs on par or better across both modalities and datasets. Therefore, for the remaining evaluations, we use the spatial block arrangement.

\begin{figure}[tbp]
\centering
\begin{subfigure}[b]{0.495\linewidth}
\centering\includegraphics[trim=0 0 0 0, clip=true,width=\linewidth]{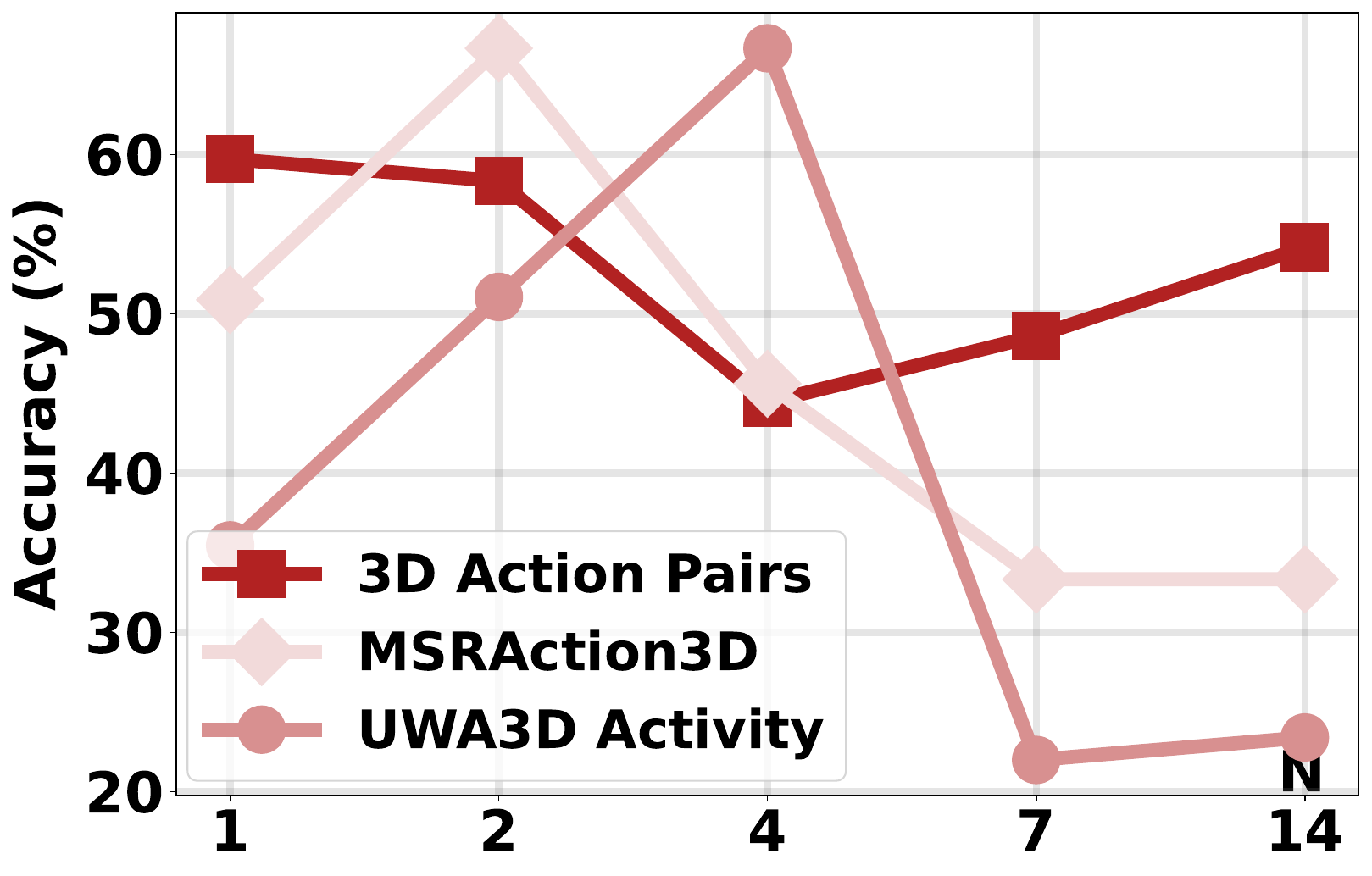}
\caption{\label{fig:trade-depth}Depth modality.}
\end{subfigure}\hfill
\begin{subfigure}[b]{0.495\linewidth}
\centering\includegraphics[trim=0 0 0 0, clip=true,width=\linewidth]{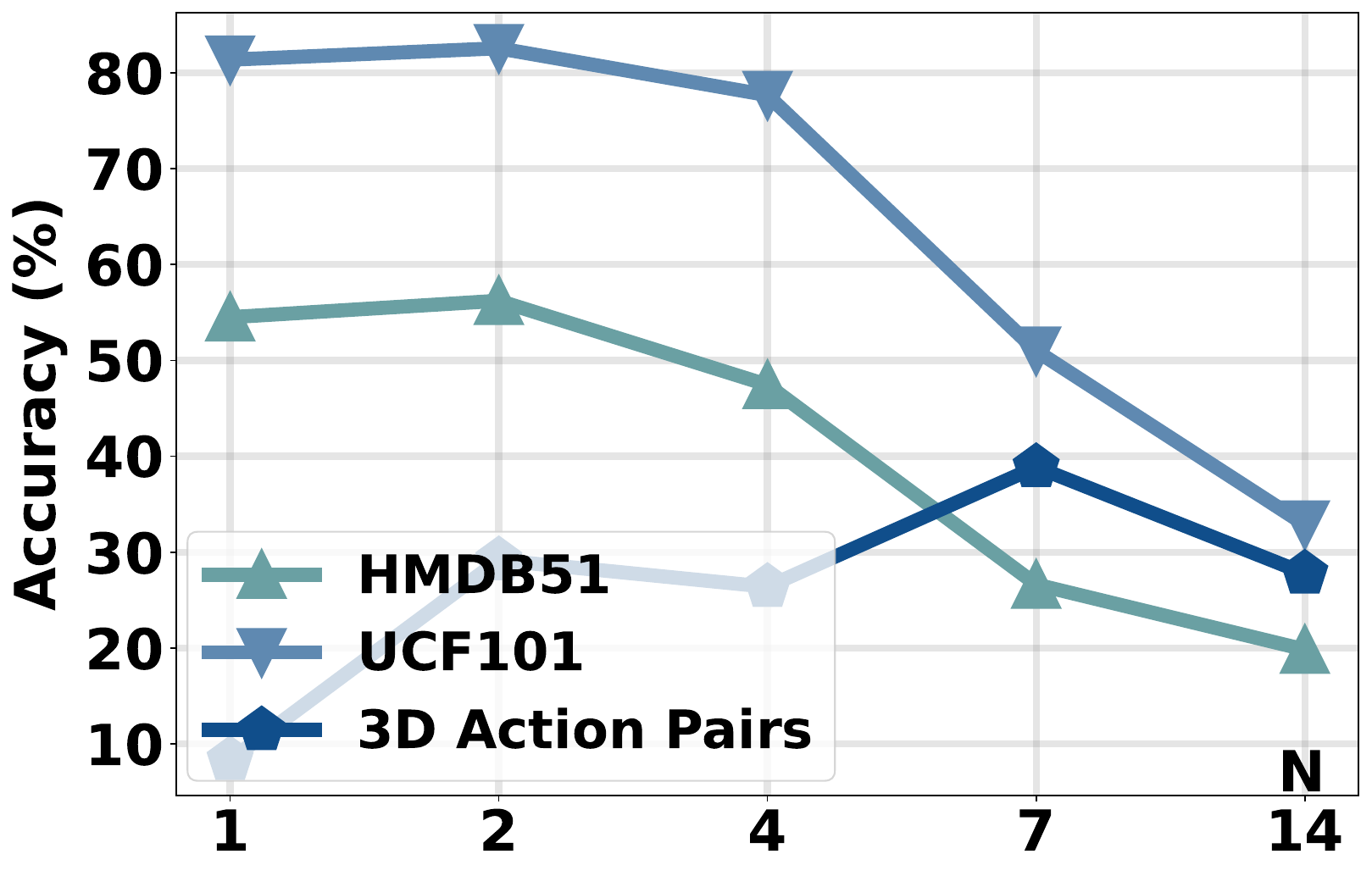}
\caption{\label{fig:trade-rgb}RGB modality.}
\end{subfigure}
\caption{Impact of the spatial-temporal balance parameter $N$ on VideoMAE. For the RGB modality, smaller values of $N$ generally yield better performance, as RGB content tends to be more visually complex. In contrast, the 3D Action Pairs dataset, containing simpler actions, achieves optimal results with a larger $N$ compared to more challenging datasets like HMDB51 and UCF101.}
\label{fig:trade-off}
\end{figure}

\noindent\textbf{$N$ as spatial-temporal balance parameter.} As shown in \cref{fig:trade-depth}, UWA3D Activity (Depth) benefits from a higher $N$ (\eg, $N\!=\!4$), likely because depth videos separate the foreground subject more clearly, providing sharper silhouettes even amidst cluttered scenes. In contrast, RGB datasets like HMDB51 and UCF101 achieve better performance with a smaller $N$ (typically $N\!=\!2$, as in~\cref{fig:trade-rgb}), where emphasizing temporal information aids in recognizing actions with complex visual environments. Increasing $N$ reduces performance, possibly due to the visual complexity and lower resolution. Notably, 3D Action Pairs (RGB), with its single, clearly defined subject and absence of significant motion ambiguity, performs better with a much higher $N$ (\eg, $N\!=\!7$), reflecting the simpler nature of its actions compared to HMDB51 and UCF101, where pose, viewpoint, and motion variability are high. These findings suggest that the optimal $N$ setting aligns with the dataset’s visual structure and complexity, with simpler or well-segmented scenes benefitting from a higher spatial-temporal balance parameter. 

\begin{figure}[tbp]
\centering
\begin{subfigure}[b]{0.495\linewidth}
\centering\includegraphics[trim=0 0 0 0, clip=true,width=\linewidth]{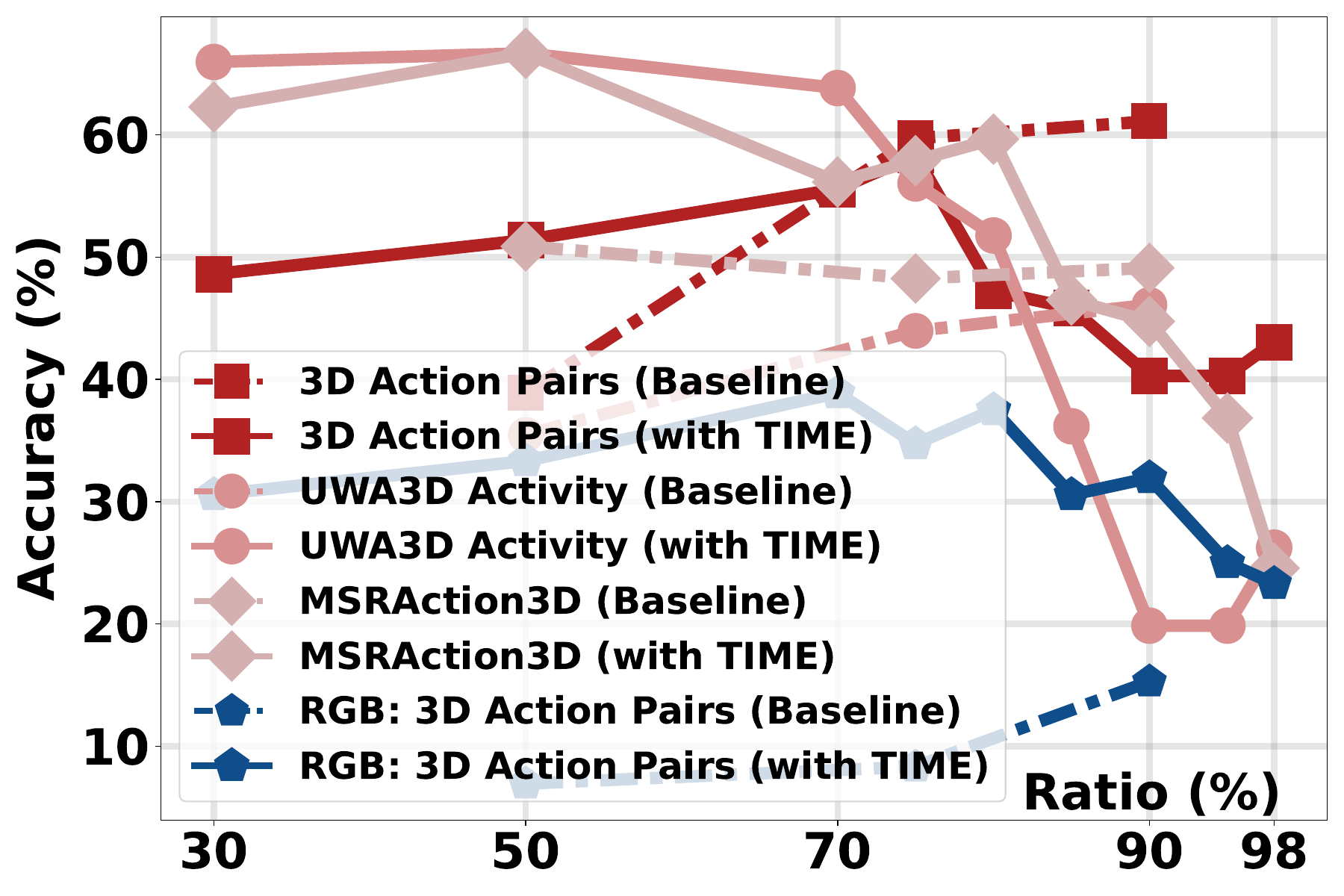}
\caption{\label{fig:mask-small}On small-scale datasets.}
\end{subfigure}\hfill
\begin{subfigure}[b]{0.495\linewidth}
\centering\includegraphics[trim=0 0 0 0, clip=true,width=\linewidth]{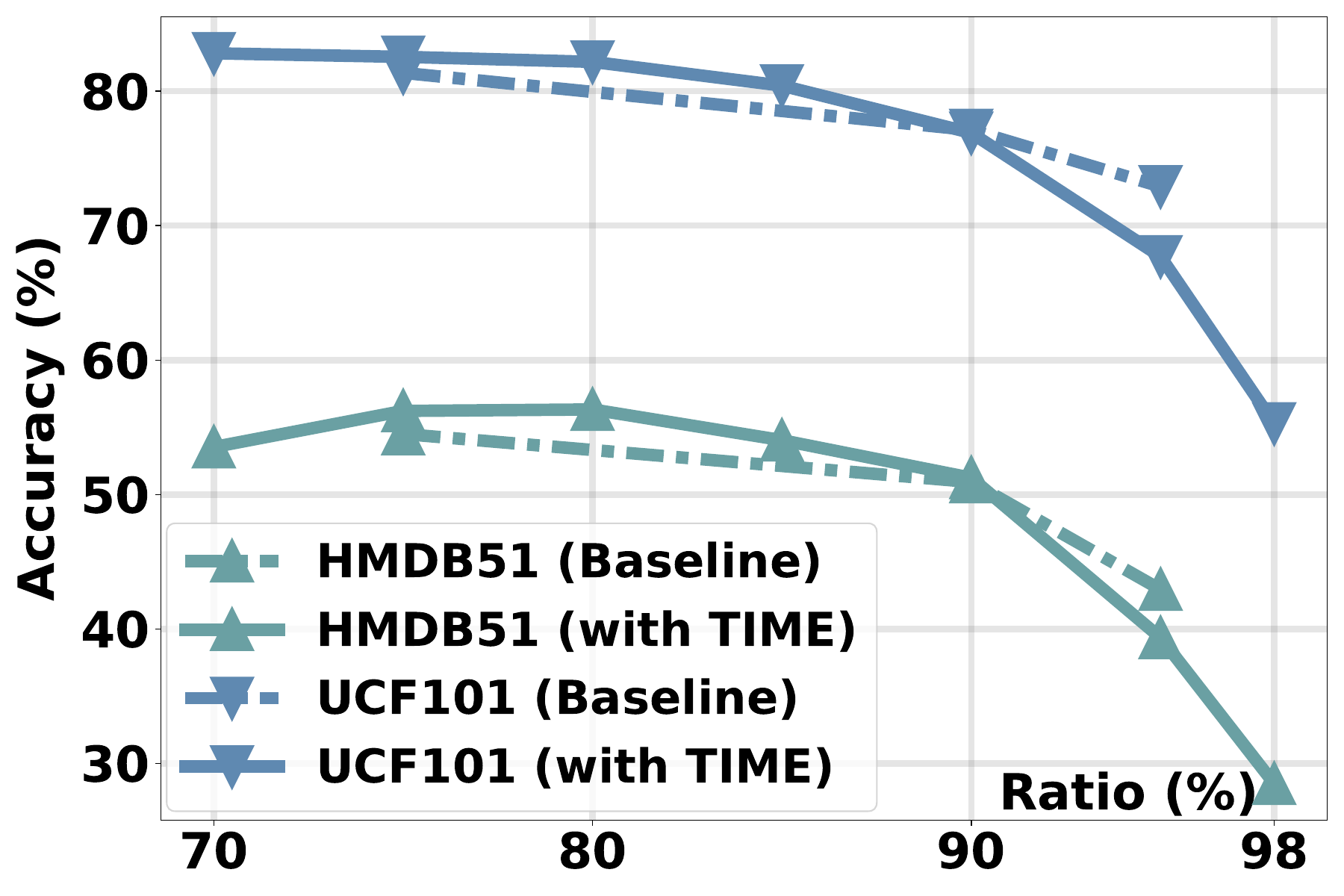}
\caption{\label{fig:mask-big} On HMDB51 and UCF101.}
\end{subfigure}
\caption{Impact of the TIME layer on VideoMAE mask ratios. We analyze RGB and depth videos across various datasets, including small-scale and challenging ones like HMDB51. Results show that the TIME layer generally favors a lower mask ratio for optimal performance on RGB videos.}
\label{fig:mask-ratio}
\end{figure}

\begin{table*}[tb]
\begin{minipage}{0.65\linewidth}
\centering
    \resizebox{\linewidth}{!}{\begin{tabular}{l l l c c c c}
    \toprule
        Backbone & Pretraining & Fine-tuning/Testing & Modality & TIME layer & Top-1 & Top-5 \\ 
        \midrule
        \multirow{12}{*}{ViT} & \multirow{12}{*}{ImageNet-1K} & \multirow{2}{*}{MSRAction3D} & \multirow{2}{*}{Depth} & $\text{\sffamily X}$ &13.16&57.90 \\ 
        & & & & $\checkmark$ ($N=7$) &\textbf{40.57} & \textbf{82.24} \\ 
        \cline{3-7} 
        & & \multirow{4}{*}{3D Action Pairs} & \multirow{2}{*}{Depth} & $\text{\sffamily X}$ &22.22&85.42 \\ 
        & &  &  &$\checkmark$ ($N=7$) &\textbf{36.46} &\textbf{97.22} \\
        \cline{4-7}
        & & & \multirow{2}{*}{RGB} & $\text{\sffamily X}$ &14.24&57.29 \\ 
        & & & & $\checkmark$ ($N=4$) &\textbf{29.51} &\textbf{94.10} \\ 
        \cline{3-7} 
        & & \multirow{2}{*}{UWA3D Activity} & \multirow{2}{*}{Depth} & $\text{\sffamily X}$ &10.99&58.16 \\
        & & &  & $\checkmark$ ($N=7$) &\textbf{38.48} &\textbf{74.29} \\ 
        \cline{3-7} 
        & & \multirow{2}{*}{HMDB51} & \multirow{2}{*}{RGB} & $\text{\sffamily X}$ &47.09&76.44 \\
        & & & & $\checkmark$ ($N=2$) &\textbf{47.29} &\textbf{77.17} \\ 
        \cline{3-7} 
        & & \multirow{2}{*}{UCF101} & \multirow{2}{*}{RGB} & $\text{\sffamily X}$&\textbf{84.26}&\textbf{95.80} \\ 
        & &  & & $\checkmark$ ($N=2$) &79.81 &95.27 \\ 
        \hline
        \multirow{12}{*}{ResNet-50} & \multirow{12}{*}{ImageNet-1K} & \multirow{2}{*}{MSRAction3D} & \multirow{2}{*}{Depth} & $\text{\sffamily X}$ &17.98&61.40 \\ 
        & & & & $\checkmark$ ($N=7$) &\textbf{66.01} &\textbf{94.96} \\
        \cline{3-7} 
        & & \multirow{4}{*}{3D Action Pairs} & \multirow{2}{*}{Depth} & $\text{\sffamily X}$ &42.36 &98.97 \\
        & & & & $\checkmark$ ($N=14$) &\textbf{87.85} &\textbf{100.00} \\
        \cline{4-7} 
        & & & \multirow{2}{*}{RGB} & $\text{\sffamily X}$ &47.57 &\textbf{100.00} \\
        & & & &  $\checkmark$ ($N=4$) & \textbf{75.00} & \textbf{100.00} \\ 
        \cline{3-7} 
        & & \multirow{2}{*}{UWA3D Activity} & \multirow{2}{*}{Depth} & $\text{\sffamily X}$ &64.01&93.62 \\
        & &  &  & $\checkmark$ ($N=4$) &\textbf{77.13} &\textbf{97.52} \\ 
        \cline{3-7}  
        & & \multirow{2}{*}{HMDB51} & \multirow{2}{*}{RGB} & $\text{\sffamily X}$ &\textbf{45.05}&\textbf{76.54} \\
         & & & & $\checkmark$ ($N=2$) &42.83 &73.17 \\ 
        \cline{3-7}  
        & & \multirow{2}{*}{UCF101} & \multirow{2}{*}{RGB} & $\text{\sffamily X}$ &\textbf{76.55} &\textbf{93.79} \\
        & & & & $\checkmark$ ($N=2$) &71.27 &92.02 \\ 
        \hline
        \multirow{22}{*}{VideoMAE} & \multirow{2}{*}{Kinetics-400}  & \multirow{2}{*}{MSRAction3D} & \multirow{2}{*}{Depth} & $\text{\sffamily X}$ &70.18&\textbf{98.25}\\
        &  & & & $\checkmark$ ($N=2$) &\textbf{73.68}&97.37 \\
        \cline{2-7} 
         & \multirow{2}{*}{MSRAction3D}  & \multirow{2}{*}{MSRAction3D} & \multirow{2}{*}{Depth} & $\text{\sffamily X}$ &50.88&87.72\\
        &  & & & $\checkmark$ ($N=2$) &\textbf{66.67}&\textbf{93.86} \\
        \cline{2-7} 
        & \multirow{4}{*}{Kinetics-400} & \multirow{4}{*}{3D Action Pairs} & \multirow{2}{*}{Depth} & $\text{\sffamily X}$ &79.17&\textbf{100.00} \\
        &  &  &  & $\checkmark$ ($N=2$) &\textbf{82.34} &\textbf{100.00} \\ 
        \cline{4-7} 
        & &  & \multirow{2}{*}{RGB} & $\text{\sffamily X}$ &\textbf{72.22} &\textbf{100.00} \\ 
        & & & & $\checkmark$ ($N=2$) & 66.88 &97.98 \\ 
        \cline{2-7} 
        & \multirow{4}{*}{3D Action Pairs} & \multirow{4}{*}{3D Action Pairs} & \multirow{2}{*}{Depth} & $\text{\sffamily X}$ &\textbf{61.11}&\textbf{100.00} \\
        &  &  &  & $\checkmark$ ($N=2$) &58.33 &\textbf{100.00} \\
        \cline{4-7} 
        &  & & \multirow{2}{*}{RGB} & $\text{\sffamily X}$ &\textbf{38.89} &\textbf{100.00} \\ 
        & & & & $\checkmark$ ($N=7$) &\textbf{38.89} &94.44 \\
        \cline{2-7}
        & \multirow{4}{*}{Kinetics-400}  & \multirow{4}{*}{UWA3D Activity} & \multirow{2}{*}{RGB} & $\text{\sffamily X}$ & \textbf{88.80} & \textbf{100.00} \\
        & &  & & $\checkmark$ ($N=2$) & 84.38 & \textbf{100.00} \\
        \cline{4-7} 
        & &  & \multirow{2}{*}{Depth} & $\text{\sffamily X}$ & \textbf{81.56} & \textbf{99.29} \\
        & &  & & $\checkmark$ ($N=2$) & 81.25 & 97.92 \\
        \cline{2-7} 
        & \multirow{2}{*}{UWA3D Activity}  & \multirow{2}{*}{UWA3D Activity} & \multirow{2}{*}{Depth} & $\text{\sffamily X}$ &46.10 &91.49 \\
        & &  & & $\checkmark$ ($N=4$) &\textbf{66.67} & \textbf{96.45} \\
        \cline{2-7} 
        & \multirow{2}{*}{HMDB51}  & \multirow{2}{*}{HMDB51} & \multirow{2}{*}{RGB} &  $\text{\sffamily X}$ &54.51 &82.97 \\ 
        & & & & $\checkmark$ ($N=2$) &\textbf{56.33} &\textbf{82.99} \\
        \cline{2-7} 
        & \multirow{2}{*}{UCF101}  & \multirow{2}{*}{UCF101} & \multirow{2}{*}{RGB} & $\text{\sffamily X}$ &81.37 &96.30 \\ 
        & & & & $\checkmark$ ($N=2$) &\textbf{82.56} &\textbf{97.36} \\
        \bottomrule
    \end{tabular}}
    \caption{Evaluation of three representative backbones on conventional RGB and depth videos. For VideoMAE, we perform pretraining and finetuning on various datasets, while the other two backbones are finetuned from ImageNet-1K pretrained models. The evaluation is conducted with and without the TIME layer, indicated by ($\checkmark$) and ($\text{\sffamily X}$), respectively. Values in parentheses represent the optimal spatial-temporal balance parameters $N$, as defined in \cref{sec:time}. For depth videos, larger values of $N$ tend to improve performance, likely due to better foreground segmentation and the lack of color information, which emphasizes object and human silhouettes. Additionally, training from scratch with the TIME layer generally outperforms the baselines.}
    \label{tab:main}
\end{minipage}\hfill
\begin{minipage}{0.32\linewidth}
\centering
\rotatebox{-90}{\includegraphics[width=2.75\linewidth]{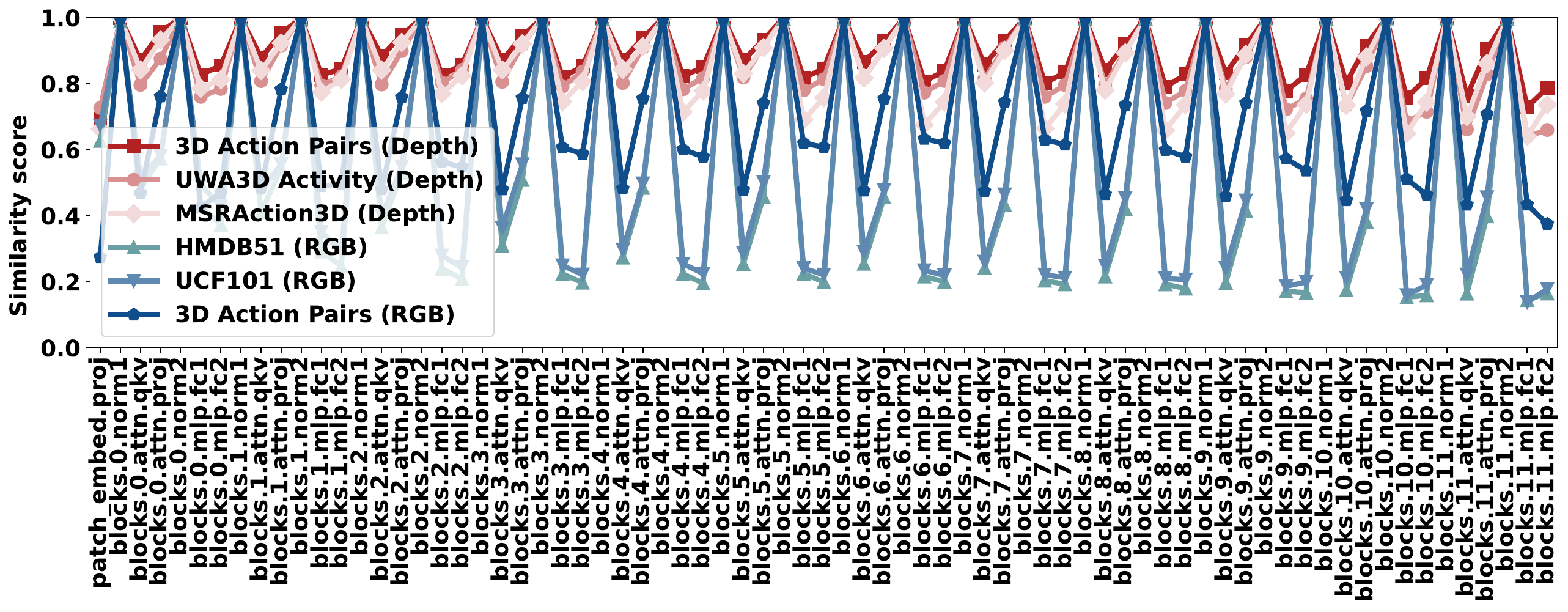}}
  \captionof{figure}{Per-layer weight similarity comparison between VideoMAE encoders with and without the TIME layer. The models are first trained from scratch, followed by fine-tuning of the encoders. The horizontal axis shows cosine similarity scores (ranging from 0 to 1), where higher values indicate greater similarity between the weights of the models. The vertical axis lists the individual layers within the VideoMAE encoder.}
  \label{fig:temp-eval}
\end{minipage}
\end{table*}

\noindent\textbf{Impact on VideoMAE mask ratios.} As shown in~\cref{fig:mask-small}, smaller datasets in the RGB modality generally perform best with higher mask ratios (\eg, 70–80\% for 3D Action Pairs), while depth modalities tend to favor lower mask ratios, \eg, 50\% for MSRAction3D and UWA3D Activity. For more challenging datasets (as shown in~\cref{fig:mask-big}), an optimal mask ratio of approximately 80\% is observed for HMDB51 and UCF101. Increasing the mask ratio beyond this point leads to a dramatic decrease in performance. This shows that the TIME layer's integration of temporal information benefits from a lower mask ratio, which retains sufficient visual details to enhance feature extraction effectively.

\noindent\textbf{TIME layer in frame- and video-level models.} As shown in~\cref{tab:main}, with the TIME layer, VideoMAE generally outperforms frame-level models such as ViT and ResNet-50. This is because VideoMAE is more effective at processing video data through mechanisms like tube masking and reconstructing missing spatiotemporal cues. 
With the addition of the TIME layer, VideoMAE further enhances its ability to extract spatial-temporal information, resulting in consistent performance improvements, particularly for training-from-scratch setups. When fine-tuned with the TIME layer on small datasets, the ImageNet-1K pre-trained models (ViT and ResNet-50) also demonstrate strong performance. This demonstrates that the TIME layer enhances frame-level models for video tasks by effectively injecting temporal information, striking a balance between spatial and temporal features.
Interestingly, we also observe that ResNet-50, as a frame-level model with the TIME layer, achieves competitive results compared to VideoMAE. This may be due to a spatial bias in existing methods, as noted in~\cite{patrick2021keeping,girdhar2023omnimae}, where a single frame can sometimes be sufficient to recognize actions in Kinetics-400 without requiring the full sequence. These findings underscore the need for datasets that demand more robust temporal reasoning to fully evaluate video-level understanding.

\noindent\textbf{TIME layer’s effect across modalities.} As shown in~\cref{tab:main}, the depth modality consistently outperforms conventional RGB across nearly all backbones. Depth videos enable easier foreground segmentation of human subjects, even in cluttered scenes, since the absence of color eliminates distractions like clothing color. This allows action recognition models to focus on extracting high-level features that describe the action, rather than handling low-level segmentation tasks.
Additionally, both ImageNet-1K and Kinetics-400 pretrained models perform effectively when fine-tuned on depth videos, often surpassing their performance on RGB video fine-tuning. This suggests that the inherent clarity and reduced noise in depth data can positively impact model performance, underscoring the role of data quality in action recognition.

\noindent\textbf{TIME layer on the effect of dataset volume.} As shown in~\cref{tab:main}, we notice that large-scale pretraining, \eg, on Kinetics-400 and then finetuning on smaller datasets generally achieve better performance compared to the training from scratch using smaller datasets. This is mainly because smaller datasets have limited motion concepts, visual appearances and scenarios, whereas large-scale datasets provide more rich information. Even for the depth videos, we observe the same trend. We also notice that ResNet-50 pretrained on ImageNet-1K, achieve quite competitive results on all the datasets and modalities.

\begin{table*}[tbp]
\begin{center}
\resizebox{0.9\textwidth}{!}{\begin{tabular}{l l c  c  c  c  c c  c  c  c  c c  c  c}
\toprule
& Training & \multicolumn{2}{c}{$V_1$ \& $V_2$} & \multicolumn{2}{c}{$V_1$ \& $V_3$} & \multicolumn{2}{c}{$V_1$ \& $V_4$} & \multicolumn{2}{c}{$V_2$ \& $V_3$} & \multicolumn{2}{c}{$V_2$ \& $V_4$} & \multicolumn{2}{c}{$V_3$ \& $V_4$} & \multirow{2}{*}{Average}\\
\cline{2-14}
& Testing & $V_3$ & $V_4$ & $V_2$ & $V_4$ & $V_2$ & $V_3$ & $V_1$ & $V_4$ & $V_1$ & $V_3$ & $V_1$ & $V_2$ & {}\\
\midrule
\multirow{3}{*}{Depth} & w/o K400 tuning & 70.52 & 62.17 & 67.67 & 63.67 & 58.27 & 63.43 & 64.31 & 57.30 & 63.94 & 66.42 & 69.89 & 67.29 & 64.73 \\
& w/ K400 tuning & 79.48 & 76.40 & 75.19 & 76.40 & 70.30 & 74.25 & 76.95 & 70.41 & 78.44 & 76.49 & 76.21 & 73.31 & 75.06\\
& + TIME layer & \textbf{89.18} & \textbf{85.02} & \textbf{87.55} & \textbf{83.52} & \textbf{78.57} & \textbf{79.48} & \textbf{86.99} & \textbf{77.15} & \textbf{89.22} & \textbf{88.81} & \textbf{89.96} & \textbf{86.47} & \textbf{85.16}\\
\hline
\multirow{2}{*}{RGB} & w/o K400 tuning & 84.76 & 84.39 & 88.06 & 84.39 & 76.12 & 77.32 & 89.96 & 84.39 & 88.48 & 82.53 & 91.08 & 86.94 &  84.59\\
& + TIME layer & \textbf{91.45} & \textbf{89.96} & \textbf{91.04} & \textbf{89.59} & \textbf{82.09} & \textbf{85.50} & \textbf{91.08} & \textbf{89.22} & \textbf{88.48} & \textbf{87.73} & \textbf{92.19} & \textbf{90.67} & \textbf{89.08}\\
\bottomrule
\end{tabular}}
\caption{Evaluations on UWA3D Multiview Activity II show that the TIME layer consistently enhances performance in VideoMAE, both with and without Kinetics-400 (K400) fine-tuning, across RGB and depth modalities in all 12 training and testing viewpoint combinations.}
\label{tab:uwa3dmulti}
\end{center}
\end{table*}

\noindent\textbf{TIME layer on multiview action recognition.}
\cref{tab:uwa3dmulti} shows results on UWA3D Multiview Activity II using VideoMAE pretrained on Kinetics-400 and fine-tuned, with and without the TIME layer, on both RGB and depth modalities. Due to the smaller size of this dataset, we focus solely on fine-tuning rather than training from scratch. We find that incorporating the TIME layer consistently improves accuracy across all 12 distinct pairs of training and testing viewpoints, yielding gains of around 10\% for depth videos and 5\% for RGB videos, highlighting the TIME layer’s effectiveness in handling viewpoint variations and motion details. Notably, the TIME layer enables fine-tuning of the depth modality even when starting from RGB-pretrained models, demonstrating its versatility across modalities.


\begin{figure*}[tbp] 
\centering
\includegraphics[trim=1.2cm 2.2cm 0.2cm 2.2cm, clip=true, width=0.9\linewidth]{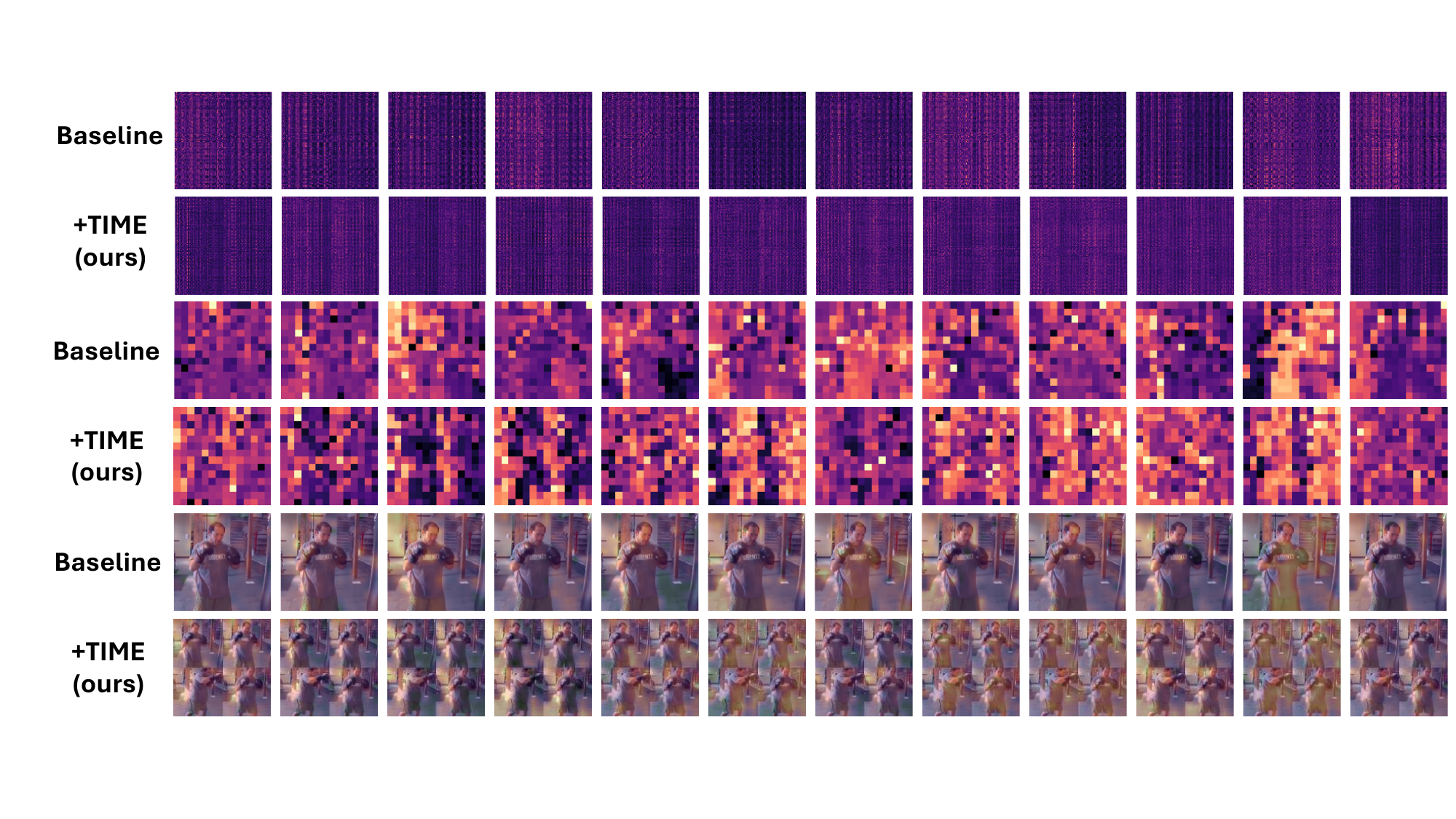}
\caption{Visualization of attention maps, class weight heatmaps, and overlaid attention maps on VideoMAE with and without the TIME layer (Baseline vs. +TIME). The first two rows show attention maps from all 12 layers of the VideoMAE encoder, capturing the spatial distribution of attention across video frames and highlighting areas of focus at each layer. The third and fourth rows depict class weight heatmaps, derived from averaging attention scores of the [CLS] token across a $14\! \times \!14$ grid of spatial patches. These heatmaps indicate which patches are most relevant for classification, with the TIME layer (+TIME) producing sharper, more distinct attention regions compared to the Baseline. The last two rows overlay class weight heatmaps onto original video frames, showing that the TIME layer enables a more focused attention on critical regions, such as central subjects and areas of motion, while the Baseline displays a more diffused focus. This demonstrates how the TIME layer enhances the model's capacity to capture key features relevant for action classification.}
\label{fig:attentionmap}
\end{figure*}

\noindent\textbf{Using TIME layer as a diagnostic tool.} We use cosine similarity to compare the per-layer weights of VideoMAE encoders between finetuned models that are firstly trained from scratch, with and without the TIME layer, across RGB and depth modalities. Notably, for all RGB modality datasets, we observe that the similarity scores for attention projection layers are smaller. This suggests that the TIME layer influences the weights of fully connected layers responsible for embedding temporal information, enriching video processing tasks with nuanced motion cues.

In contrast, for depth datasets, the similarity scores tend to be higher compared to RGB datasets. Although the TIME layer still affects the attention projection layers, this result indicates that depth videos offer cleaner and more distinct silhouette information, which may reduce the impact of temporal adjustments.
Our findings suggest that the TIME layer can serve as a valuable tool to assess model behavior and identify challenging datasets. For example, datasets like HMDB51 and UCF101 show much lower similarity scores for attention layers, indicating potential difficulties in processing complex actions with high viewpoint, pose, and motion variability. This insight can help fine-tune model performance and inform dataset selection. 

\cref{fig:attentionmap} provides a visual comparison of attention maps generated with and without the use of the TIME layer. The TIME layer enhances attention maps by sharpening focus on key regions, especially areas with significant motion and central subjects, which are often critical for action recognition. This indicates that the TIME layer effectively directs the model's attention to relevant features, improving spatial precision and temporal sensitivity.

\section{Conclusion}
\label{sec:concl}
We introduced the TIME layer, a module designed to enhance temporal dynamics in video action recognition models with minimal architectural adjustments. By providing a flexible spatial-temporal balance, the TIME layer enables both short-term and long-term temporal dynamics, resulting in improved model performance on complex action recognition tasks. Results show that the TIME layer effectively enriches temporal cues while preserving spatial fidelity, outperforming traditional methods. Its adaptability across diverse models, including ResNet-50, ViT, and VideoMAE, positions the TIME layer as a scalable and powerful solution for advancing future video processing research.

\section*{Acknowledgments}
Huilin Chen conducted this research under the supervision of Lei Wang for her final year honors research project at ANU.  
This work was supported by the National Computational Merit Allocation Scheme 2024 (NCMAS 2024), with computational resources provided by NCI Australia, an NCRIS-enabled capability supported by the Australian Government.

{
    \small
    \bibliographystyle{ieeenat_fullname}
    \bibliography{main}

\begin{thebibliography}{41}
\providecommand{\natexlab}[1]{#1}
\providecommand{\url}[1]{\texttt{#1}}
\expandafter\ifx\csname urlstyle\endcsname\relax
  \providecommand{\doi}[1]{doi: #1}\else
  \providecommand{\doi}{doi: \begingroup \urlstyle{rm}\Url}\fi

\bibitem[Arnab et~al.(2021)Arnab, Dehghani, Heigold, Sun, Lu{\v{c}}i{\'c}, and Schmid]{arnab2021vivit}
Anurag Arnab, Mostafa Dehghani, Georg Heigold, Chen Sun, Mario Lu{\v{c}}i{\'c}, and Cordelia Schmid.
\newblock Vivit: A video vision transformer.
\newblock In \emph{Proceedings of the IEEE/CVF international conference on computer vision}, pages 6836--6846, 2021.

\bibitem[Bertasius et~al.(2021)Bertasius, Wang, and Torresani]{bertasius2021space}
Gedas Bertasius, Heng Wang, and Lorenzo Torresani.
\newblock Is space-time attention all you need for video understanding?
\newblock In \emph{ICML}, page~4, 2021.

\bibitem[Bilen et~al.(2016)Bilen, Fernando, Gavves, Vedaldi, and Gould]{Bilen_2016_CVPR}
Hakan Bilen, Basura Fernando, Efstratios Gavves, Andrea Vedaldi, and Stephen Gould.
\newblock Dynamic image networks for action recognition.
\newblock In \emph{CVPR}, 2016.

\bibitem[Carreira and Zisserman(2017)]{carreira2017quo}
Joao Carreira and Andrew Zisserman.
\newblock Quo vadis, action recognition? a new model and the kinetics dataset.
\newblock In \emph{proceedings of the IEEE Conference on Computer Vision and Pattern Recognition}, pages 6299--6308, 2017.

\bibitem[Chen et~al.(2024)Chen, Wang, Koniusz, and Gedeon]{chen2024motion}
Qixiang Chen, Lei Wang, Piotr Koniusz, and Tom Gedeon.
\newblock Motion meets attention: Video motion prompts.
\newblock In \emph{The 16th Asian Conference on Machine Learning (Conference Track)}, 2024.

\bibitem[Dai et~al.(2017)Dai, Singh, Zhang, Davis, and Qiu~Chen]{dai2017temporal}
Xiyang Dai, Bharat Singh, Guyue Zhang, Larry~S Davis, and Yan Qiu~Chen.
\newblock Temporal context network for activity localization in videos.
\newblock In \emph{Proceedings of the IEEE International Conference on Computer Vision}, pages 5793--5802, 2017.

\bibitem[Deng et~al.(2009)Deng, Dong, Socher, Li, Li, and Fei-Fei]{5206848}
Jia Deng, Wei Dong, Richard Socher, Li-Jia Li, Kai Li, and Li Fei-Fei.
\newblock Imagenet: A large-scale hierarchical image database.
\newblock In \emph{2009 IEEE Conference on Computer Vision and Pattern Recognition}, pages 248--255, 2009.

\bibitem[Dosovitskiy et~al.(2021)Dosovitskiy, Beyer, Kolesnikov, Weissenborn, Zhai, Unterthiner, Dehghani, Minderer, Heigold, Gelly, Uszkoreit, and Houlsby]{dosovitskiy2021an}
Alexey Dosovitskiy, Lucas Beyer, Alexander Kolesnikov, Dirk Weissenborn, Xiaohua Zhai, Thomas Unterthiner, Mostafa Dehghani, Matthias Minderer, Georg Heigold, Sylvain Gelly, Jakob Uszkoreit, and Neil Houlsby.
\newblock An image is worth 16x16 words: Transformers for image recognition at scale.
\newblock In \emph{International Conference on Learning Representations}, 2021.

\bibitem[Feichtenhofer(2020)]{feichtenhofer2020x3d}
Christoph Feichtenhofer.
\newblock X3d: Expanding architectures for efficient video recognition.
\newblock In \emph{Proceedings of the IEEE/CVF conference on computer vision and pattern recognition}, pages 203--213, 2020.

\bibitem[Feichtenhofer et~al.(2016)Feichtenhofer, Pinz, and Zisserman]{feichtenhofer2016convolutional}
Christoph Feichtenhofer, Axel Pinz, and Andrew Zisserman.
\newblock Convolutional two-stream network fusion for video action recognition.
\newblock In \emph{Proceedings of the IEEE conference on computer vision and pattern recognition}, pages 1933--1941, 2016.

\bibitem[Feichtenhofer et~al.(2019)Feichtenhofer, Fan, Malik, and He]{feichtenhofer2019slowfast}
Christoph Feichtenhofer, Haoqi Fan, Jitendra Malik, and Kaiming He.
\newblock Slowfast networks for video recognition.
\newblock In \emph{ICCV}, pages 6202--6211, 2019.

\bibitem[Girdhar et~al.(2023)Girdhar, El-Nouby, Singh, Alwala, Joulin, and Misra]{girdhar2023omnimae}
Rohit Girdhar, Alaaeldin El-Nouby, Mannat Singh, Kalyan~Vasudev Alwala, Armand Joulin, and Ishan Misra.
\newblock Omnimae: Single model masked pretraining on images and videos.
\newblock In \emph{Proceedings of the IEEE/CVF conference on computer vision and pattern recognition}, pages 10406--10417, 2023.

\bibitem[He et~al.(2016)He, Zhang, Ren, and Sun]{he2016deep}
Kaiming He, Xiangyu Zhang, Shaoqing Ren, and Jian Sun.
\newblock Deep residual learning for image recognition.
\newblock In \emph{Proceedings of the IEEE conference on computer vision and pattern recognition}, pages 770--778, 2016.

\bibitem[Kim et~al.(2022)Kim, Gowda, Aodha, and Sevilla-Lara]{kim2022capturing}
Kiyoon Kim, Shreyank~N Gowda, Oisin~Mac Aodha, and Laura Sevilla-Lara.
\newblock Capturing temporal information in a single frame: Channel sampling strategies for action recognition.
\newblock In \emph{BMVC}. {BMVA} Press, 2022.

\bibitem[Koniusz et~al.(2021)Koniusz, Wang, and Cherian]{koniusz2021tensor}
Piotr Koniusz, Lei Wang, and Anoop Cherian.
\newblock Tensor representations for action recognition.
\newblock \emph{IEEE Transactions on Pattern Analysis and Machine Intelligence}, 44\penalty0 (2):\penalty0 648--665, 2021.

\bibitem[Kuehne et~al.(2011)Kuehne, Jhuang, Garrote, Poggio, and Serre]{kuehne2011hmdb}
Hildegard Kuehne, Hueihan Jhuang, Est{\'\i}baliz Garrote, Tomaso Poggio, and Thomas Serre.
\newblock {HMDB}: {A} large video database for human motion recognition.
\newblock In \emph{ICCV}, pages 2556--2563, 2011.

\bibitem[Li et~al.(2010)Li, Zhang, and Liu]{li_msraction3d}
Wanqing Li, Zhengyou Zhang, and Zicheng Liu.
\newblock {Action recognition based on a bag of 3D points}.
\newblock \emph{CVPR Workshop}, pages 9--14, 2010.

\bibitem[Lin et~al.(2019)Lin, Gan, and Han]{lin2019tsm}
Ji Lin, Chuang Gan, and Song Han.
\newblock Tsm: Temporal shift module for efficient video understanding.
\newblock In \emph{Proceedings of the IEEE/CVF international conference on computer vision}, pages 7083--7093, 2019.

\bibitem[Liu et~al.(2019)Liu, Shahroudy, Perez, Wang, Duan, and Kot]{Liu_2019_NTURGBD120}
Jun Liu, Amir Shahroudy, Mauricio Perez, Gang Wang, Ling-Yu Duan, and Alex~C. Kot.
\newblock Ntu rgb+d 120: A large-scale benchmark for 3d human activity understanding.
\newblock \emph{IEEE Transactions on Pattern Analysis and Machine Intelligence}, 2019.

\bibitem[Oreifej and Liu(2013)]{Oreifej2013}
Omar Oreifej and Zicheng Liu.
\newblock {HON4D: Histogram of Oriented 4D Normals for Activity Recognition from Depth Sequences}.
\newblock In \emph{CVPR}, pages 716--723, 2013.

\bibitem[Patrick et~al.(2021)Patrick, Campbell, Asano, Misra, Metze, Feichtenhofer, Vedaldi, and Henriques]{patrick2021keeping}
Mandela Patrick, Dylan Campbell, Yuki Asano, Ishan Misra, Florian Metze, Christoph Feichtenhofer, Andrea Vedaldi, and Joao~F Henriques.
\newblock Keeping your eye on the ball: Trajectory attention in video transformers.
\newblock \emph{Advances in neural information processing systems}, 34:\penalty0 12493--12506, 2021.

\bibitem[Rahmani et~al.(2014)Rahmani, Mahmood, Huynh, and Mian]{RahmaniHOPC2014}
Hossein Rahmani, Arif Mahmood, Du~Q. Huynh, and Ajmal Mian.
\newblock {HOPC: Histogram of Oriented Principal Components of 3D Pointclouds for Action Recognition}.
\newblock In \emph{ECCV}, pages 742--757, 2014.

\bibitem[Rahmani et~al.(2016)Rahmani, Mahmood, Huynh, and Mian]{Rahmani2016}
Hossein Rahmani, Arif Mahmood, Du~Q. Huynh, and Ajmal Mian.
\newblock {Histogram of Oriented Principal Components for Cross-View Action Recognition}.
\newblock \emph{TPAMI}, pages 2430--2443, 2016.

\bibitem[Ryoo et~al.(2020)Ryoo, Piergiovanni, Tan, and Angelova]{Ryoo2020AssembleNet}
Michael~S. Ryoo, AJ Piergiovanni, Mingxing Tan, and Anelia Angelova.
\newblock Assemblenet: Searching for multi-stream neural connectivity in video architectures.
\newblock In \emph{International Conference on Learning Representations}, 2020.

\bibitem[Shahroudy et~al.(2016)Shahroudy, Liu, Ng, and Wang]{shahroudy2016ntu}
Amir Shahroudy, Jun Liu, Tian-Tsong Ng, and Gang Wang.
\newblock Ntu rgb+ d: A large scale dataset for 3d human activity analysis.
\newblock \emph{CVPR}, pages 1010--1019, 2016.

\bibitem[Simonyan and Zisserman(2014)]{simonyan2014two}
Karen Simonyan and Andrew Zisserman.
\newblock Two-stream convolutional networks for action recognition in videos.
\newblock \emph{Advances in neural information processing systems}, 27, 2014.

\bibitem[Soomro et~al.(2012)Soomro, Zamir, and Shah]{Soomro2012UCF101AD}
Khurram Soomro, Amir Zamir, and Mubarak Shah.
\newblock Ucf101: A dataset of 101 human actions classes from videos in the wild.
\newblock \emph{ArXiv}, abs/1212.0402, 2012.

\bibitem[Tong et~al.(2022)Tong, Song, Wang, and Wang]{tong2022videomae}
Zhan Tong, Yibing Song, Jue Wang, and Limin Wang.
\newblock Videomae: Masked autoencoders are data-efficient learners for self-supervised video pre-training.
\newblock \emph{Advances in neural information processing systems}, 35:\penalty0 10078--10093, 2022.

\bibitem[Tran et~al.(2015)Tran, Bourdev, Fergus, Torresani, and Paluri]{tran2015learning}
Du Tran, Lubomir Bourdev, Rob Fergus, Lorenzo Torresani, and Manohar Paluri.
\newblock Learning spatiotemporal features with 3d convolutional networks.
\newblock In \emph{Proceedings of the IEEE international conference on computer vision}, pages 4489--4497, 2015.

\bibitem[Wang and Koniusz(2021)]{wang2021self}
Lei Wang and Piotr Koniusz.
\newblock Self-supervising action recognition by statistical moment and subspace descriptors.
\newblock In \emph{Proceedings of the 29th ACM international conference on multimedia}, pages 4324--4333, 2021.

\bibitem[Wang and Koniusz(2024)]{wang2024flow}
Lei Wang and Piotr Koniusz.
\newblock Flow dynamics correction for action recognition.
\newblock In \emph{ICASSP 2024-2024 IEEE International Conference on Acoustics, Speech and Signal Processing (ICASSP)}, pages 3795--3799. IEEE, 2024.

\bibitem[Wang et~al.(2016)Wang, Xiong, Wang, Qiao, Lin, Tang, and Van~Gool]{wang2016temporal}
Limin Wang, Yuanjun Xiong, Zhe Wang, Yu Qiao, Dahua Lin, Xiaoou Tang, and Luc Van~Gool.
\newblock Temporal segment networks: Towards good practices for deep action recognition.
\newblock In \emph{European conference on computer vision}, pages 20--36. Springer, 2016.

\bibitem[Wang et~al.(2019{\natexlab{a}})Wang, Koniusz, and Huynh]{wang2019hallucinating}
Lei Wang, Piotr Koniusz, and Du~Q Huynh.
\newblock Hallucinating idt descriptors and i3d optical flow features for action recognition with cnns.
\newblock In \emph{Proceedings of the IEEE/CVF international conference on computer vision}, pages 8698--8708, 2019{\natexlab{a}}.

\bibitem[Wang et~al.(2019{\natexlab{b}})Wang, Xiong, Wang, Qiao, Lin, Tang, and Van~Gool]{8454294}
Limin Wang, Yuanjun Xiong, Zhe Wang, Yu Qiao, Dahua Lin, Xiaoou Tang, and Luc Van~Gool.
\newblock Temporal segment networks for action recognition in videos.
\newblock \emph{IEEE Transactions on Pattern Analysis and Machine Intelligence}, 41\penalty0 (11):\penalty0 2740--2755, 2019{\natexlab{b}}.

\bibitem[Wang et~al.(2023)Wang, Huang, Zhao, Tong, He, Wang, Wang, and Qiao]{wang2023videomae}
Limin Wang, Bingkun Huang, Zhiyu Zhao, Zhan Tong, Yinan He, Yi Wang, Yali Wang, and Yu Qiao.
\newblock Videomae v2: Scaling video masked autoencoders with dual masking.
\newblock In \emph{Proceedings of the IEEE/CVF Conference on Computer Vision and Pattern Recognition}, pages 14549--14560, 2023.

\bibitem[Wang et~al.(2024{\natexlab{a}})Wang, Sun, and Koniusz]{wang2024high}
Lei Wang, Ke Sun, and Piotr Koniusz.
\newblock High-order tensor pooling with attention for action recognition.
\newblock In \emph{ICASSP 2024-2024 IEEE International Conference on Acoustics, Speech and Signal Processing (ICASSP)}, pages 3885--3889. IEEE, 2024{\natexlab{a}}.

\bibitem[Wang et~al.(2024{\natexlab{b}})Wang, Yuan, Gedeon, and Zheng]{wangtaylor}
Lei Wang, Xiuyuan Yuan, Tom Gedeon, and Liang Zheng.
\newblock Taylor videos for action recognition.
\newblock In \emph{Forty-first International Conference on Machine Learning}, 2024{\natexlab{b}}.

\bibitem[Wang et~al.(2024{\natexlab{c}})Wang, Li, Li, Yu, He, Chen, Pei, Zheng, Xu, Wang, et~al.]{wang2024internvideo2}
Yi Wang, Kunchang Li, Xinhao Li, Jiashuo Yu, Yinan He, Guo Chen, Baoqi Pei, Rongkun Zheng, Jilan Xu, Zun Wang, et~al.
\newblock Internvideo2: Scaling video foundation models for multimodal video understanding.
\newblock \emph{ECCV}, 2024{\natexlab{c}}.

\bibitem[Woo et~al.(2023)Woo, Lee, Park, Nugroho, and Kim]{woo2023towards}
Sangmin Woo, Sumin Lee, Yeonju Park, Muhammad~Adi Nugroho, and Changick Kim.
\newblock Towards good practices for missing modality robust action recognition.
\newblock In \emph{Proceedings of the AAAI Conference on Artificial Intelligence}, pages 2776--2784, 2023.

\bibitem[Yang et~al.(2020)Yang, Xu, Shi, Dai, and Zhou]{yang2020temporal}
Ceyuan Yang, Yinghao Xu, Jianping Shi, Bo Dai, and Bolei Zhou.
\newblock Temporal pyramid network for action recognition.
\newblock In \emph{Proceedings of the IEEE/CVF conference on computer vision and pattern recognition}, pages 591--600, 2020.

\bibitem[Zhu et~al.(2024)Zhu, Wang, Raj, Gedeon, and Chen]{msad2024}
Liyun Zhu, Lei Wang, Arjun Raj, Tom Gedeon, and Chen Chen.
\newblock Advancing video anomaly detection: A concise review and a new dataset.
\newblock In \emph{The Thirty-eighth Conference on Neural Information Processing Systems Datasets and Benchmarks Track}, 2024.

\end{thebibliography}
}

\clearpage
\appendix

\clearpage
\setcounter{page}{1}
\maketitlesupplementary

\section{Evaluations on UCF101, NTU-60 and NTU-120}
\label{appendix:ntu}

\begin{table}[tbp]
\setlength{\tabcolsep}{0.15em}
\renewcommand{\arraystretch}{0.70}
\fontsize{9}{9}\selectfont
    \centering
    \resizebox{\linewidth}{!}{
    \begin{tabular}{l l l c c c}
    \toprule
    Backbone & Pretraining & Fine-tuning/Testing & Modality & TIME layer & Top-1 \\ 
    \midrule
        \multirow{10}{*}{VideoMAE} & \multirow{2}{*}{UCF101}  & \multirow{2}{*}{UCF101} & \multirow{2}{*}{RGB} & $\text{\sffamily X}$ &88.3\\
        &  & & & $\checkmark$ ($N=2$) &\textbf{89.4}\\
        \cline{2-6} 
        & \multirow{4}{*}{NTU-60} & \multirow{4}{*}{NTU-60} & \multirow{2}{*}{Depth} & $\text{\sffamily X}$ & 91.7 \\
        &  &  &  & $\checkmark$ ($N=2$) &\textbf{95.0}\\ 
        \cline{4-6} 
        & &  & \multirow{2}{*}{RGB} & $\text{\sffamily X}$ & 85.3 \\ 
        & & & & $\checkmark$ ($N=2$) & \textbf{87.2}\\ 
        \cline{2-6} 
        & \multirow{4}{*}{NTU-120} & \multirow{4}{*}{NTU-120} & \multirow{2}{*}{Depth} & $\text{\sffamily X}$ &88.0\\
        &  &  &  & $\checkmark$ ($N=2$) & \textbf{92.5} \\
        \cline{4-6} 
        &  & & \multirow{2}{*}{RGB} & $\text{\sffamily X}$ & 84.5 \\ 
        & & & & $\checkmark$ ($N=2$) & \textbf{85.3}\\
    \hline
    \end{tabular}}
    \caption{Evaluation on the UCF101 and large-scale NTU-60 and NTU-120 datasets. VideoMAE is trained from scratch and fine-tuned for the action recognition task. The evaluation compares performance with and without the TIME layer, denoted by ($\checkmark$) and ($\text{\sffamily X}$), respectively. On UCF101, we train VideoMAE from scratch for 3200 epochs, followed by fine-tuning for an additional 100 epochs.}
    \label{tab:ntu}
\end{table}

Below, we present the evaluations conducted on the UCF101 and large-scale NTU-60 and NTU-120 datasets. 

As shown in~\cref{tab:ntu}, incorporating the TIME layer enhances performance across both the RGB and depth modalities on NTU-60 and NTU-120. With the TIME layer, we achieve a performance boost of approximately 4\% on both NTU-60 and NTU-120 datasets compared to the baseline, for the depth modality.

On UCF101, we observe that pre-training VideoMAE for 3200 epochs consistently enhances performance, compared to the 800-epoch pre-training used in~\cref{tab:main} of the main paper. Notably, the inclusion of the TIME layer further amplifies these improvements.

\section{Additional Results on UCF101}
\label{appendix:ucf101}

\begin{table}[tbp]
\centering
\label{tab:comparison_ucf_temporal}
\resizebox{0.45\linewidth}{!}{\begin{tabular}{c c c c c}
\toprule
\textbf{Mask ratio} & \textbf{Top-1} & \textbf{Top-5} \\ 
\midrule
70\%  &89.40 & 98.39\\ 
75\%  & 88.77 & 98.28 \\ 
80\%  & 87.58 & 98.20 \\ 
85\% & 85.99 & 97.73 \\ 
90\% & 84.17 & 97.15 \\ 
95\% & 75.63 & 93.50 \\ 
98\%  & 65.66 & 88.61 \\ 
\bottomrule
\end{tabular}}
\caption{Evaluation of the mask ratio on UCF101.}
\label{tab:ucf-maskratio}
\end{table}

\cref{tab:ucf-maskratio} presents the evaluation of mask ratios on UCF101. Unlike the standard VideoMAE, incorporating the TIME layer achieves better performance with lower mask ratios, such as 70\% and 75\%.

\section{Additional Results on UWA3D Multiview Activity II}
\label{appendix:uwa3dmulti}

\begin{table*}[tbp]
\begin{center}
\resizebox{0.9\textwidth}{!}{\begin{tabular}{l l c  c  c  c  c c  c  c  c  c c  c  c}
\toprule
& Training & \multicolumn{2}{c}{$V_1$ \& $V_2$} & \multicolumn{2}{c}{$V_1$ \& $V_3$} & \multicolumn{2}{c}{$V_1$ \& $V_4$} & \multicolumn{2}{c}{$V_2$ \& $V_3$} & \multicolumn{2}{c}{$V_2$ \& $V_4$} & \multicolumn{2}{c}{$V_3$ \& $V_4$} & \multirow{2}{*}{Average}\\
\cline{2-14}
& Testing & $V_3$ & $V_4$ & $V_2$ & $V_4$ & $V_2$ & $V_3$ & $V_1$ & $V_4$ & $V_1$ & $V_3$ & $V_1$ & $V_2$ & {}\\
\midrule
\multirow{4}{*}{Depth} & w/o K400 tuning & 70.52 & 62.17 & 67.67 & 63.67 & 58.27 & 63.43 & 64.31 & 57.30 & 63.94 & 66.42 & 69.89 & 67.29 & 64.73 \\
& + TIME layer & \textbf{84.04} & \textbf{76.18} & \textbf{83.15} & \textbf{79.26} & \textbf{76.69} & \textbf{77.74} & \textbf{80.16} & \textbf{72.70} & \textbf{84.60} & \textbf{82.50} & \textbf{84.76} & \textbf{80.59} & \textbf{80.20}\\
& w/ K400 tuning & 79.48 & 76.40 & 75.19 & 76.40 & 70.30 & 74.25 & 76.95 & 70.41 & 78.44 & 76.49 & 76.21 & 73.31 & 75.06\\
& + TIME layer & \textbf{89.18} & \textbf{85.02} & \textbf{87.55} & \textbf{83.52} & \textbf{78.57} & \textbf{79.48} & \textbf{86.99} & \textbf{77.15} & \textbf{89.22} & \textbf{88.81} & \textbf{89.96} & \textbf{86.47} & \textbf{85.16}\\
\bottomrule
\end{tabular}}
\caption{Evaluations on UWA3D Multiview Activity II show that the TIME layer consistently enhances performance in VideoMAE, both with and without Kinetics-400 (K400) fine-tuning, in all 12 training and testing viewpoint combinations.}
\label{tab:uwa3dmulti_add}
\end{center}
\end{table*}

As shown in~\cref{tab:uwa3dmulti_add}, for depth videos, the inclusion of the TIME layer significantly enhances performance. Specifically, it achieves improvements of over 15\% without Kinetics-400 pre-training and over 10\% with pre-training. These results demonstrate the TIME layer's ability to substantially boost action recognition performance.

\section{Per-layer Weight Similarity Comparison}
\label{appendix:similarity}

\begin{figure*}[tbp] 
\centering
\includegraphics[width=0.9\linewidth]{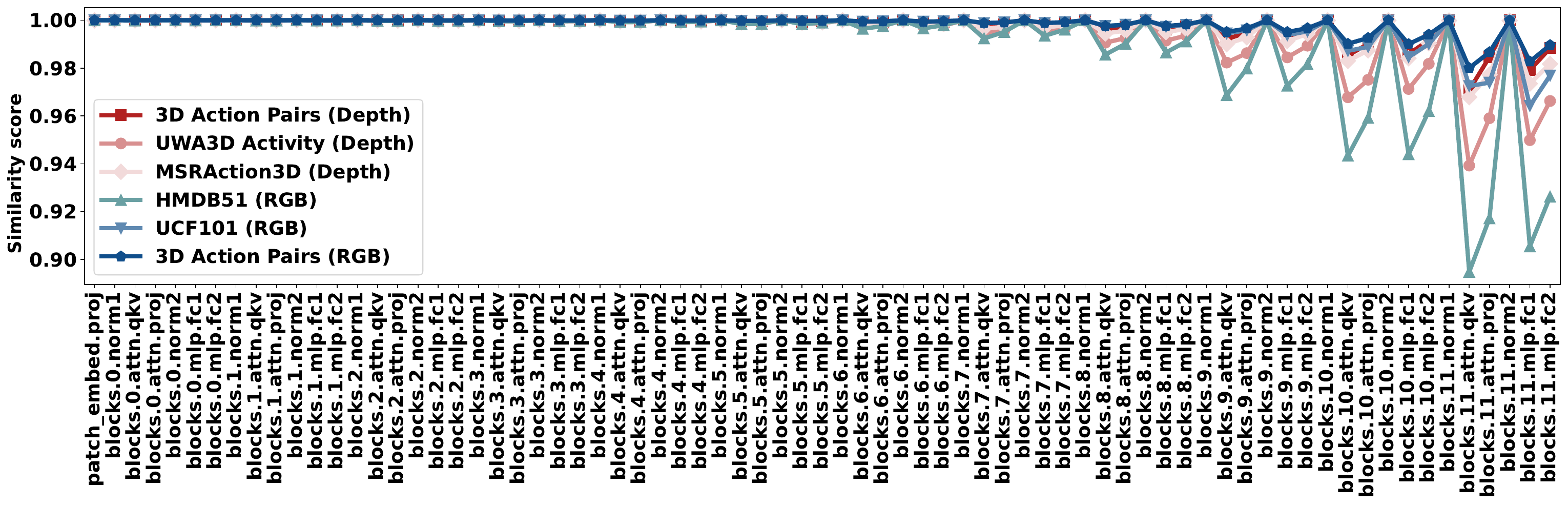}
\caption{Comparison of per-layer weight similarity between the baseline model (without the TIME layer) and the model with the TIME layer, using VideoMAE pretrained on Kinetics-400 and fine-tuned across RGB and depth modalities on various datasets.}
\label{fig:weightsimilarity_k400}
\end{figure*}

\begin{figure*}[tbp] 
\centering
\includegraphics[width=0.9\linewidth]{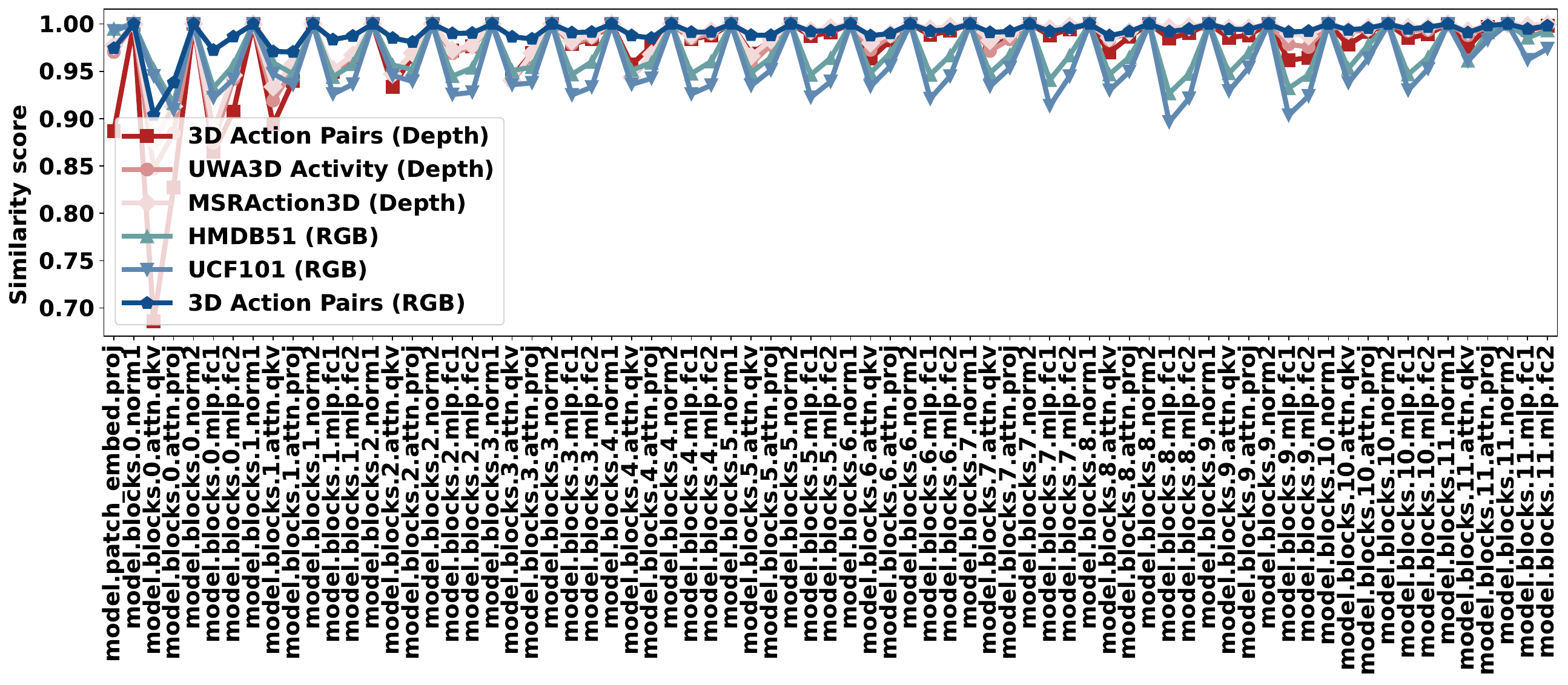}
\caption{Comparison of per-layer weight similarity between the baseline model (without the TIME layer) and the model with the TIME layer, using ViT pretrained on ImageNet-1K and fine-tuned across RGB and depth modalities on various datasets.}
\label{fig:weightsimilarity_vit}
\end{figure*}

\begin{figure*}[tbp] 
\centering
\includegraphics[width=0.9\linewidth]{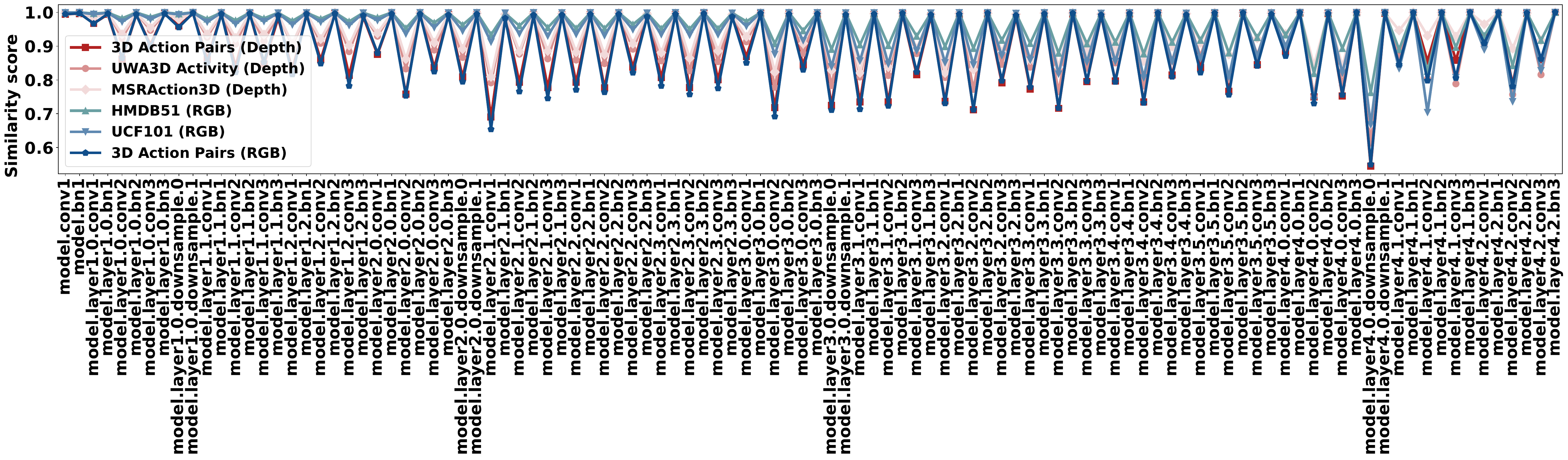}
\caption{Comparison of per-layer weight similarity between the baseline model (without the TIME layer) and the model with the TIME layer, using ResNet-50 pretrained on ImageNet-1K and fine-tuned across RGB and depth modalities on various datasets.}
\label{fig:weightsimilarity_resnet}
\end{figure*}

Below, we present a detailed comparison of per-layer weight similarity between the baseline model (without the TIME layer) and the model with the TIME layer, using VideoMAE, ViT, and ResNet-50 pretrained on either Kinetics-400 or ImageNet-1K. This comparison is performed across both RGB and depth modalities on various datasets.

We observe that fine-tuning with our TIME layer, particularly on VideoMAE (\cref{fig:weightsimilarity_k400}), significantly influences the weights of the later encoder layers. This suggests that the TIME layer is effectively extracting higher-level, abstract features, likely capturing temporal information essential for action recognition tasks. The adaptation of these later layers to incorporate action-related features highlights the impact of our TIME layer in enhancing the model's ability to capture and use temporal dynamics.

Compared to \cref{fig:temp-eval} in the main paper, we observe that training from scratch has a more pronounced effect on the layer weights. This highlights the importance of incorporating the TIME layer during large-scale pretraining, as it has the potential to further enhance downstream motion-related tasks, such as action recognition and anomaly detection.

Interestingly, we observe that fine-tuning ViT with the TIME layer leads to significant changes in the weights of the earlier layers (\cref{fig:weightsimilarity_vit}), particularly the attention and projection layers. This suggests that ViT has a limited capacity to effectively capture and process temporal information.

Additionally, we observe that the TIME layer, when fine-tuning the ResNet-50 pretrained models, significantly influences multiple layers across both RGB and depth modalities (\cref{fig:weightsimilarity_resnet}). This indicates that the TIME layer enables a 2D model to begin extracting spatial information, which is later used as temporal information for motion-related tasks. This provides a novel perspective on incorporating temporal information into traditional 2D CNNs, which have predominantly been used for image classification tasks. With the integration of the TIME layer, these models can now be adapted to enhance video processing tasks, such as action recognition.

\section{Additional Results on Spatial vs. Temporal Arrangements}
\label{appendix:arrangement}

\begin{figure*}[tbp] 
\centering
\includegraphics[width=0.9\linewidth]{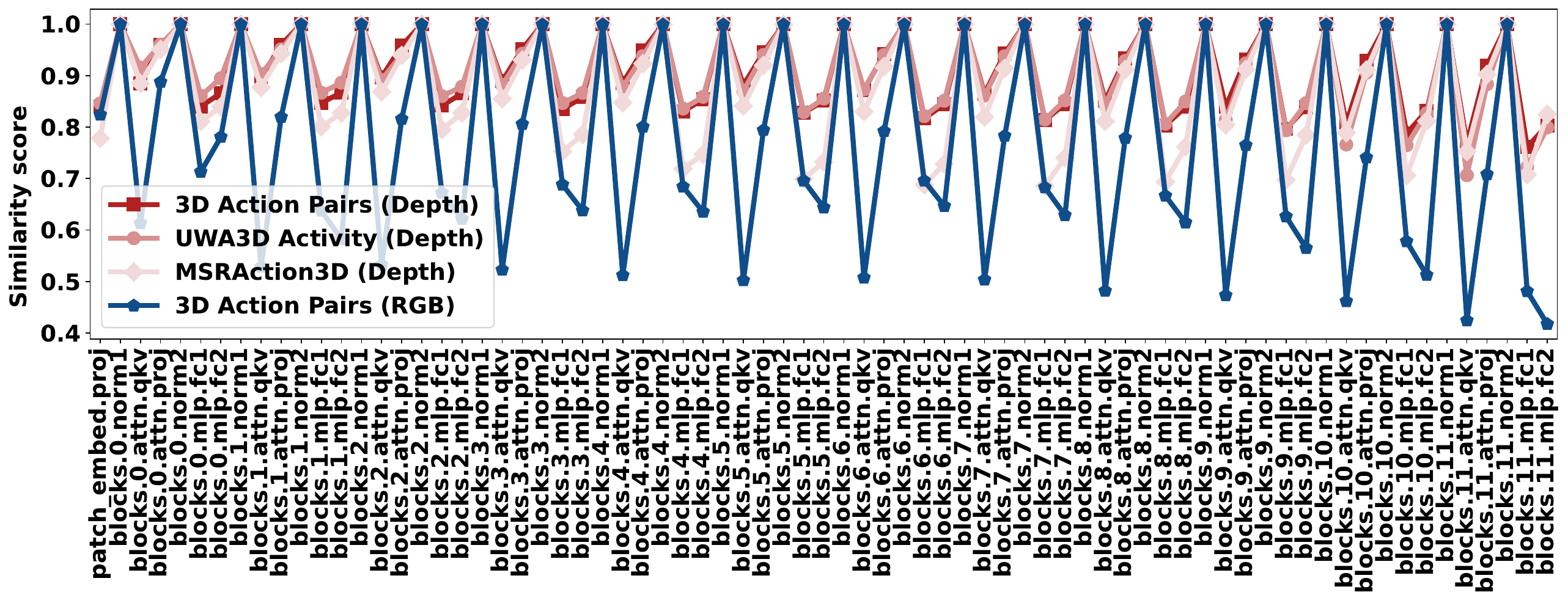}
\caption{Comparison of per-layer weight similarity between spatial and temporal block arrangements using VideoMAE trained from scratch on three depth datasets (3D Action Pairs, UWA3D Activity, MSRAction3D) and one RGB dataset (3D Action Pairs).}
\label{fig:weightsimilarity_videomae_spatialtemporal}
\end{figure*}

\begin{figure*}[tbp] 
\centering
\includegraphics[width=0.9\linewidth]{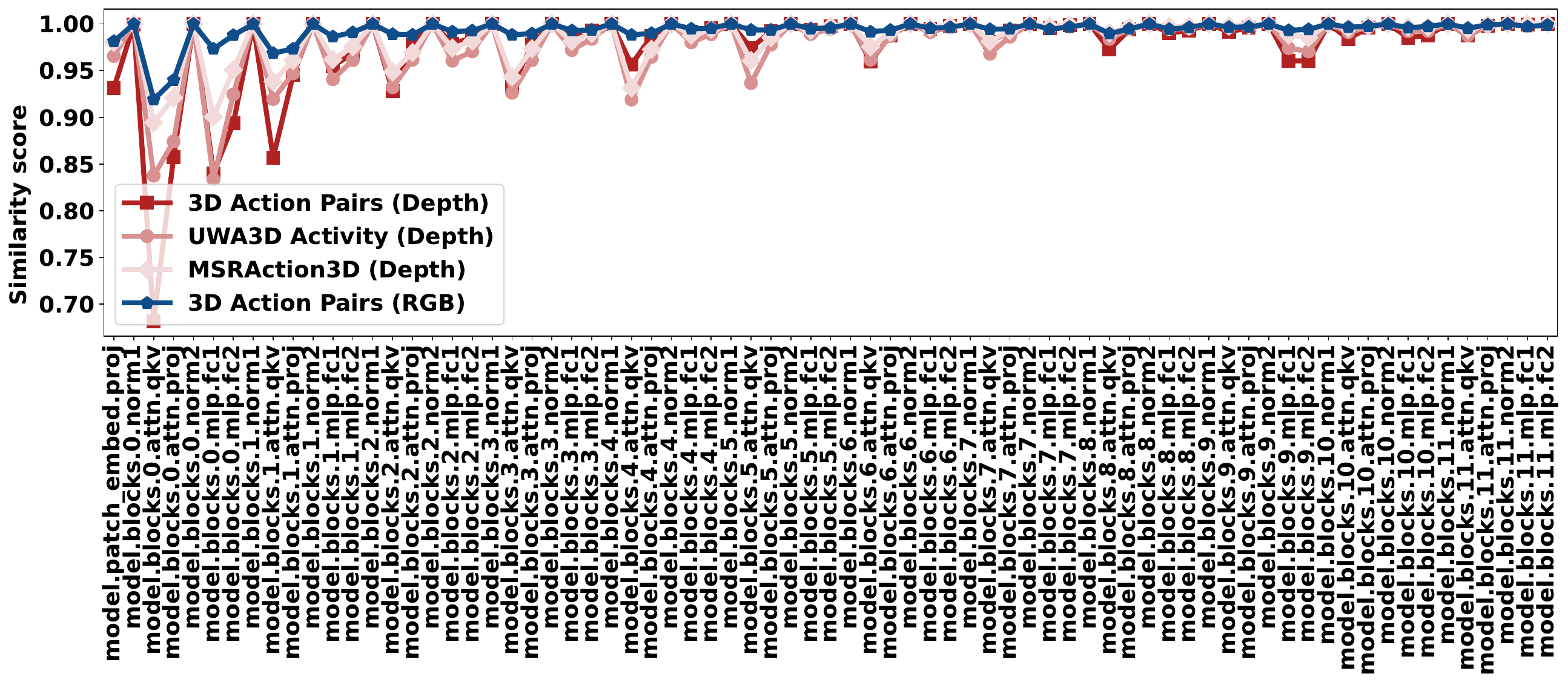}
\caption{Comparison of per-layer weight similarity between spatial and temporal block arrangements using ViT pretrained on ImageNet-1K and fine-tuned on three depth datasets (3D Action Pairs, UWA3D Activity, MSRAction3D) and one RGB dataset (3D Action Pairs).}
\label{fig:weightsimilarity_vit_spatialtemporal}
\end{figure*}

\begin{figure*}[tbp] 
\centering
\includegraphics[width=0.9\linewidth]{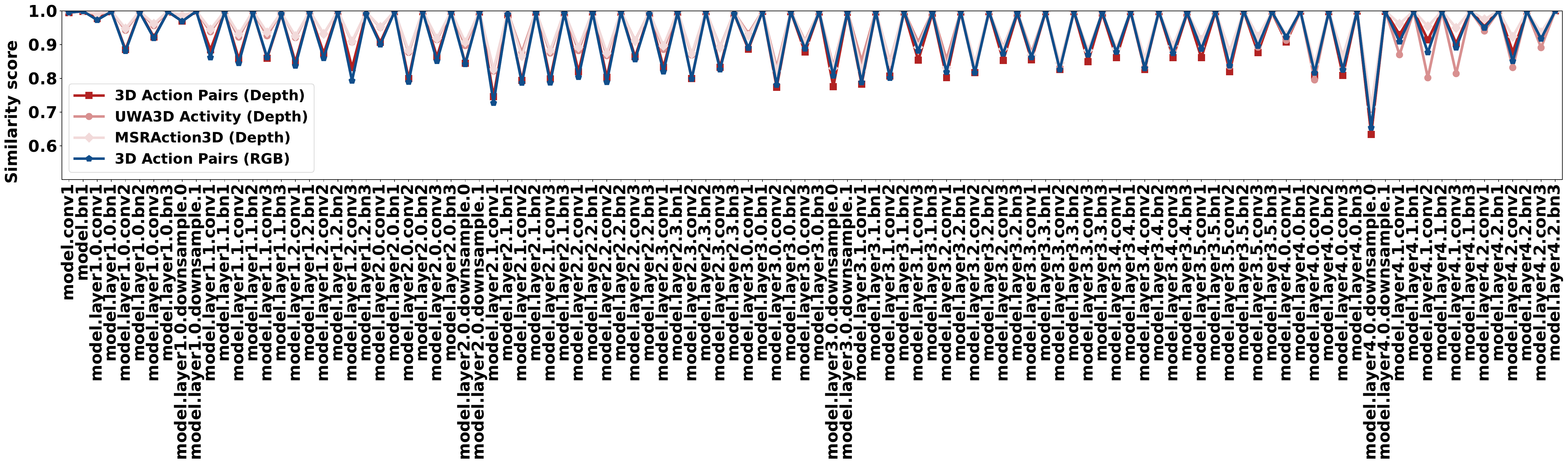}
\caption{Comparison of per-layer weight similarity between spatial and temporal block arrangements using ResNet-50 pretrained on ImageNet-1K and fine-tuned on three depth datasets (3D Action Pairs, UWA3D Activity, MSRAction3D) and one RGB dataset (3D Action Pairs).}
\label{fig:weightsimilarity_resnet_spatialtemporal}
\end{figure*}

We now present comparisons of per-layer weight similarity between spatial and temporal block arrangements on VideoMAE, ViT, and ResNet-50.

As illustrated in Fig.\ref{fig:weightsimilarity_videomae_spatialtemporal},~\ref{fig:weightsimilarity_vit_spatialtemporal}, and~\ref{fig:weightsimilarity_resnet_spatialtemporal}, the spatial and temporal block arrangements indeed lead to distinct weight patterns in all three different models.

In VideoMAE, both arrangements primarily influence the attention layers and some MLP layers. In contrast, on ViT, the effect is more pronounced in earlier layers, particularly attention and projection layers. Interestingly, on ResNet-50, the block arrangements impact nearly all convolutional layers. These observations suggest that VideoMAE excels at capturing both spatial and temporal information, while ResNet-50 is more focused on extracting spatial details. Overall, these results demonstrate that our TIME layer enables image-based models, such as ViT and ResNet-50, to be effectively adapted for video processing tasks.

\begin{figure*}[tbp] 
\centering
\includegraphics[trim=0.3cm 1.5cm 0.5cm 1cm, clip=true, width=0.9\linewidth]{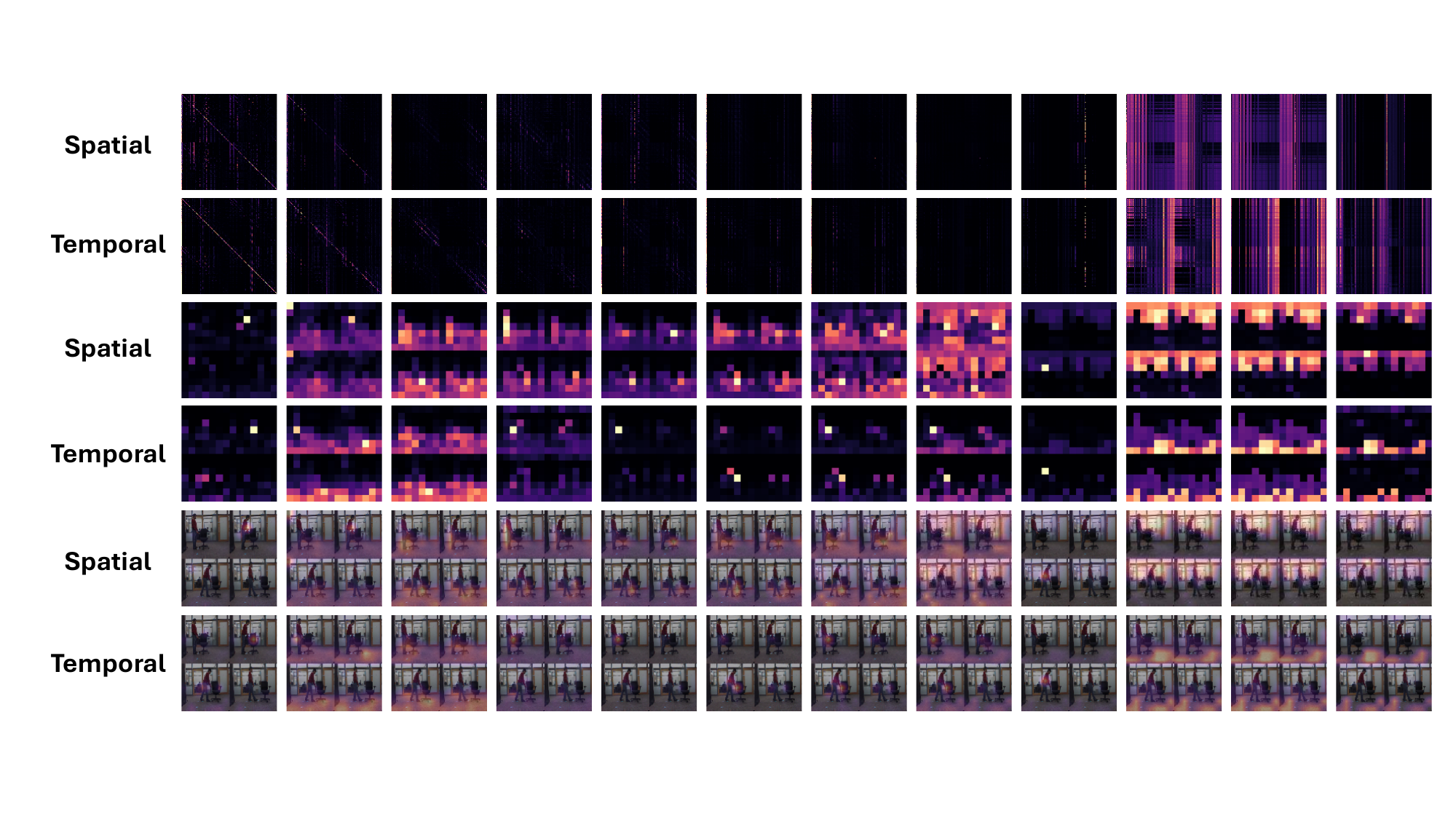}
\caption{Visual comparison of attention maps for the action \textit{push the chair}, between spatial and temporal block arrangements, using ViT (pretrained on ImageNet-1K and fine-tuned on 3D Action Pairs RGB).}
\label{fig:attentionmap_vit_rgb_spatialtemporal}
\end{figure*}

\begin{figure*}[tbp] 
\centering
\includegraphics[trim=0.3cm 1.5cm 0.5cm 1cm, clip=true, width=0.9\linewidth]{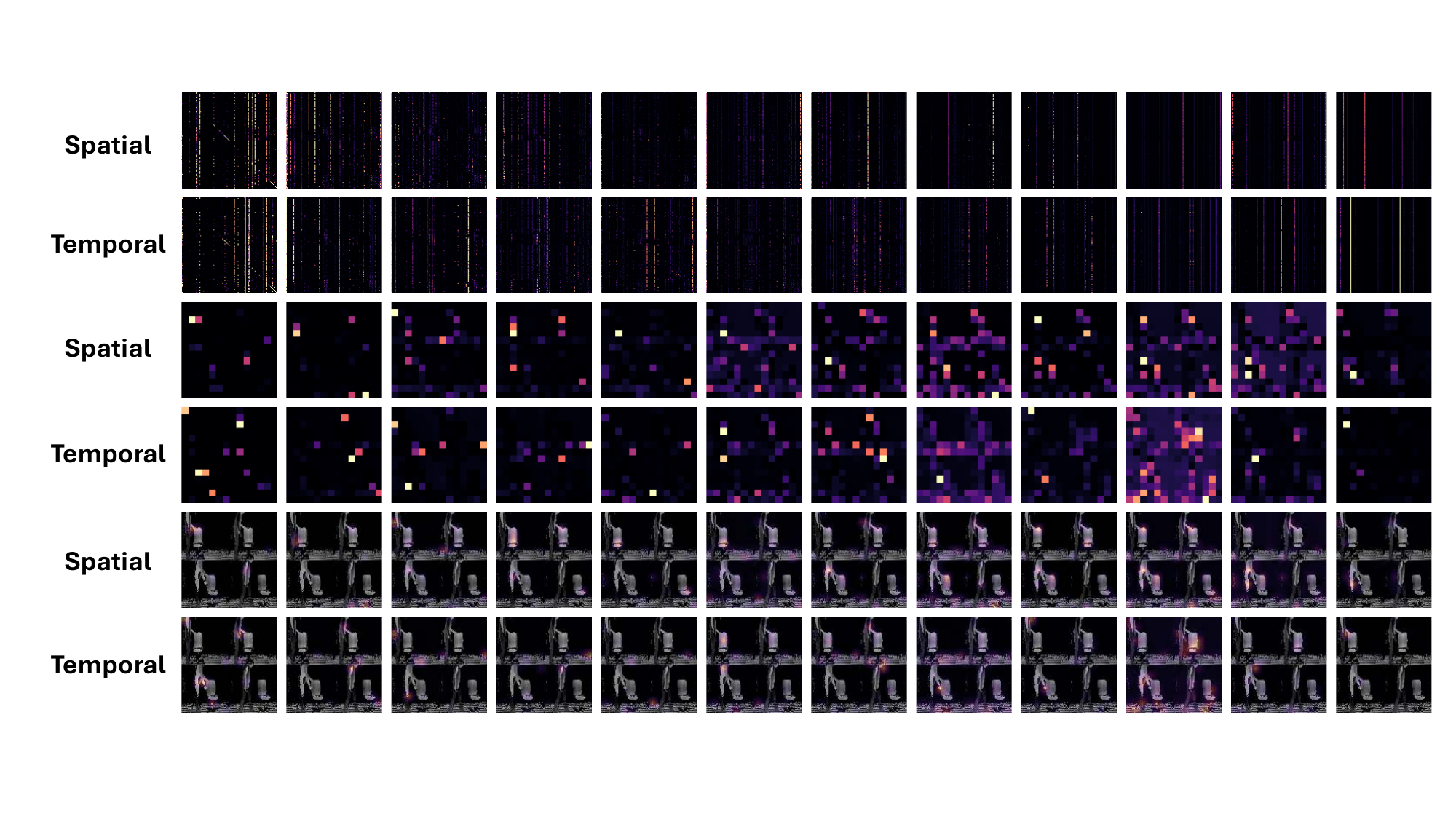}
\caption{Visual comparison of attention maps for the action \textit{push the chair}, between spatial and temporal block arrangements, using ViT (pretrained on ImageNet-1K and fine-tuned on 3D Action Pairs Depth).}
\label{fig:attentionmap_vit_depth_spatialtemporal}
\end{figure*}

\begin{figure*}[tbp] 
\centering
\includegraphics[trim=1.2cm 1.5cm 0.5cm 1cm, clip=true, width=0.9\linewidth]{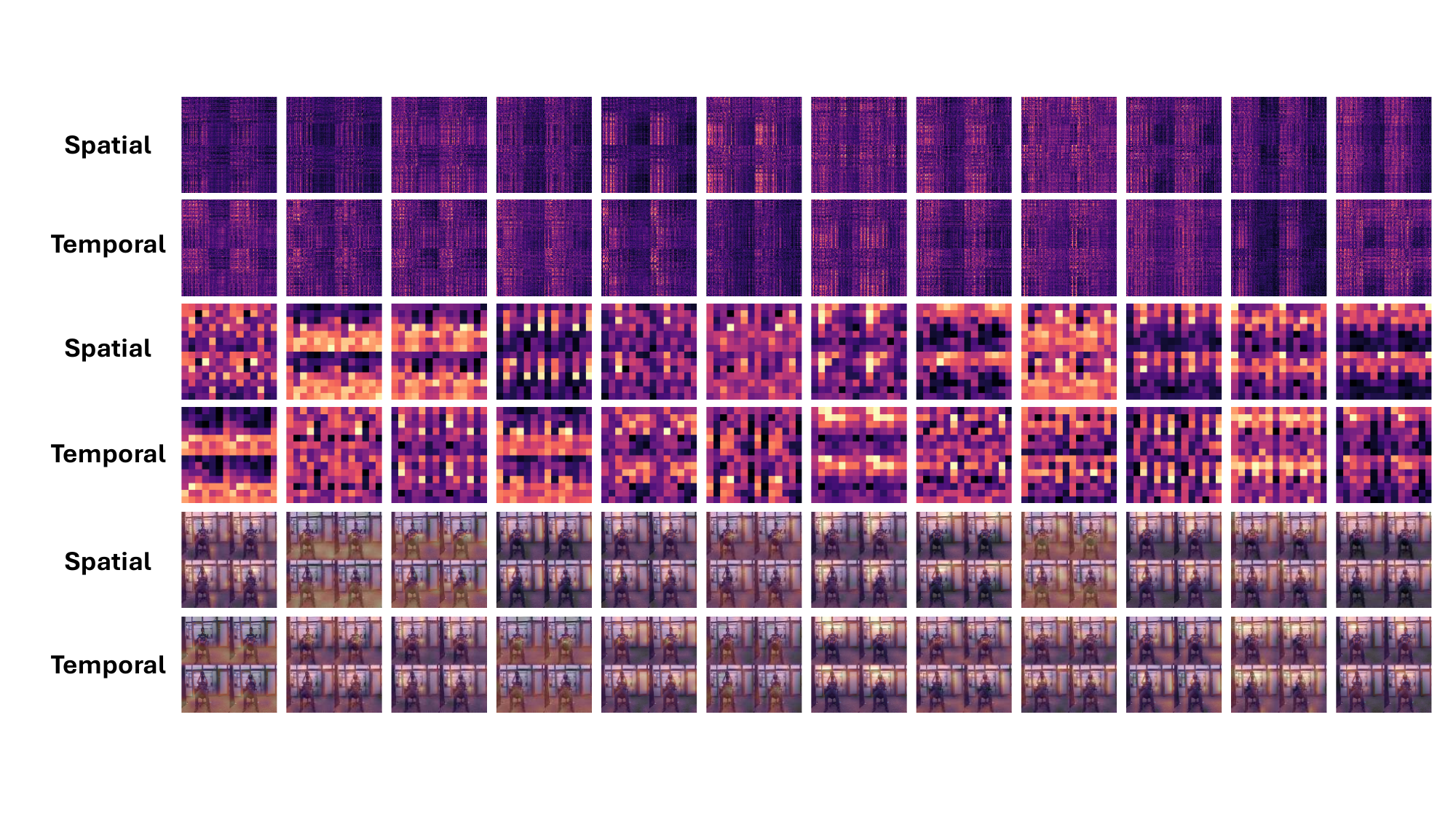}
\caption{Visual comparison of attention maps for the action \textit{put hat on}, between spatial and temporal block arrangements, using VideoMAE (trained from scratch on 3D Action Pairs RGB).}
\label{fig:attentionmap_videomae_rgb_spatialtemporal}
\end{figure*}

\begin{figure*}[tbp] 
\centering
\includegraphics[trim=1.2cm 1.5cm 0.5cm 1cm, clip=true, width=0.9\linewidth]{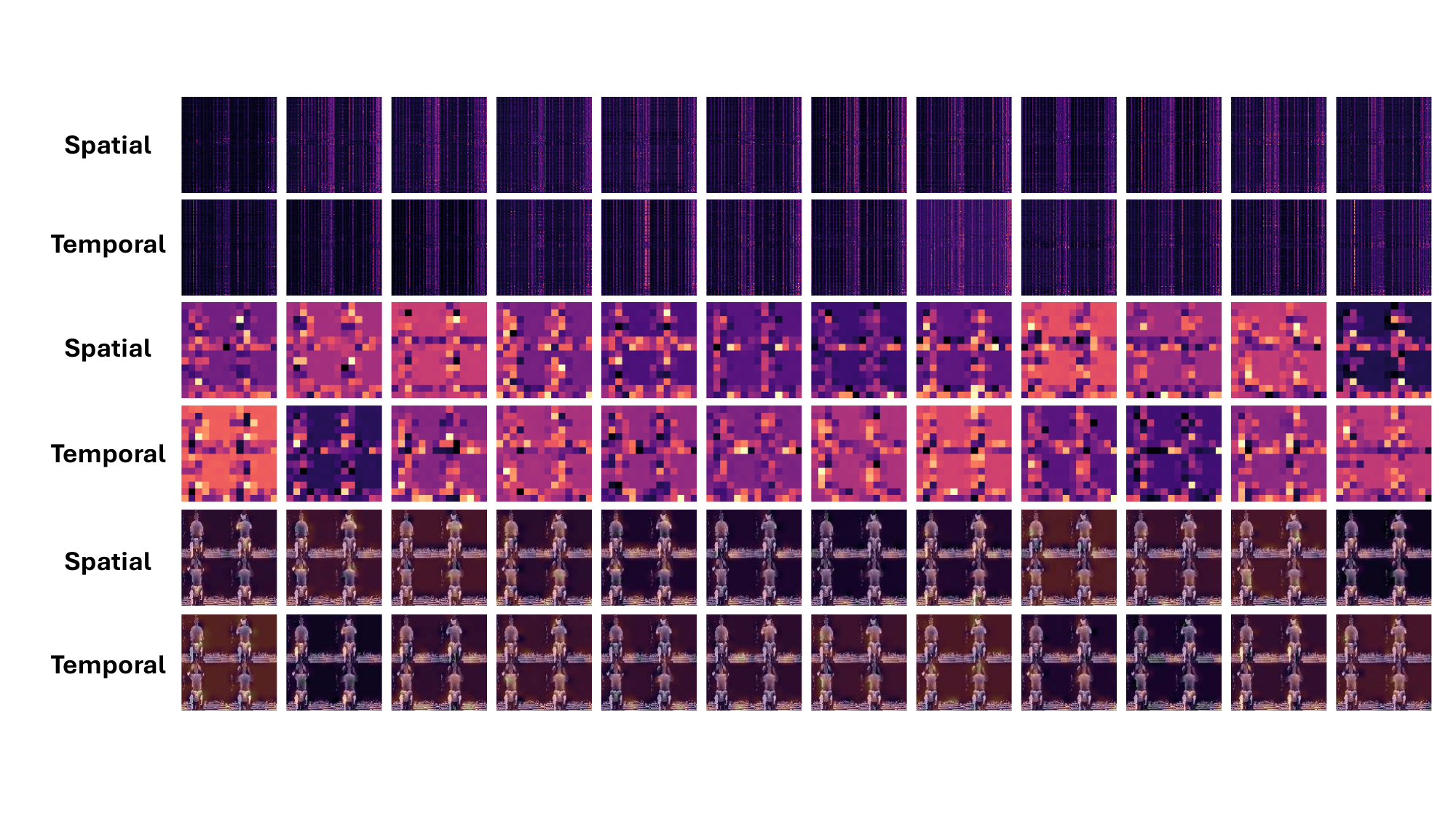}
\caption{Visual comparison of attention maps for the action \textit{put hat on}, between spatial and temporal block arrangements, using VideoMAE (trained from scratch on 3D Action Pairs Depth).}
\label{fig:attentionmap_videomae_depth_spatialtemporal}
\end{figure*}

\begin{figure*}[tbp] 
\centering
\includegraphics[trim=1.2cm 1.5cm 1.2cm 1cm, clip=true, width=0.9\linewidth]{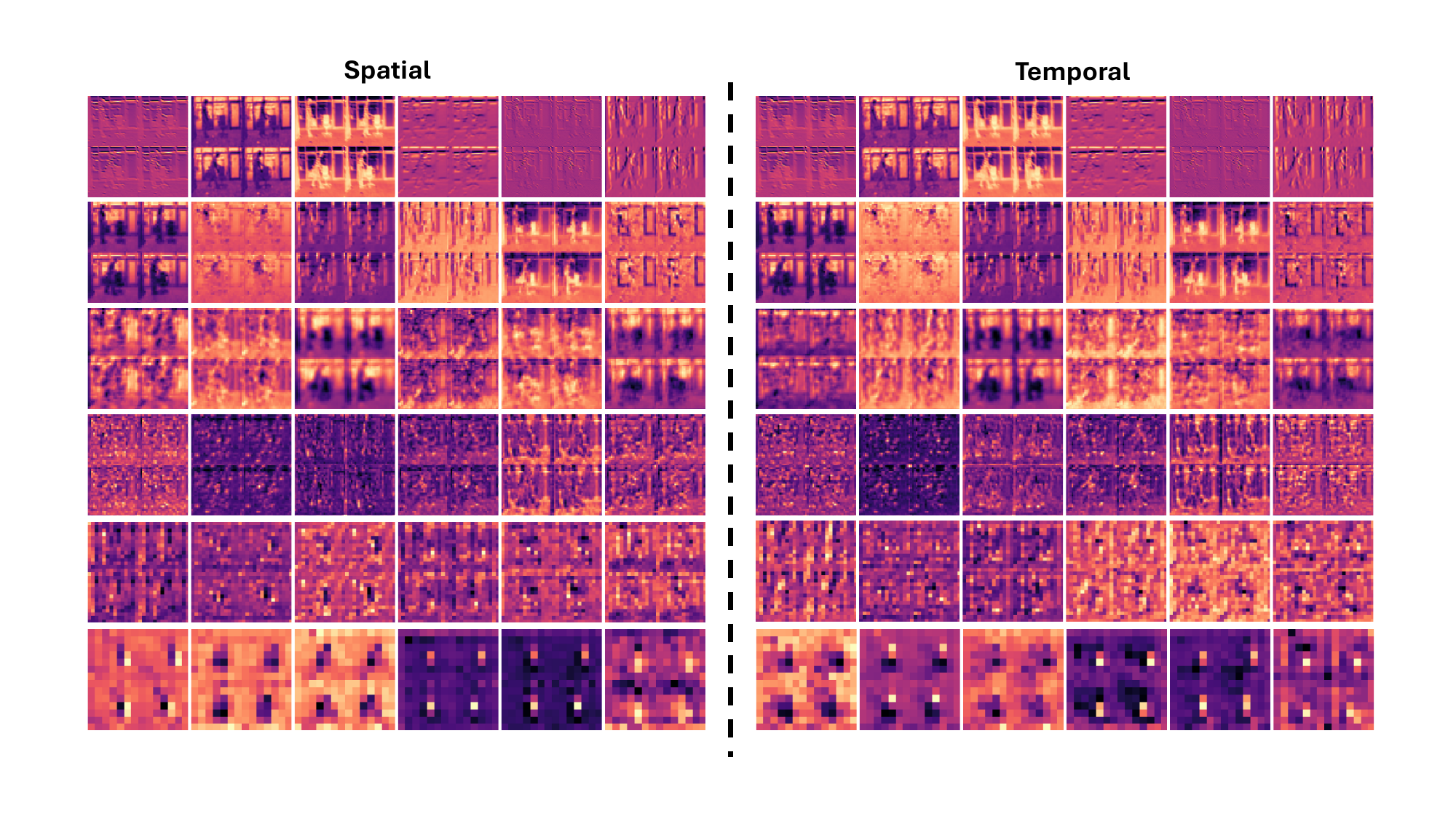}
\caption{Visual comparison of attention maps for the action \textit{push the chair}, between spatial and temporal block arrangements, using ResNet-50 (pretrained on ImageNet-1K and fine-tuned on 3D Action Pairs RGB).}
\label{fig:attentionmap_resnet_rgb_spatialtemporal}
\end{figure*}

\begin{figure*}[tbp] 
\centering
\includegraphics[trim=1.2cm 1.5cm 1.2cm 1cm, clip=true, width=0.9\linewidth]{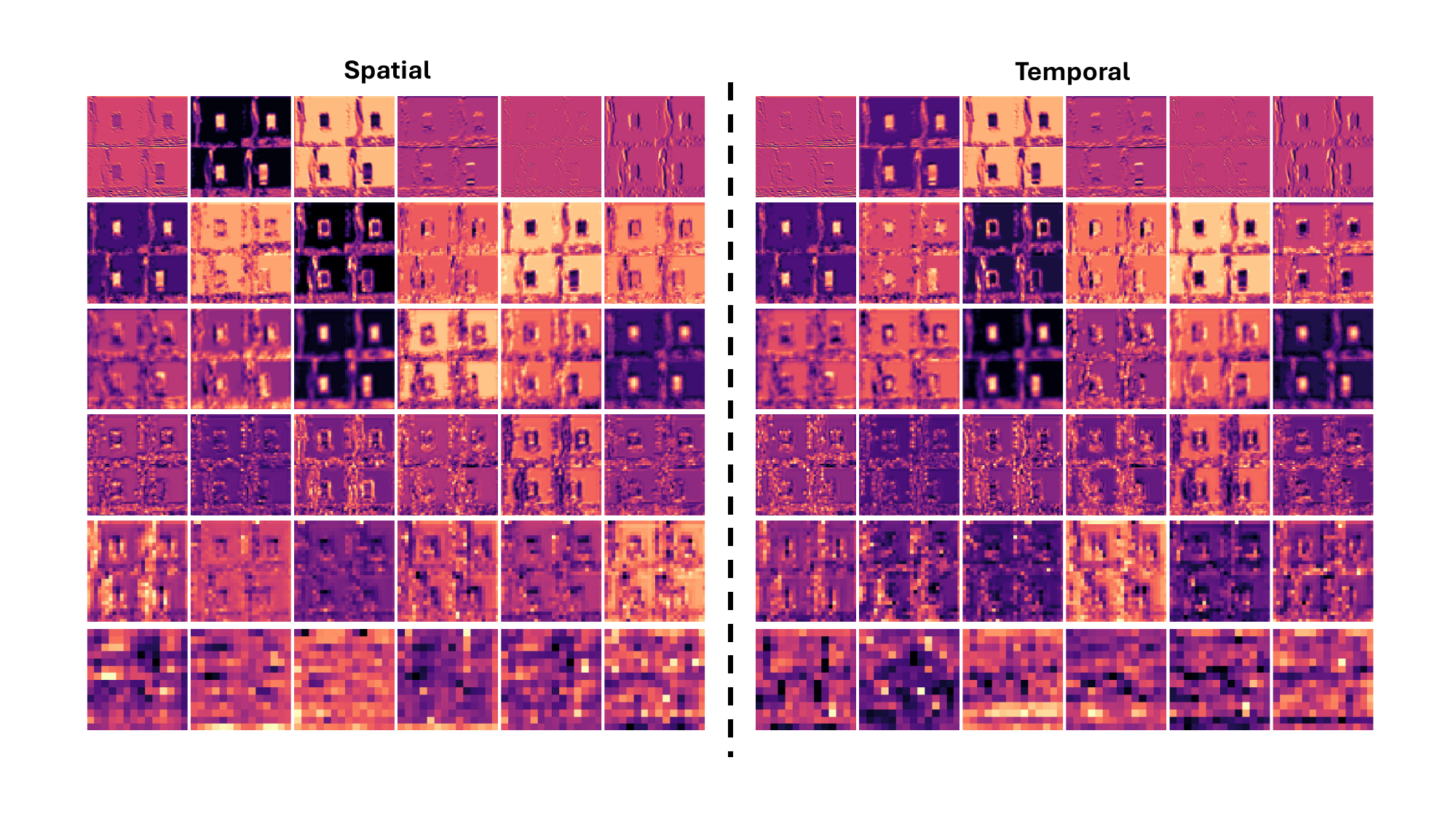}
\caption{Visual comparison of attention maps for the action \textit{push the chair}, between spatial and temporal block arrangements, using ResNet-50 (pretrained on ImageNet-1K and fine-tuned on 3D Action Pairs Depth).}
\label{fig:attentionmap_resnet_depth_spatialtemporal}
\end{figure*}

We further visualize the attention / feature maps generated by both spatial and temporal block arrangements on VideoMAE, ViT, and ResNet-50, for both RGB and depth modalities.

As shown in~\cref{fig:attentionmap_vit_rgb_spatialtemporal} and~\cref{fig:attentionmap_vit_depth_spatialtemporal}, for the RGB modality, the spatial block arrangement tends to exhibit more concentrated attention. This could be attributed to the 3D Action Pairs dataset, which consists of action pairs with similar motion trajectories, resulting in more focused attention on specific regions where both motion and frame order are crucial. In contrast, for the depth modality, both spatial and temporal arrangements show similar attention patterns. This may be due to the depth videos being clearer and more compact compared to the RGB videos, which reduces the variability in attention across different arrangements.

In~\cref{fig:attentionmap_resnet_rgb_spatialtemporal} and~\cref{fig:attentionmap_resnet_depth_spatialtemporal}, we observe that for ResNet-50, both spatial and temporal block arrangements behave similarly. This is likely because ResNet-50, a model originally designed for image classification, excels at capturing spatial details, which reduces the difference in attention patterns between the two arrangements.

As shown in~\cref{fig:attentionmap_videomae_rgb_spatialtemporal} and~\cref{fig:attentionmap_videomae_depth_spatialtemporal}, when training VideoMAE from scratch, both spatial and temporal block arrangements exhibit similar attention patterns. This suggests that large-scale pretraining with the TIME layer in VideoMAE has the potential to improve performance on downstream tasks. VideoMAE excels at capturing both spatial and temporal information, and the TIME layer, acting as a bridge between these two modalities, further enhances the model’s ability to extract spatio-temporal features essential for motion-related tasks.

\section{Visualizations of Attention and Feature Maps}
\label{appendix:attn}

\begin{figure*}[tbp] 
\centering
\includegraphics[trim=1.2cm 2.2cm 0.2cm 2cm, clip=true, width=0.9\linewidth]{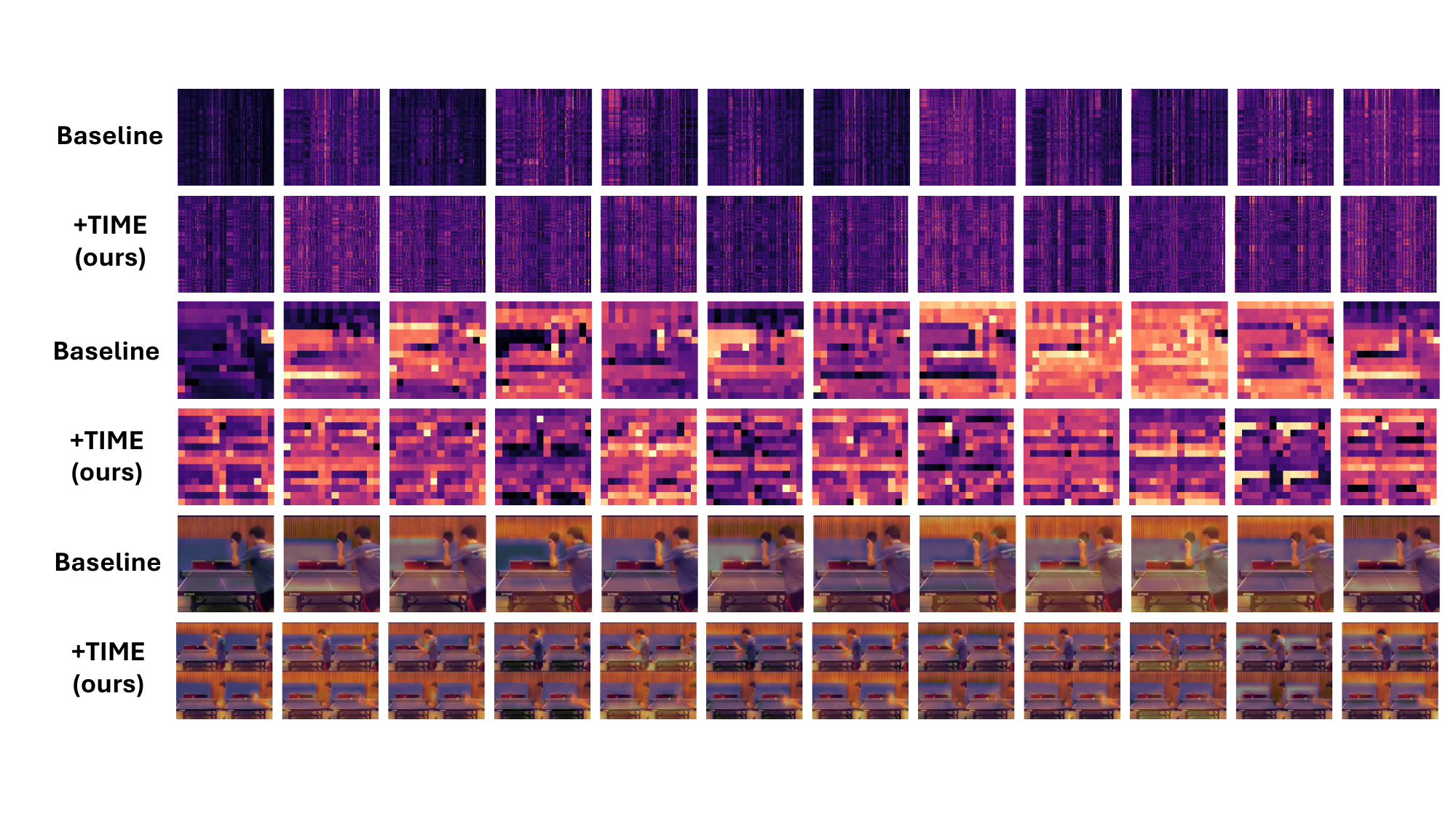}
\caption{Visual comparison of attention maps for the action \textit{table tennis}, between the baseline (without TIME layer) and the model with the TIME layer, using VideoMAE (train from scratch on UCF101 RGB).}
\label{fig:attentionmap_ucf_videomae}
\end{figure*}

\begin{figure*}[tbp] 
\centering
\includegraphics[trim=1.2cm 2.2cm 0.2cm 2cm, clip=true, width=0.9\linewidth]{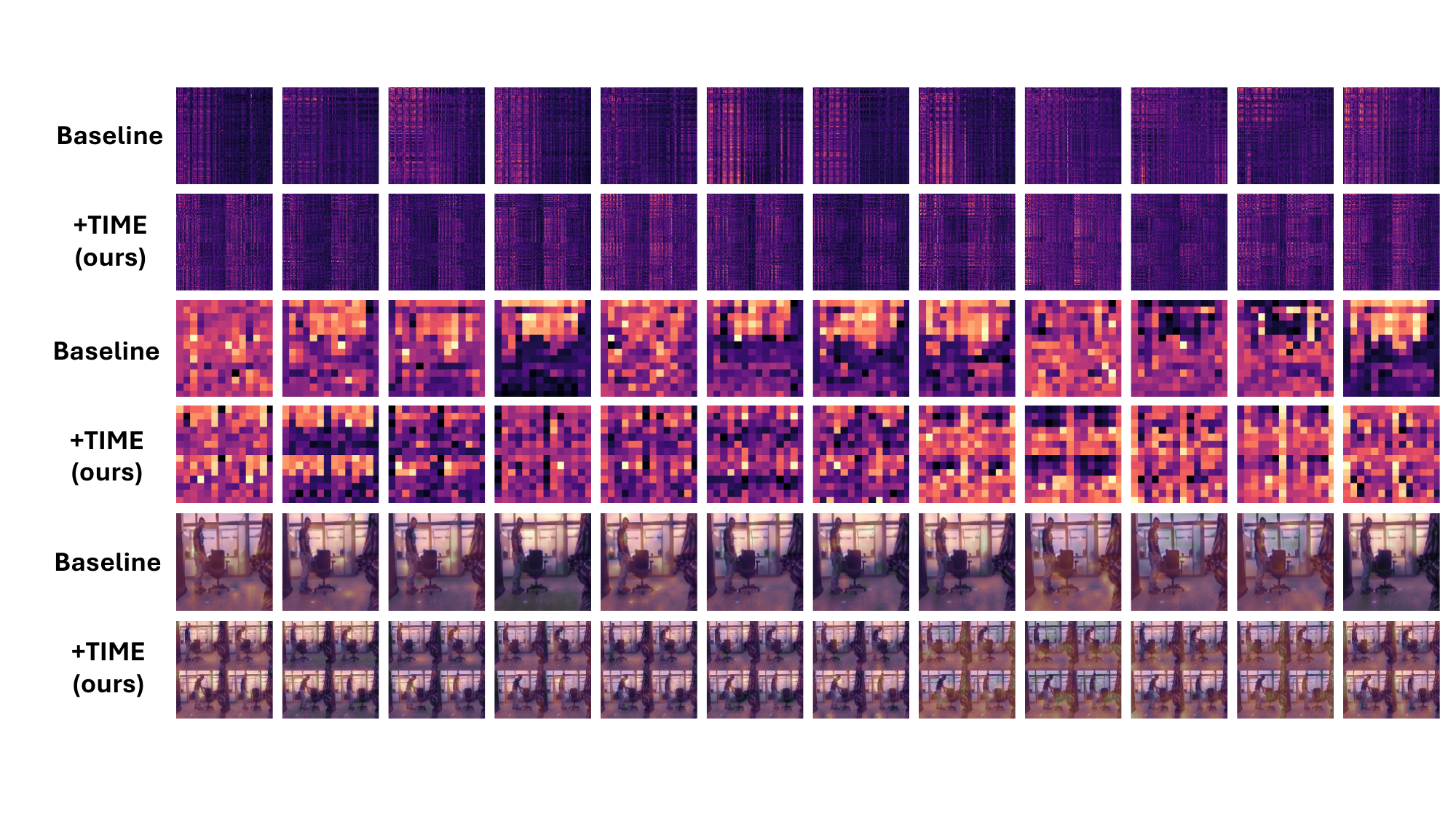}
\caption{Visual comparison of attention maps for the action \textit{push the chair}, between the baseline (without TIME layer) and the model with the TIME layer, using VideoMAE (train from scratch on 3D Action Pairs RGB).}
\label{fig:attentionmap_3D_rgb_videomae}
\end{figure*}

\begin{figure*}[tbp] 
\centering
\includegraphics[trim=1.2cm 2.2cm 0.2cm 1.5cm, clip=true, width=0.9\linewidth]{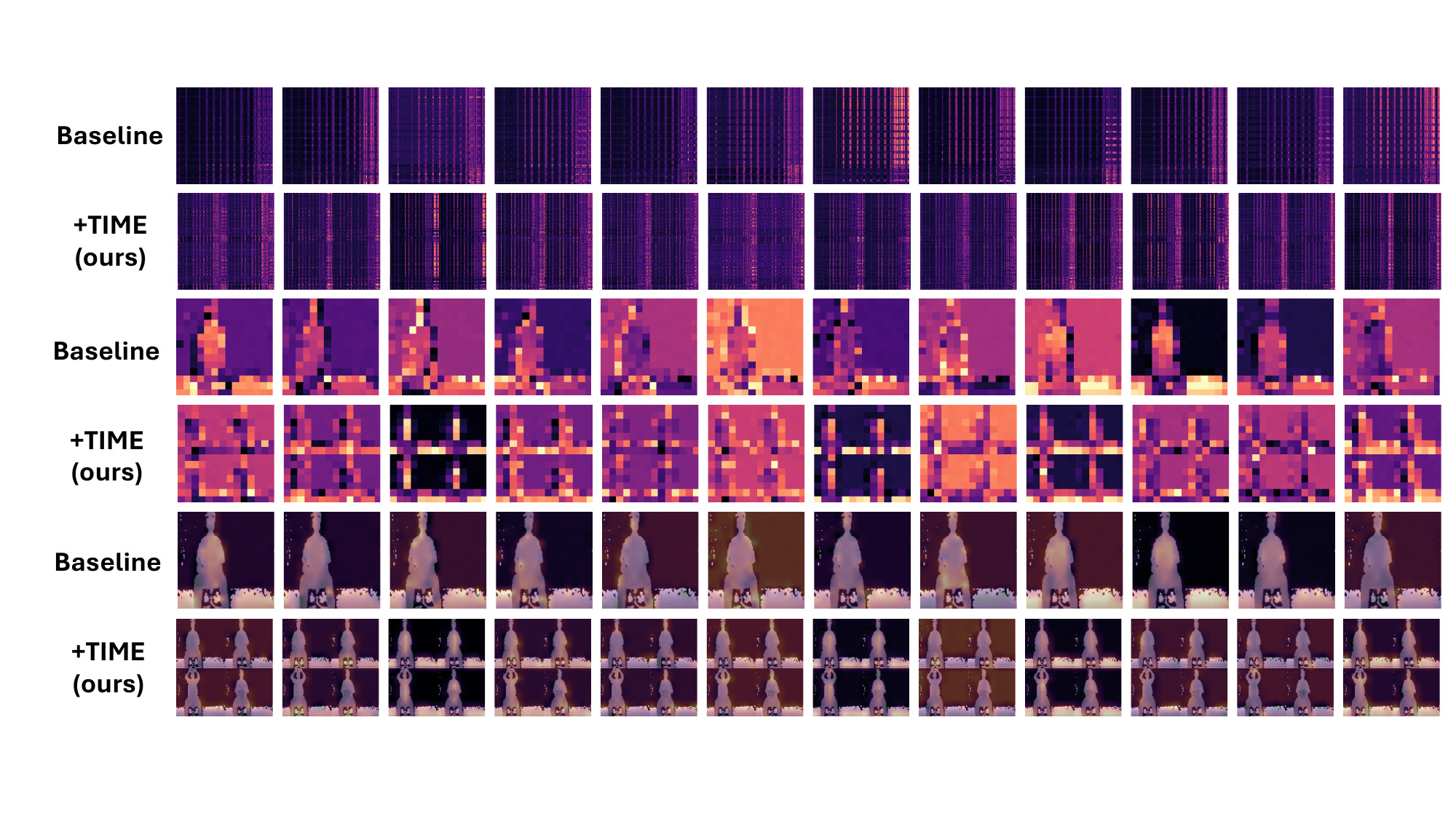}
\caption{Visual comparison of attention maps for the action \textit{take off hat}, between the baseline (without TIME layer) and the model with the TIME layer, using VideoMAE (train from scratch on 3D Action Pairs Depth).}
\label{fig:attentionmap_3D_depth_videomae}
\end{figure*}

\begin{figure*}[tbp] 
\centering
\includegraphics[trim=1.2cm 2.2cm 0.2cm 1.5cm, clip=true, width=0.9\linewidth]{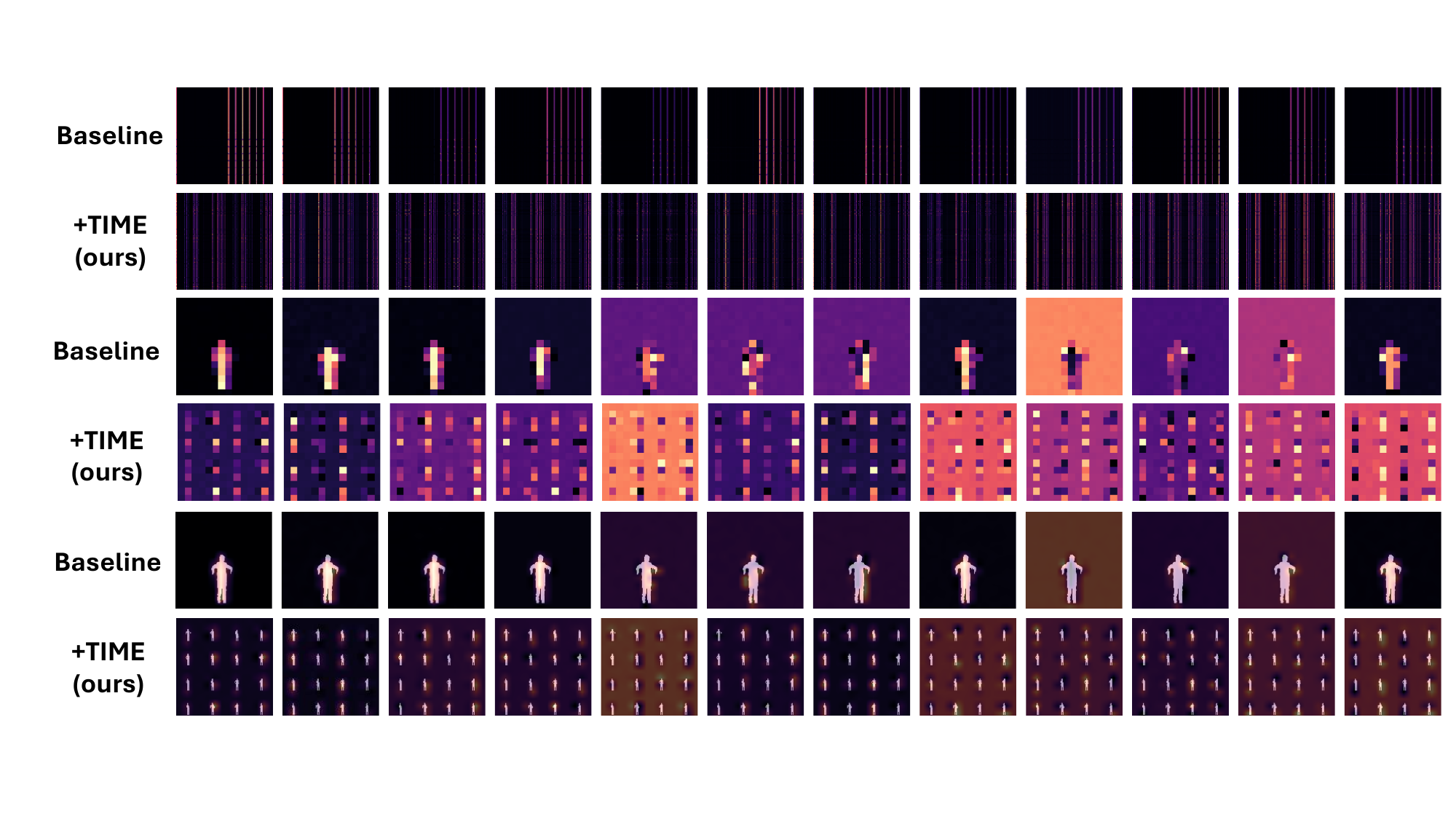}
\caption{Visual comparison of attention maps for the action \textit{hands on waist}, between the baseline (without TIME layer) and the model with the TIME layer, using VideoMAE (train from scratch on UWA3D Activity Depth).}
\label{fig:attentionmap_uwa_depth_videomae}
\end{figure*}

\begin{figure*}[tbp] 
\centering
\includegraphics[trim=1.2cm 2.2cm 0.2cm 2.2cm, clip=true, width=0.9\linewidth]{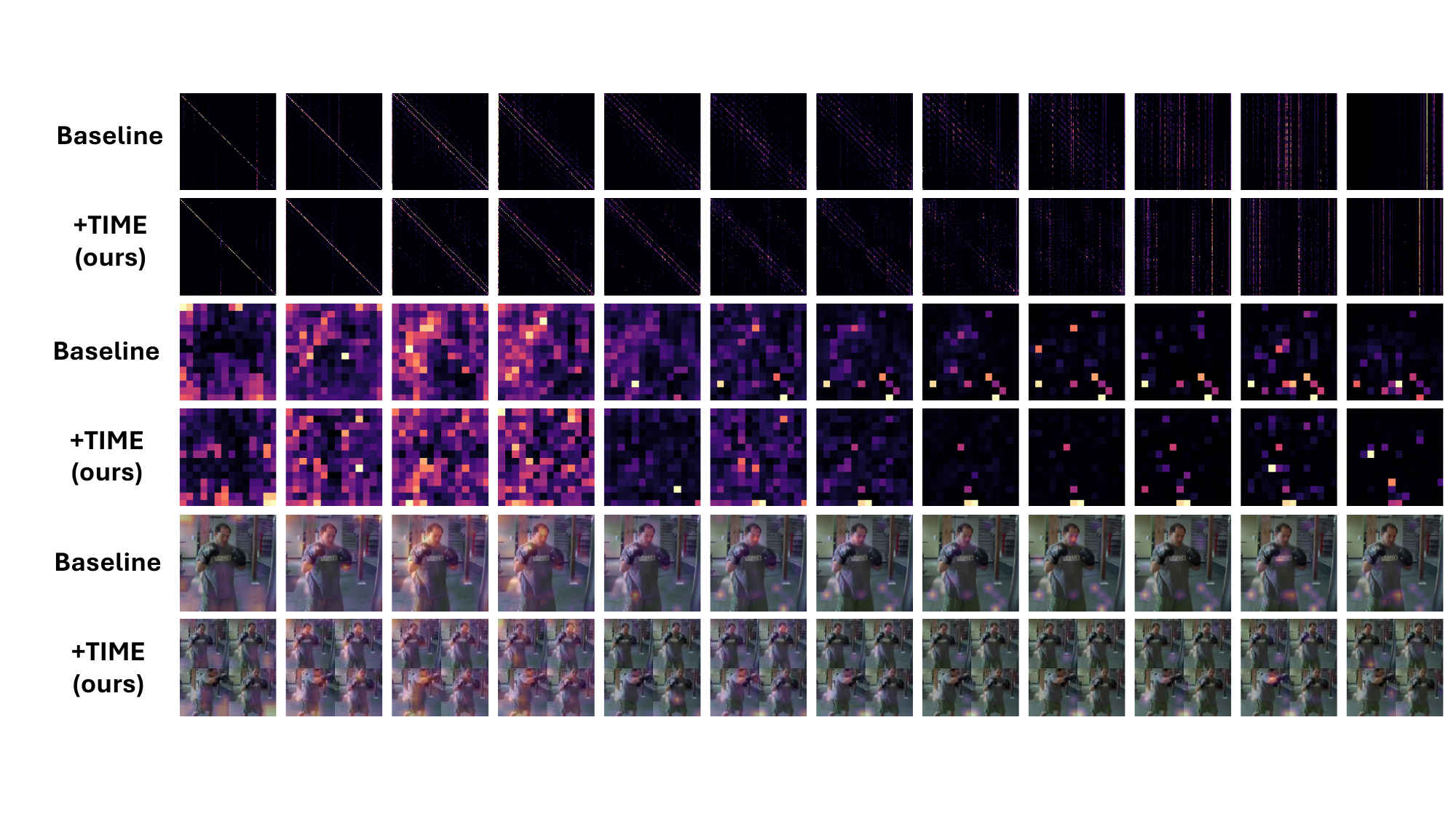}
\caption{Visual comparison of attention maps for the action \textit{boxing}, between the baseline (without TIME layer) and the model with the TIME layer, using ViT (pretrained on ImageNet-1K and fine-tuned on HMDB51 RGB).}
\label{fig:attentionmap_vit_hmdb51}
\end{figure*}

\begin{figure*}[tbp] 
\centering
\includegraphics[trim=1.2cm 2.2cm 0.2cm 2cm, clip=true, width=0.9\linewidth]{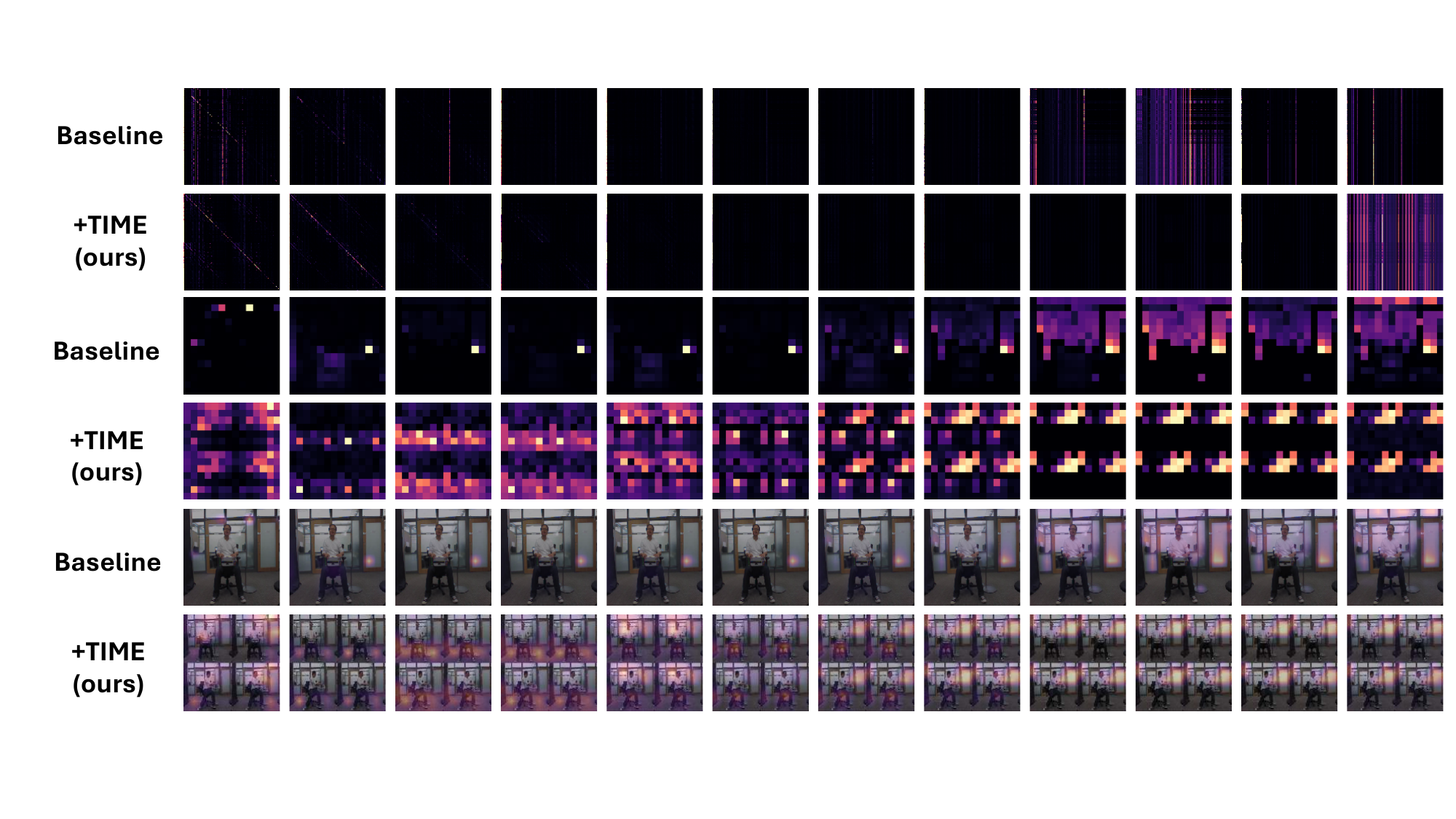}
\caption{Visual comparison of attention maps for the action \textit{put box on the desk}, between the baseline (without TIME layer) and the model with the TIME layer, using ViT (pretrained on ImageNet-1K and fine-tuned on 3D Action Pairs RGB).}
\label{fig:attentionmap_vit_3D_rgb}
\end{figure*}

\begin{figure*}[tbp] 
\centering
\includegraphics[trim=1.2cm 2.2cm 0.2cm 2cm, clip=true, width=0.9\linewidth]{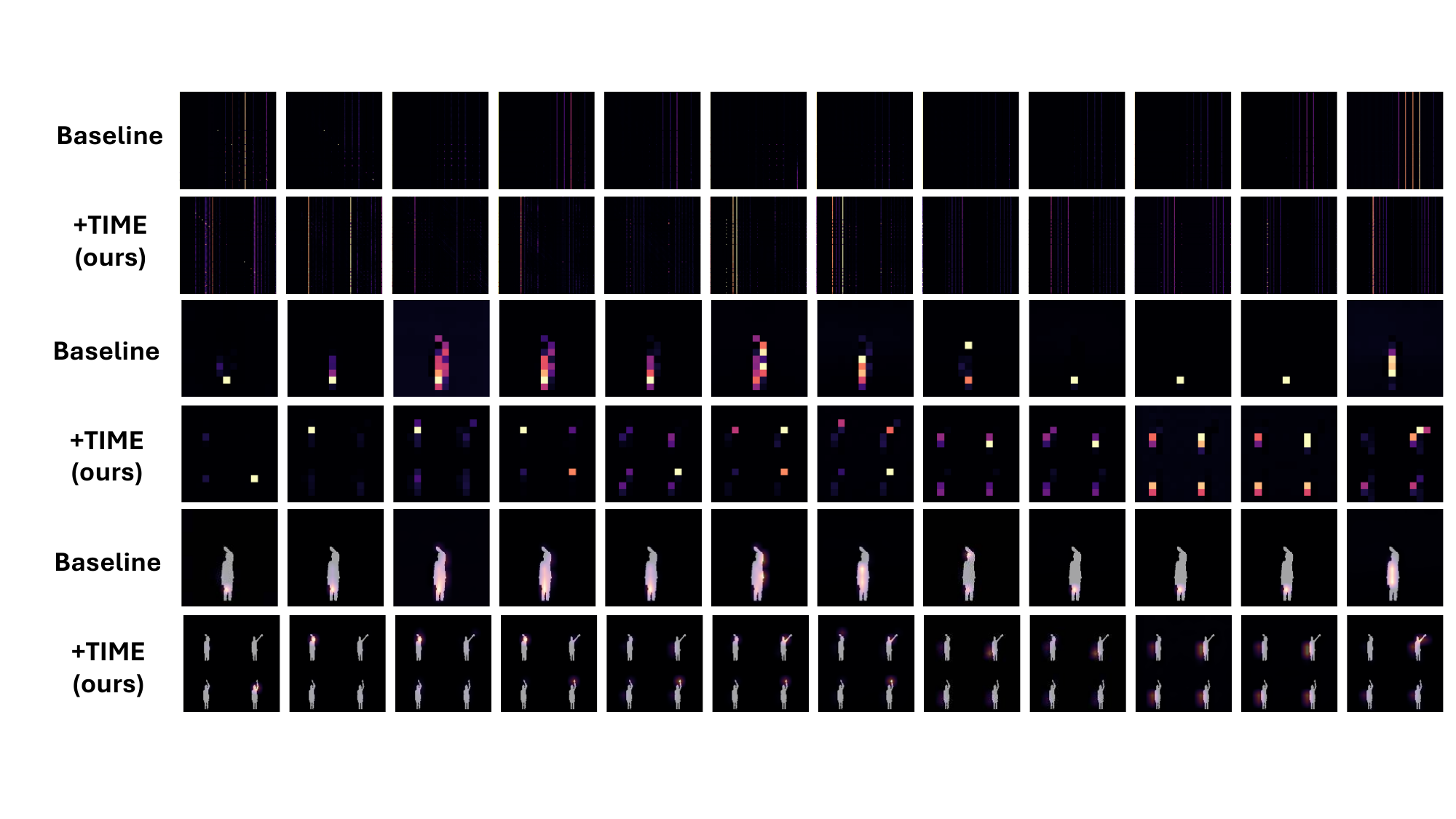}
\caption{Visual comparison of attention maps for the action \textit{high arm wave}, between the baseline (without TIME layer) and the model with the TIME layer, using ViT (pretrained on ImageNet-1K and fine-tuned on UWA3D Activity Depth).}
\label{fig:attentionmap_vit_uwa_depth}
\end{figure*}

\begin{figure*}[tbp] 
\centering
\includegraphics[trim=1.2cm 1.7cm 1.2cm 1cm, clip=true, width=0.9\linewidth]{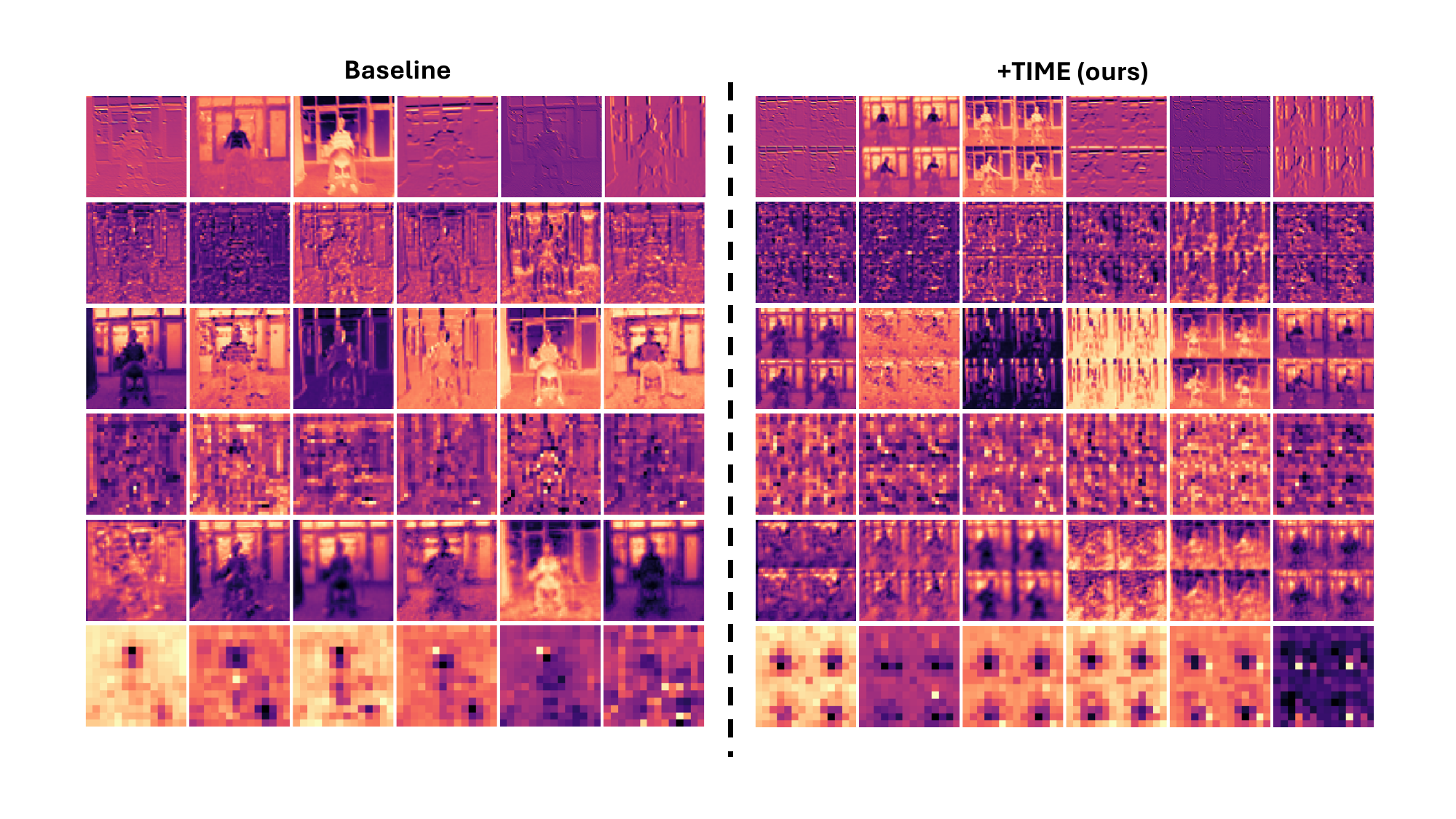}
\caption{Visual comparison of feature maps for the action \textit{pick up box on the desk}, between the baseline (without TIME layer) and the model with the TIME layer, using ResNet-50 (pretrained on ImageNet-1K and fine-tuned on 3D Action Pairs RGB). Each row displays the feature maps from a specific layer, listed from top to bottom in the following order: \texttt{conv1}, \texttt{layer1[0].conv1}, \texttt{layer1[0].conv2}, \texttt{layer2[0].conv1}, \texttt{layer3[0].conv1}, and \texttt{layer4[0].conv1}. One feature map per layer is shown.}
\label{fig:attentionmap_resnet_3D_rgb}
\end{figure*}

\begin{figure*}[tbp] 
\centering
\includegraphics[trim=1.2cm 1.5cm 1.2cm 1cm, clip=true, width=0.9\linewidth]{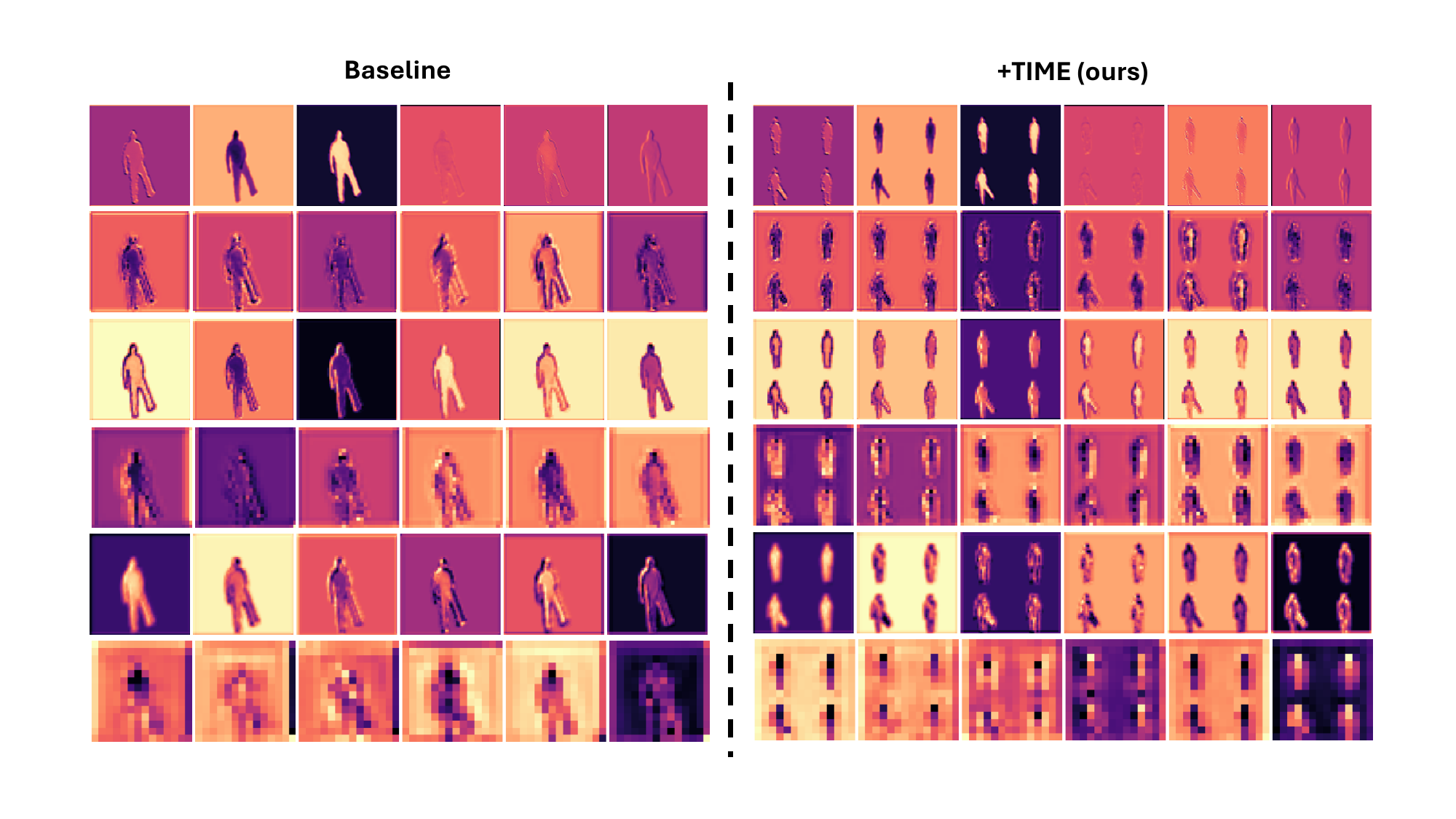}
\caption{Visual comparison of feature maps for the action \textit{side kick}, between the baseline (without TIME layer) and the model with the TIME layer, using ResNet-50 (pretrained on ImageNet-1K and fine-tuned on MSRAction3D Depth). Each row displays the feature maps from a specific layer, listed from top to bottom in the following order: \texttt{conv1}, \texttt{layer1[0].conv1}, \texttt{layer1[0].conv2}, \texttt{layer2[0].conv1}, \texttt{layer3[0].conv1}, and \texttt{layer4[0].conv1}. One feature map per layer is shown.}
\label{fig:attentionmap_resnet_msr}
\end{figure*}

\cref{fig:attentionmap_ucf_videomae} to \ref{fig:attentionmap_resnet_msr} present visual comparisons of attention and feature maps between the baseline (without the TIME layer) and the model with the TIME layer. These comparisons use different model architectures, including both training from scratch and fine-tuning, across various video modalities.

Overall, when applied to depth videos, the use of the TIME layer leads to more clear and compact attention maps and feature representations compared to those generated from RGB videos (\eg, \cref{fig:attentionmap_3D_rgb_videomae} vs. \cref{fig:attentionmap_3D_depth_videomae} for VideoMAE, \cref{fig:attentionmap_vit_3D_rgb} vs. \cref{fig:attentionmap_vit_uwa_depth} for ViT, \cref{fig:attentionmap_resnet_3D_rgb} vs. \cref{fig:attentionmap_resnet_msr}). This enhancement is particularly noticeable in terms of how the model captures and organizes spatial and temporal information.

The attention maps from Vision Transformers (ViT) tend to exhibit weaker responses to temporal dynamics, struggling to fully capture the evolution of features across time. In contrast, the VideoMAE model demonstrates a more uniform focus across all regions of temporal dynamics, effectively maintaining attention throughout the video sequence (\eg, \cref{fig:attentionmap_uwa_depth_videomae} vs. \cref{fig:attentionmap_vit_uwa_depth} for depth, \cref{fig:attentionmap_3D_rgb_videomae} vs. \cref{fig:attentionmap_vit_3D_rgb} for RGB).

When using a ResNet-50 backbone, depth video feature maps are generally more distinct and compact, showing clearer boundaries and less noise than those derived from RGB videos (\eg, \cref{fig:attentionmap_resnet_3D_rgb} vs. \cref{fig:attentionmap_resnet_msr}). In contrast, the feature maps from RGB videos often suffer from fuzzier, more ambiguous features with less defined boundaries.

In summary, the use of the TIME layer consistently improves the model’s ability to capture temporal information across various architectures, from CNNs (\eg, ResNet-50) to Transformers (\eg, ViT), including self-supervised pretraining models like VideoMAE. This enhancement is observed for both RGB and depth modalities, underscoring the TIME layer’s ability to enhance temporal understanding regardless of the model architecture or input modality.

\section{Visualizations of Sequence Reconstructions}
\label{appendix:reconstr}

\begin{figure*}[tbp] 
\centering
\includegraphics[width=0.9\linewidth]{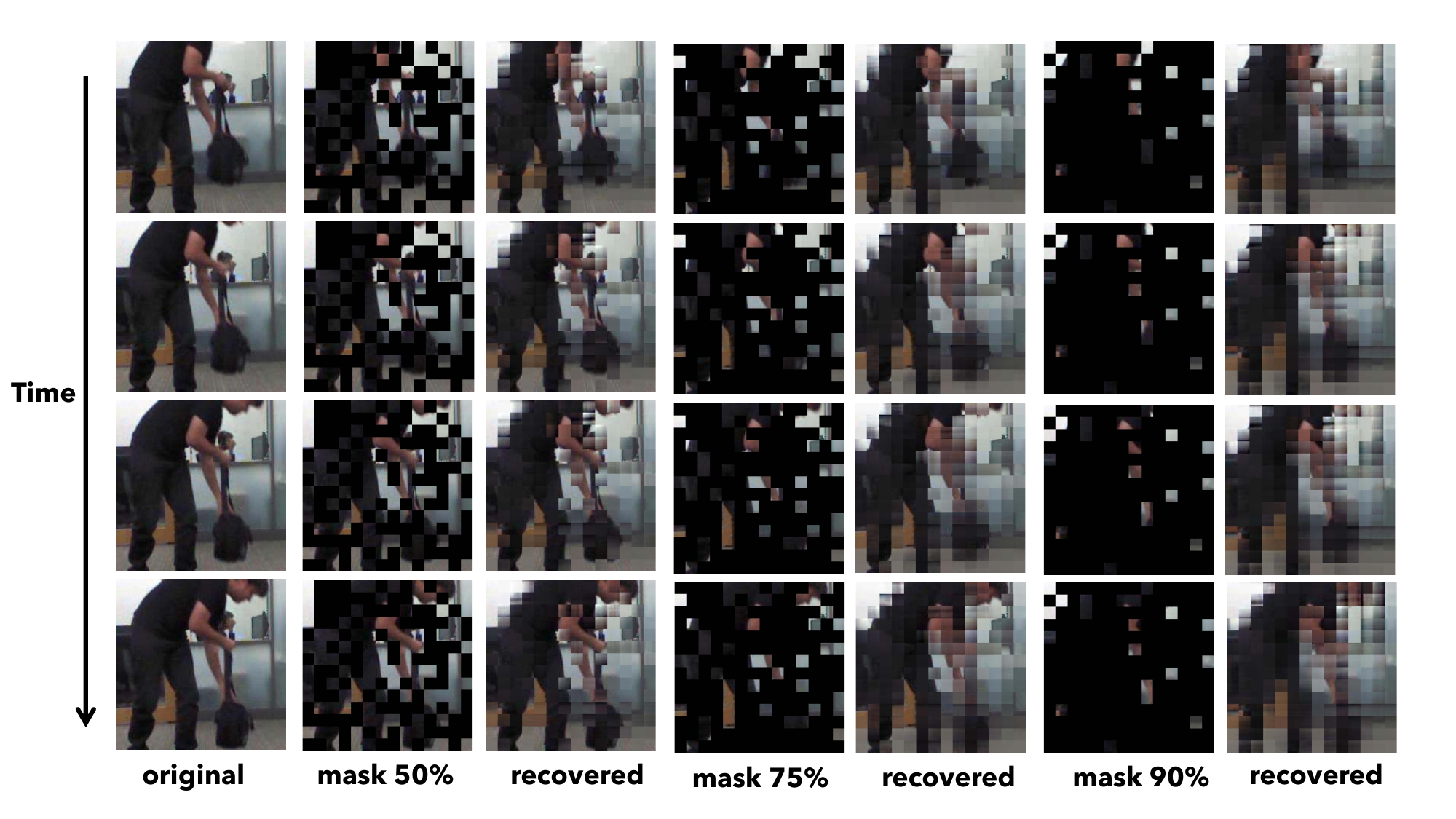}
\caption{We present our new video sequence ($N=1$, Baseline) along with reconstructions at different masking ratios. The video reconstructions are predicted by VideoMAE with the TIME layer, pre-trained using masking ratios of 50\%, 75\%, and 90\%. For this visualization, we select the action \textit{put bag on the floor} from the 3D Action Pairs RGB dataset.}
\label{fig:3D_rgb_1}
\end{figure*}

\begin{figure*}[tbp] 
\centering
\includegraphics[width=0.9\linewidth]{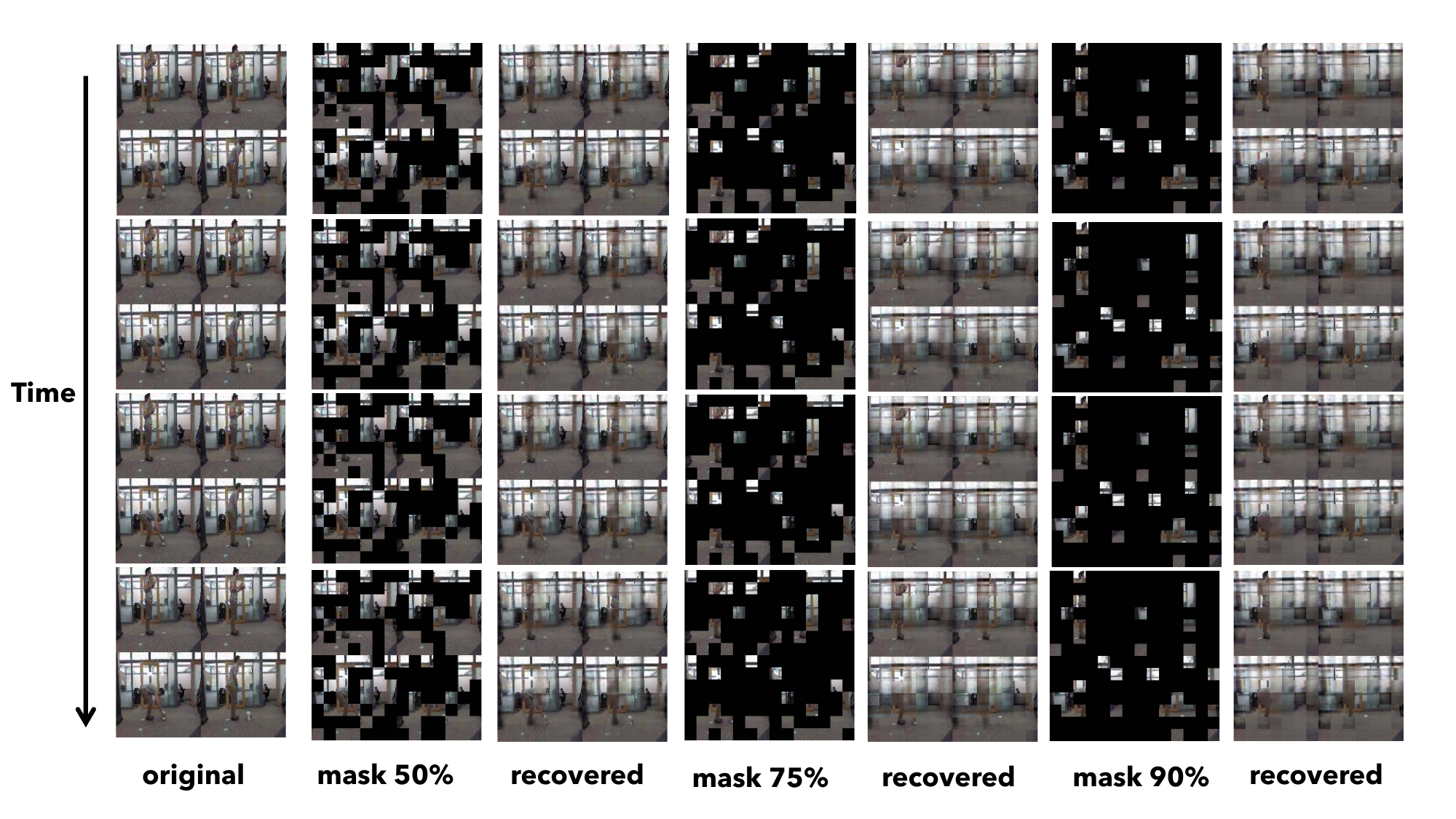}
\caption{We present our new video sequence ($N=2$) along with reconstructions at different masking ratios. The video reconstructions are predicted by VideoMAE with the TIME layer, pre-trained using masking ratios of 50\%, 75\%, and 90\%. For this visualization, we select the action \textit{put box on the floor} from the 3D Action Pairs RGB dataset.}
\label{fig:3D_rgb_2}
\end{figure*}

\begin{figure*}[tbp] 
\centering
\includegraphics[width=0.9\linewidth]{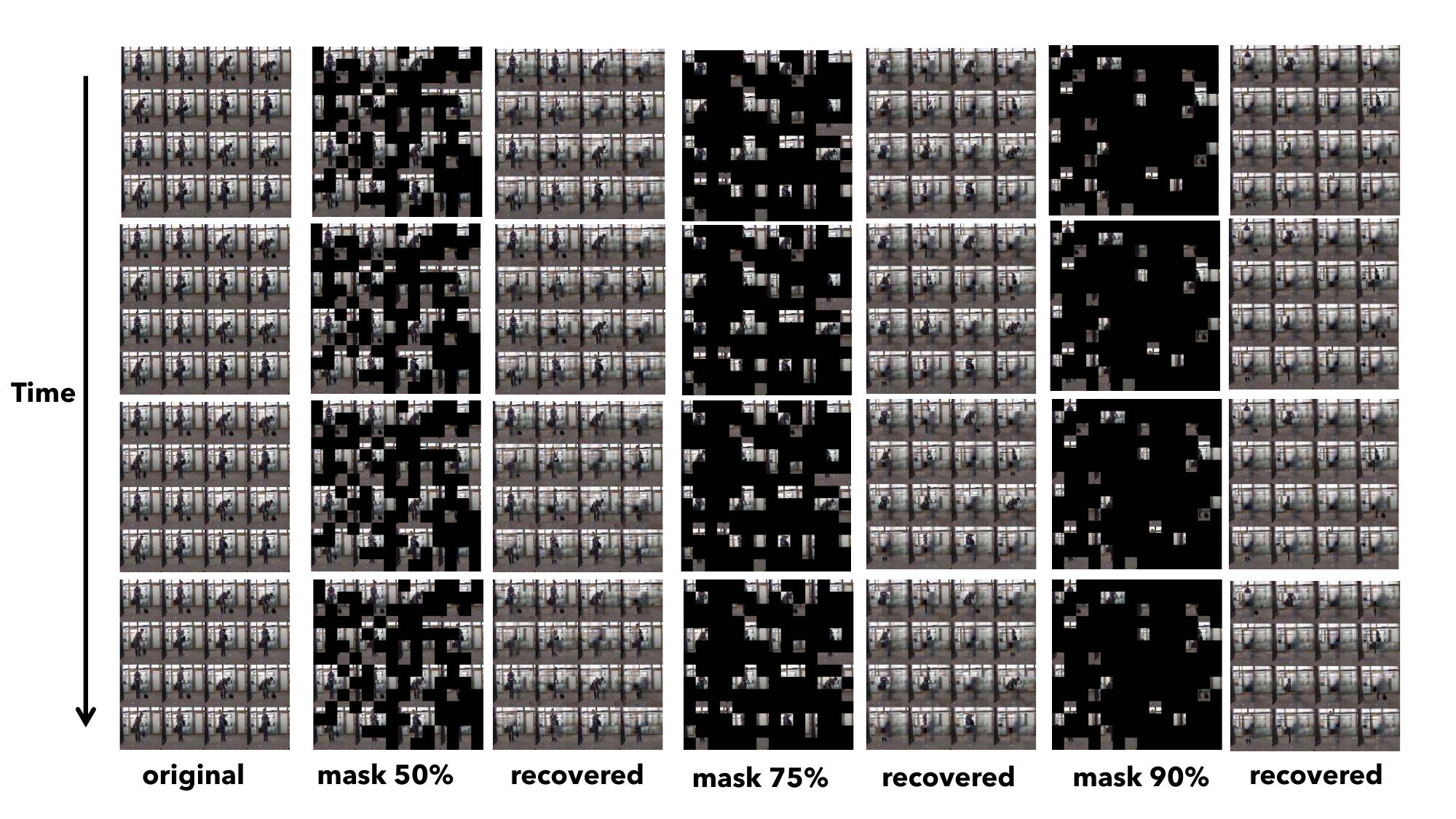}
\caption{We present our new video sequence ($N=4$) along with reconstructions at different masking ratios. The video reconstructions are predicted by VideoMAE with the TIME layer, pre-trained using masking ratios of 50\%, 75\%, and 90\%. For this visualization, we select the action \textit{pick up bag and put it on back} from the 3D Action Pairs RGB dataset.}
\label{fig:3D_rgb_4}
\end{figure*}

\begin{figure*}[tbp] 
\centering
\includegraphics[width=0.9\linewidth]{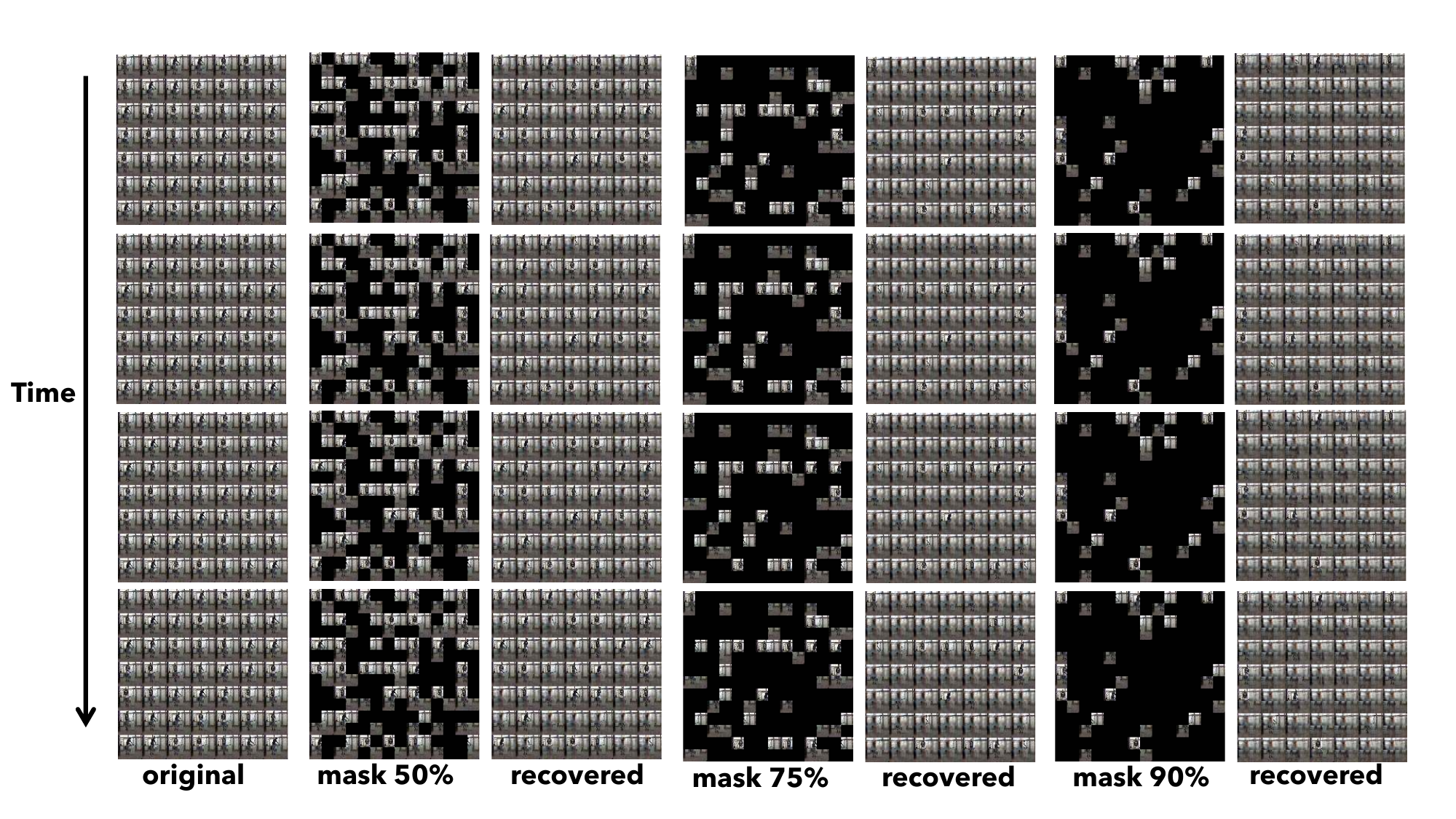}
\caption{We present our new video sequence ($N=7$) along with reconstructions at different masking ratios. The video reconstructions are predicted by VideoMAE with the TIME layer, pre-trained using masking ratios of 50\%, 75\%, and 90\%. For this visualization, we select the action \textit{pick up hanging paper from desk side} from the 3D Action Pairs RGB dataset.}
\label{fig:3D_rgb_7}
\end{figure*}

\begin{figure*}[tbp] 
\centering
\includegraphics[width=0.9\linewidth]{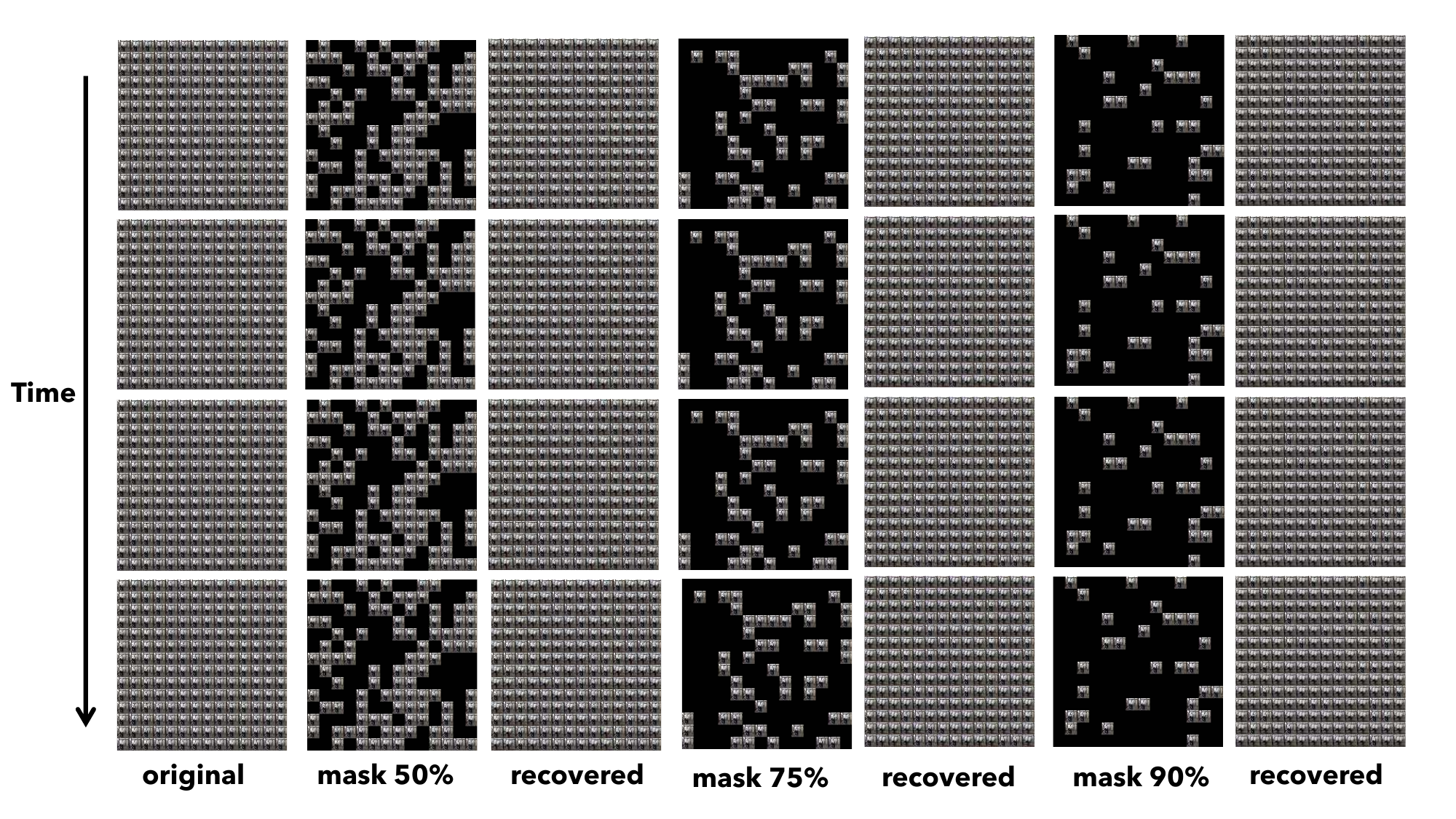}
\caption{We present our new video sequence ($N=14$) along with reconstructions at different masking ratios. The video reconstructions are predicted by VideoMAE with the TIME layer, pre-trained using masking ratios of 50\%, 75\%, and 90\%. For this visualization, we select the action \textit{put hat on head} from the 3D Action Pairs RGB dataset.}
\label{fig:3D_rgb_14}
\end{figure*}

\begin{figure*}[tbp] 
\centering
\includegraphics[width=0.9\linewidth]{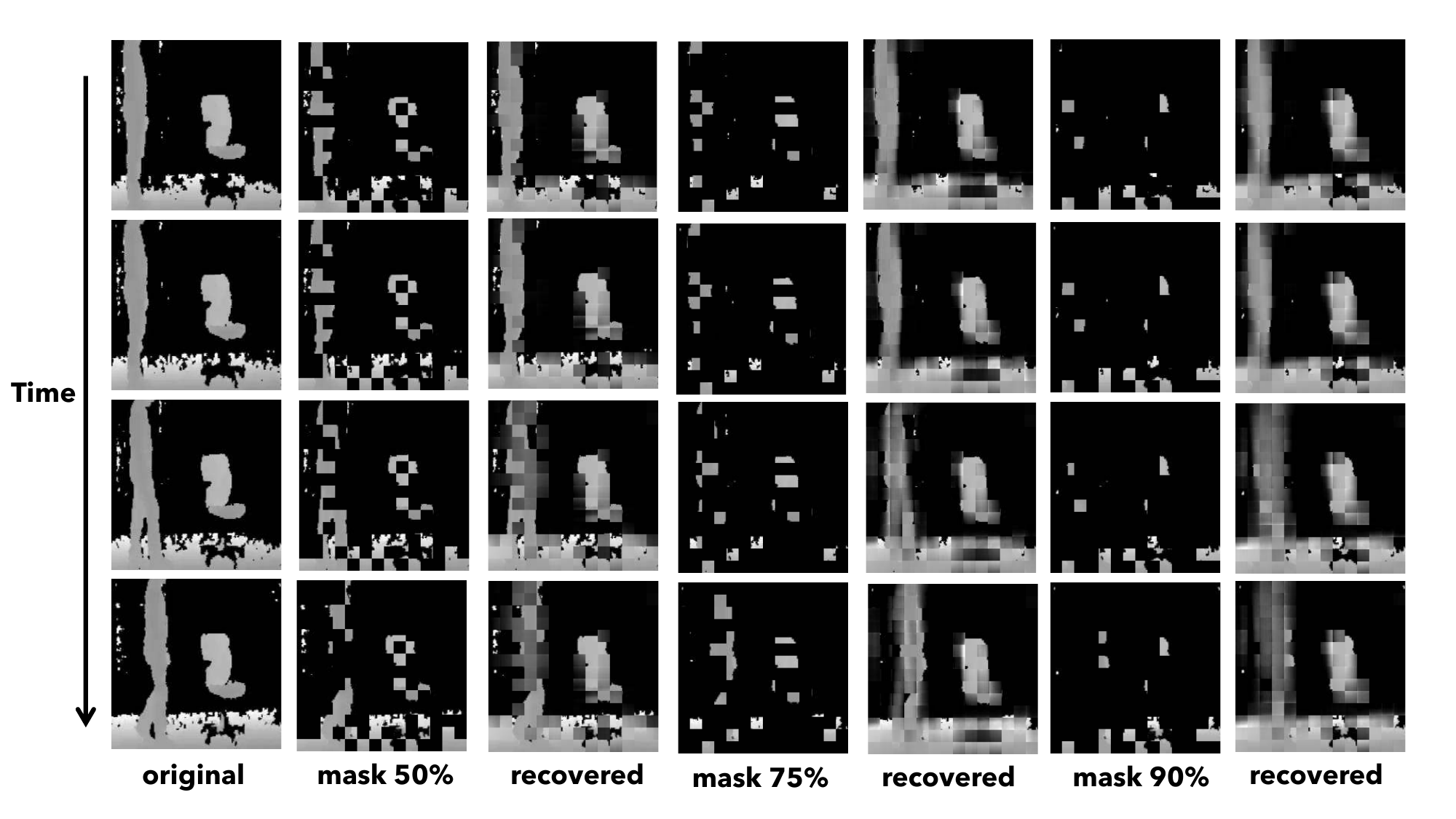}
\caption{We present our new video sequence ($N=1$, Baseline) along with reconstructions at different masking ratios. The video reconstructions are predicted by VideoMAE with the TIME layer, pre-trained using masking ratios of 50\%, 75\%, and 90\%. For this visualization, we select the action \textit{pull up the chair} from the 3D Action Pairs Depth dataset.}
\label{fig:3D_depth_1}
\end{figure*}

\begin{figure*}[tbp] 
\centering
\includegraphics[width=0.9\linewidth]{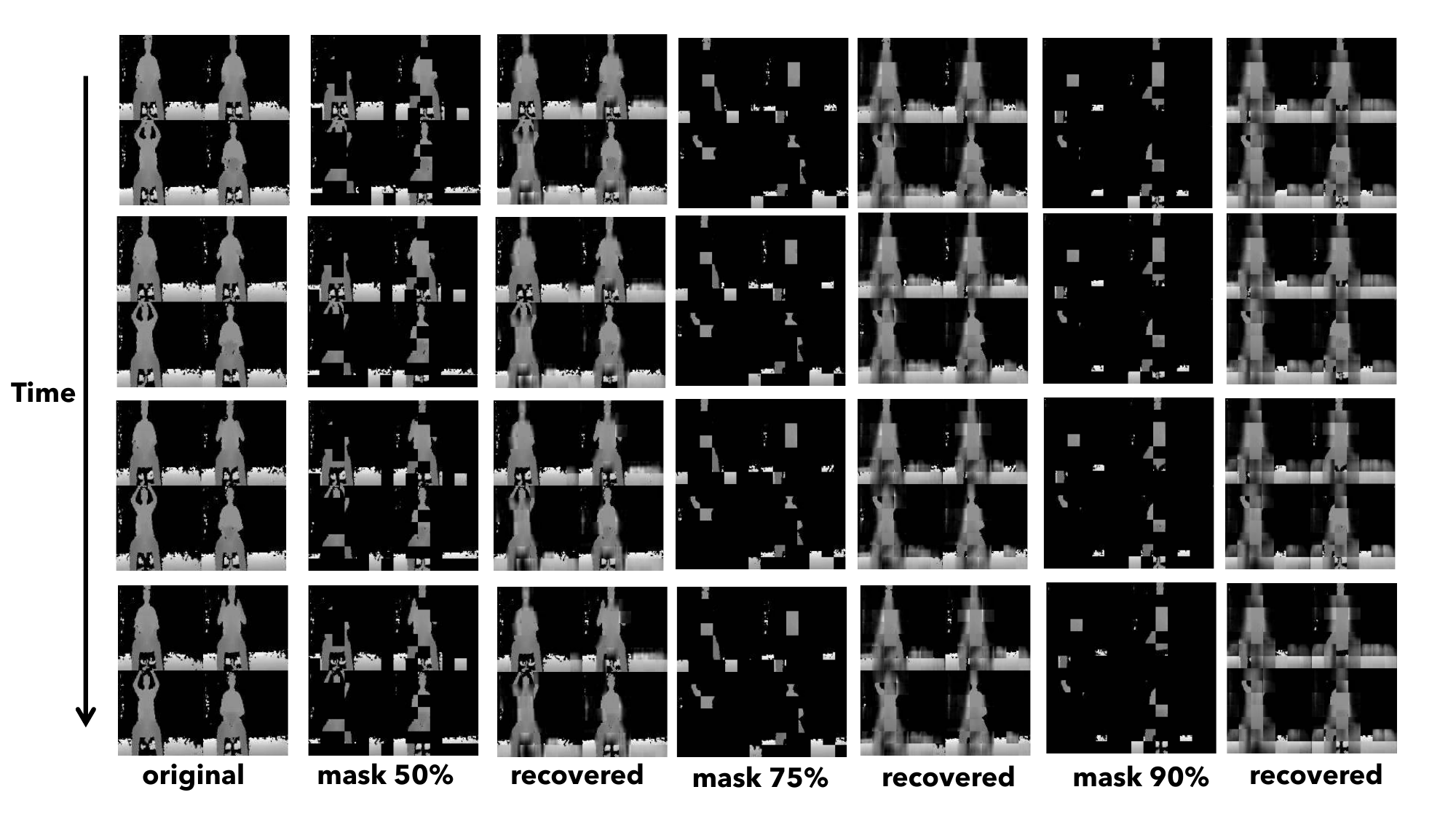}
\caption{We present our new video sequence ($N=2$) along with reconstructions at different masking ratios. The video reconstructions are predicted by VideoMAE with the TIME layer, pre-trained using masking ratios of 50\%, 75\%, and 90\%. For this visualization, we select the action \textit{take off the hat} from the 3D Action Pairs Depth dataset.}
\label{fig:3D_depth_2}
\end{figure*}

\begin{figure*}[tbp] 
\centering
\includegraphics[width=0.9\linewidth]{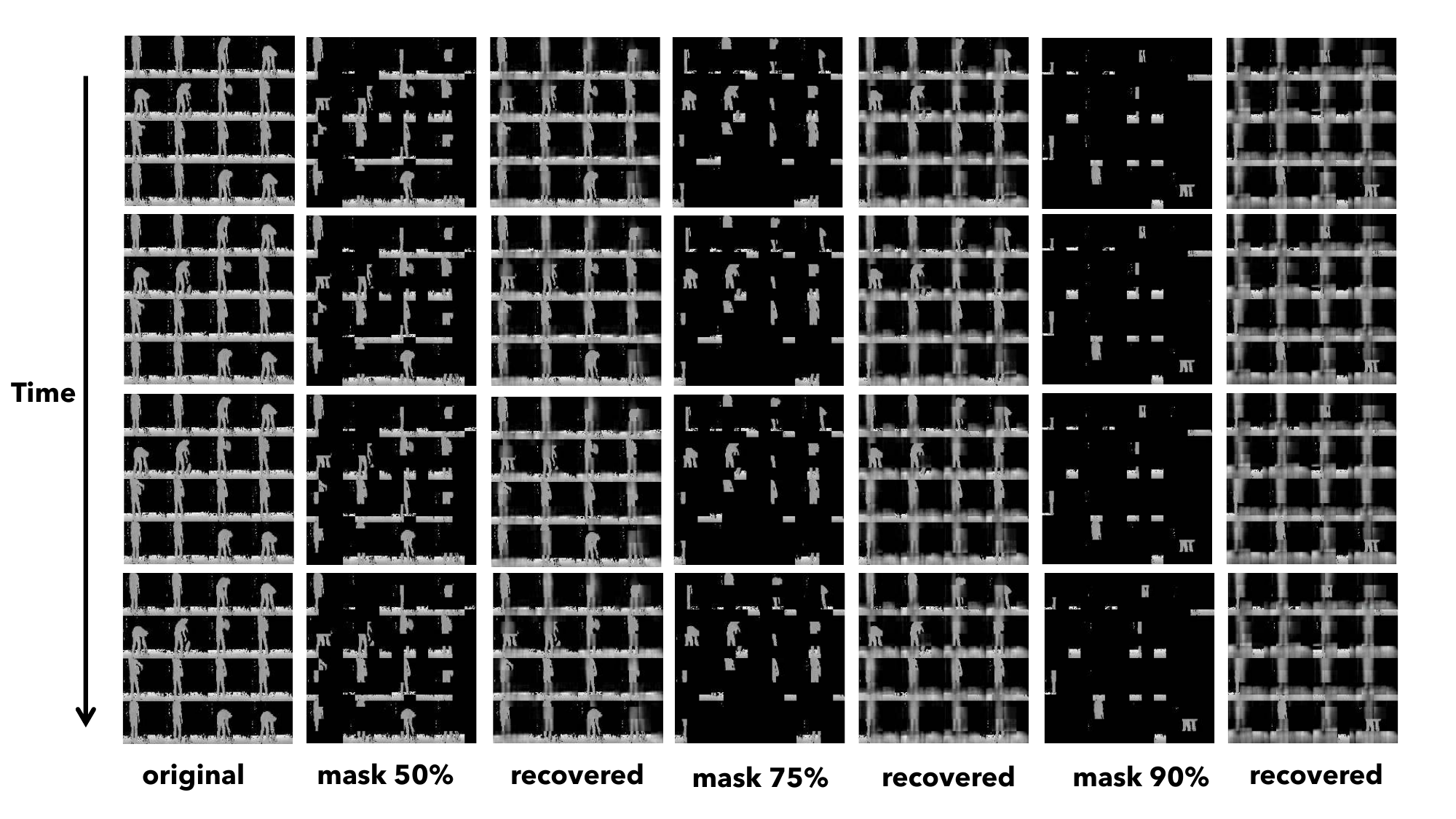}
\caption{We present our new video sequence ($N=4$) along with reconstructions at different masking ratios. The video reconstructions are predicted by VideoMAE with the TIME layer, pre-trained using masking ratios of 50\%, 75\%, and 90\%. For this visualization, we select the action \textit{pick up the bag and put it on back} from the 3D Action Pairs Depth dataset.}
\label{fig:3D_depth_4}
\end{figure*}

\begin{figure*}[tbp] 
\centering
\includegraphics[width=0.9\linewidth]{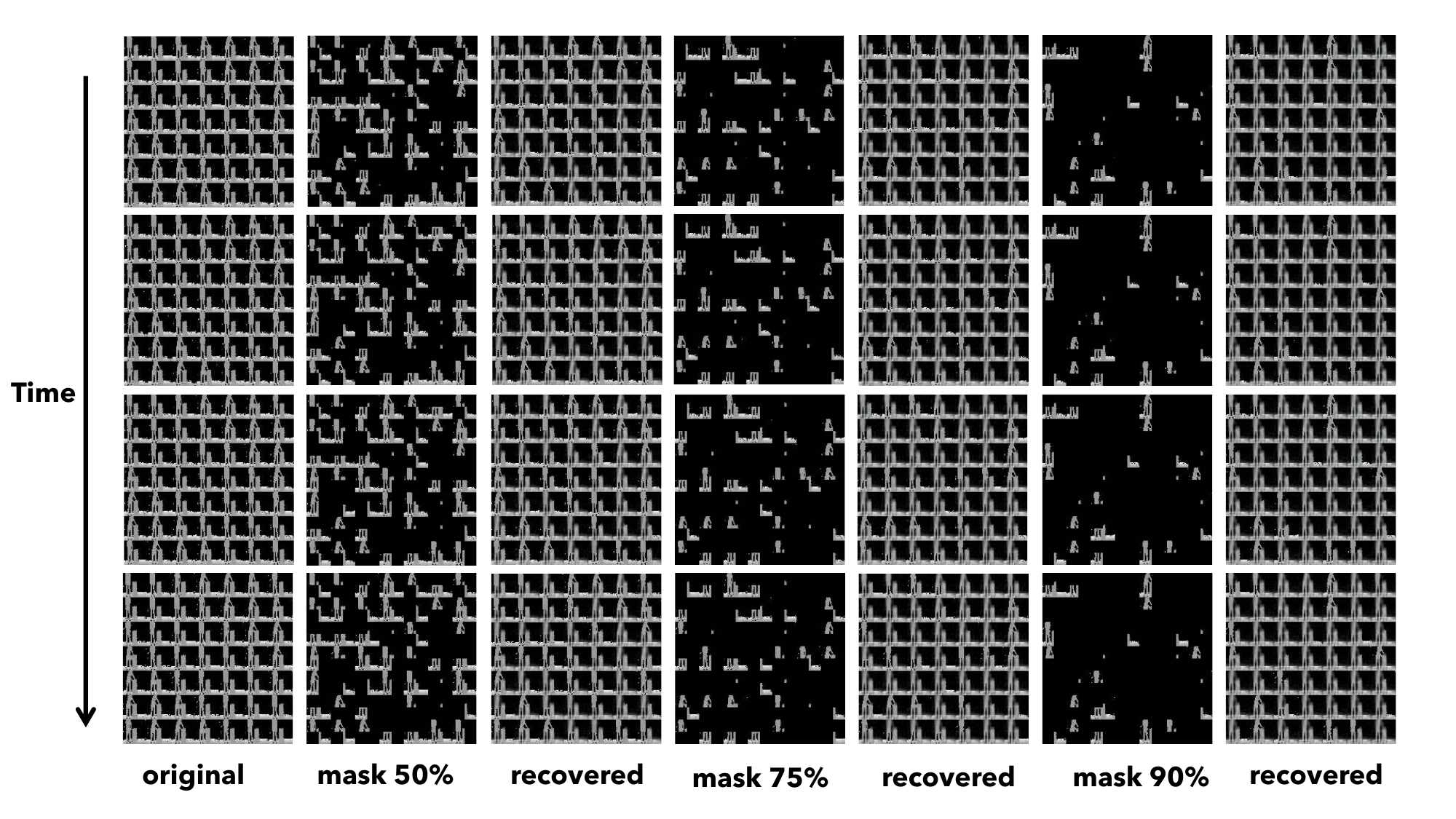}
\caption{We present our new video sequence ($N=7$) along with reconstructions at different masking ratios. The video reconstructions are predicted by VideoMAE with the TIME layer, pre-trained using masking ratios of 50\%, 75\%, and 90\%. For this visualization, we select the action \textit{pick up hanging paper from desk side} from the 3D Action Pairs Depth dataset.}
\label{fig:3D_depth_7}
\end{figure*}

\begin{figure*}[tbp] 
\centering
\includegraphics[width=0.9\linewidth]{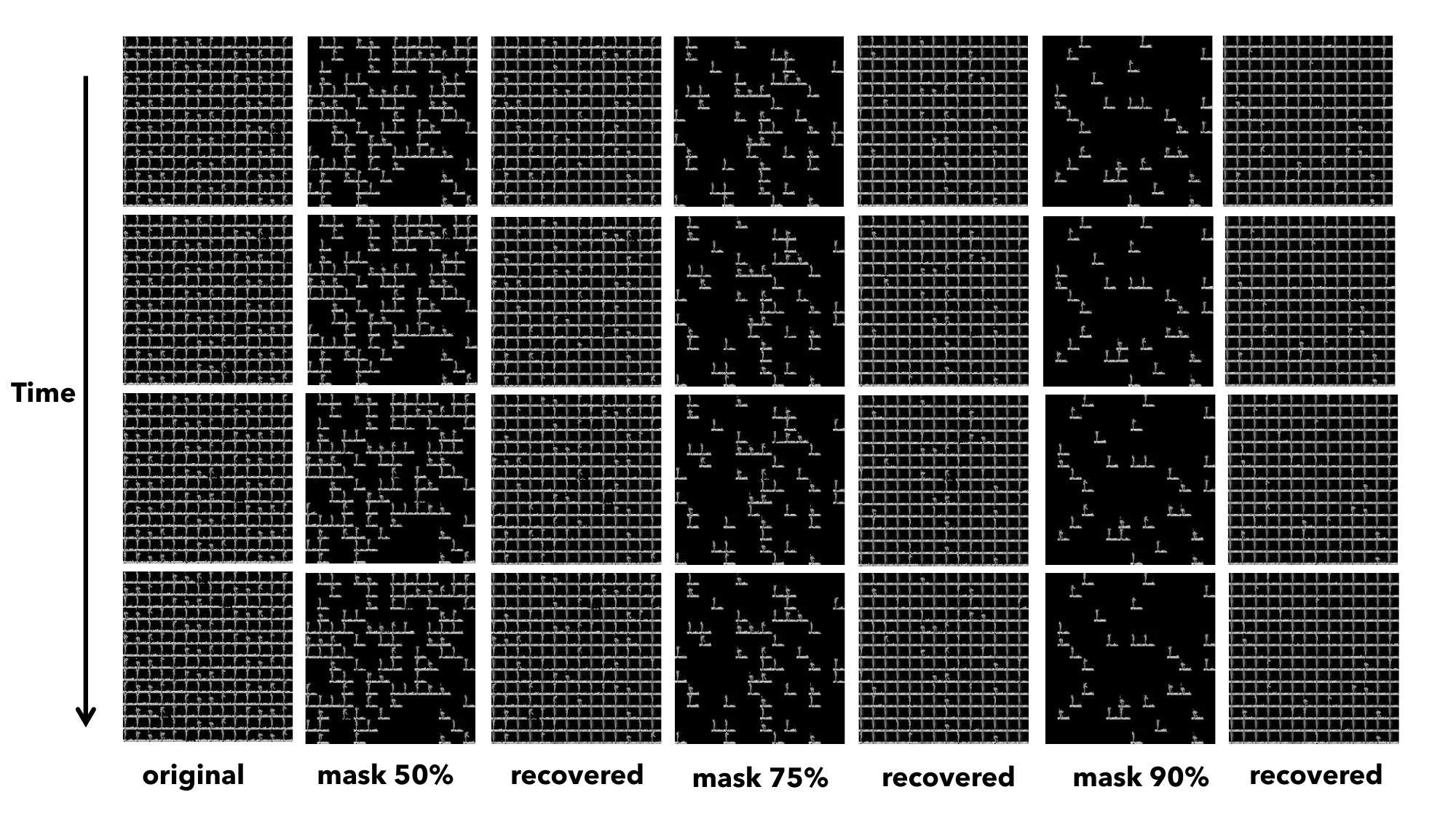}
\caption{We present our new video sequence ($N=14$) along with reconstructions at different masking ratios. The video reconstructions are predicted by VideoMAE with the TIME layer, pre-trained using masking ratios of 50\%, 75\%, and 90\%. For this visualization, we select the action \textit{pick up box on the floor} from the 3D Action Pairs Depth dataset.}
\label{fig:3D_depth_14}
\end{figure*}

Below, we present visualizations of our new video sequences with varying spatial-temporal balance parameters ($N$) and the corresponding reconstructed sequences generated by VideoMAE using masking ratios of 50\%, 75\%, and 90\% for both RGB and depth modalities.

As illustrated in~\cref{fig:3D_rgb_1} to \ref{fig:3D_depth_14}, the TIME layer effectively reconstructs temporal information in both RGB and depth video sequences. This highlights the TIME layer's ability to enhance temporal dynamics in video frames while simultaneously preserving spatial details.


\end{document}